\documentclass[journal]{IEEEtran}

\usepackage{bookmark}
\usepackage{subfig}
\usepackage{graphicx}
\usepackage{upgreek}
\usepackage{booktabs}  
\usepackage{threeparttable}  
\usepackage{multirow}
\usepackage{epstopdf}
\usepackage{hyperref} 
\usepackage{amssymb}
\usepackage{algorithm}
\usepackage{algorithmicx}
\usepackage{algpseudocode}
\usepackage{setspace}
\usepackage{color}
\usepackage{cite}
\usepackage[cmex10]{amsmath}
\usepackage{array}
\usepackage{url}

\begin{document}

\title{3D Visual Tracking Framework with Deep Learning for Asteroid Exploration}

\author{
    Dong~Zhou,~Guanghui~Sun,~\IEEEmembership{Member,~IEEE},
    and~Xiaopeng~Hong,~\IEEEmembership{Member,~IEEE}%
    \IEEEcompsocitemizethanks{
        \IEEEcompsocthanksitem D. Zhou and G. Sun are with the Department of Control Science and Engineering, Harbin Institute of Technology, Harbin, China, 150001. E-mail:  dongzhou@hit.edu.cn, guanghuisun@hit.edu.cn
        \IEEEcompsocthanksitem X. Hong is with the Faculty of Electronic and Information Engineering, Xi’an Jiaotong University, Xi'an, China, 710049. E-mail: hongxiaopeng@mail.xjtu.edu.cn
    }
}

\markboth{}%
{\MakeLowercase{\textit{Zhou et al.}}: 3D Visual Tracking Framework with Deep Learning for Asteroid Exploration}

\maketitle

\begin{abstract}
    3D visual tracking is significant to deep space exploration programs, which can guarantee spacecraft to flexibly approach the target. In this paper, we focus on the studied accurate and real-time method for 3D tracking. Considering the fact that there are almost no public dataset for this topic, A new large-scale 3D asteroid tracking dataset is presented, including binocular video sequences, depth maps, and point clouds of diverse asteroids with various shapes and textures. Benefitting from the power and convenience of simulation platform, all the 2D and 3D annotations are automatically generated. Meanwhile, we propose a deep-learning based 3D tracking framework, named as Track3D, which involves 2D monocular tracker and a novel light-weight amodal axis-aligned bounding-box network, A3BoxNet. The evaluation results demonstrate that Track3D achieves state-of-the-art 3D tracking performance in both accuracy and precision, comparing to a baseline algorithm. Moreover, our framework has great generalization ability to 2D monocular tracking performance. 
\end{abstract}


\IEEEpeerreviewmaketitle

\section{Introduction}
\IEEEPARstart{M}{illions} of asteroids exist in solar system, many the shattered remnants of planetesimals, bodies within the young Sun's solar nebula that never grew large enough to become planets \cite{kortenkamp_asteroids_2011}. The vast majority of known asteroids orbit within the main asteroid belt located between the orbits of Mars and Jupiter. To further investigate how planets formed and how life began, as well as improve our understanding to asteroids that could impact Earth, some deep exploration programs were proposed, e.g. \emph{Hayabusa}\cite{fujiwara_rubble-pile_2006, yoshikawa_hayabusa_2006}, \emph{Hayabusa2}\cite{sugita_geomorphology_2019, watanabe_hayabusa2_2017}, and \emph{OSIRIS-Rex}\cite{lauretta2017osiris, golish2020ground}. The program objectives involve orbiting observation, autonomous landing, geological sampling, and so on. 

\begin{figure}[t]
	\centering
	\subfloat[Itokawa]{
        \includegraphics[width=0.1\textwidth]{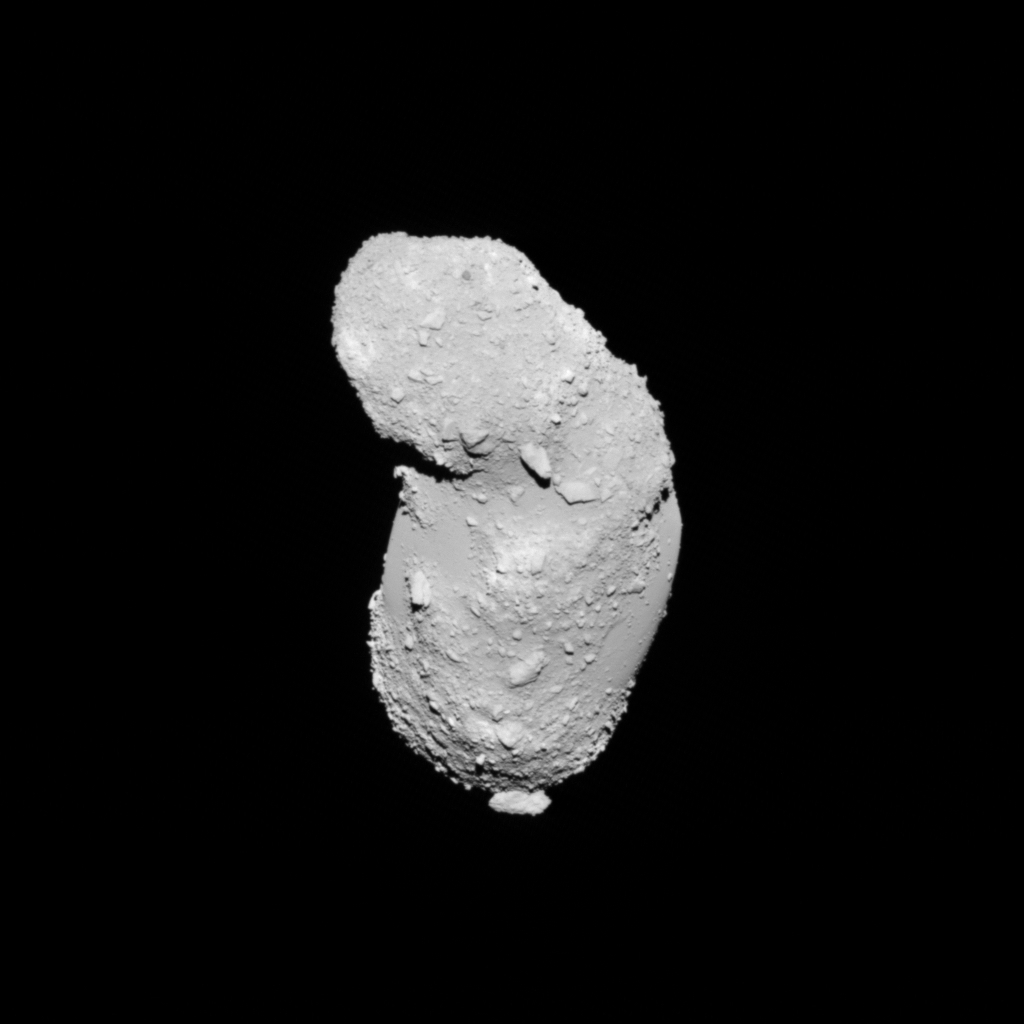} 
        \includegraphics[width=0.1\textwidth]{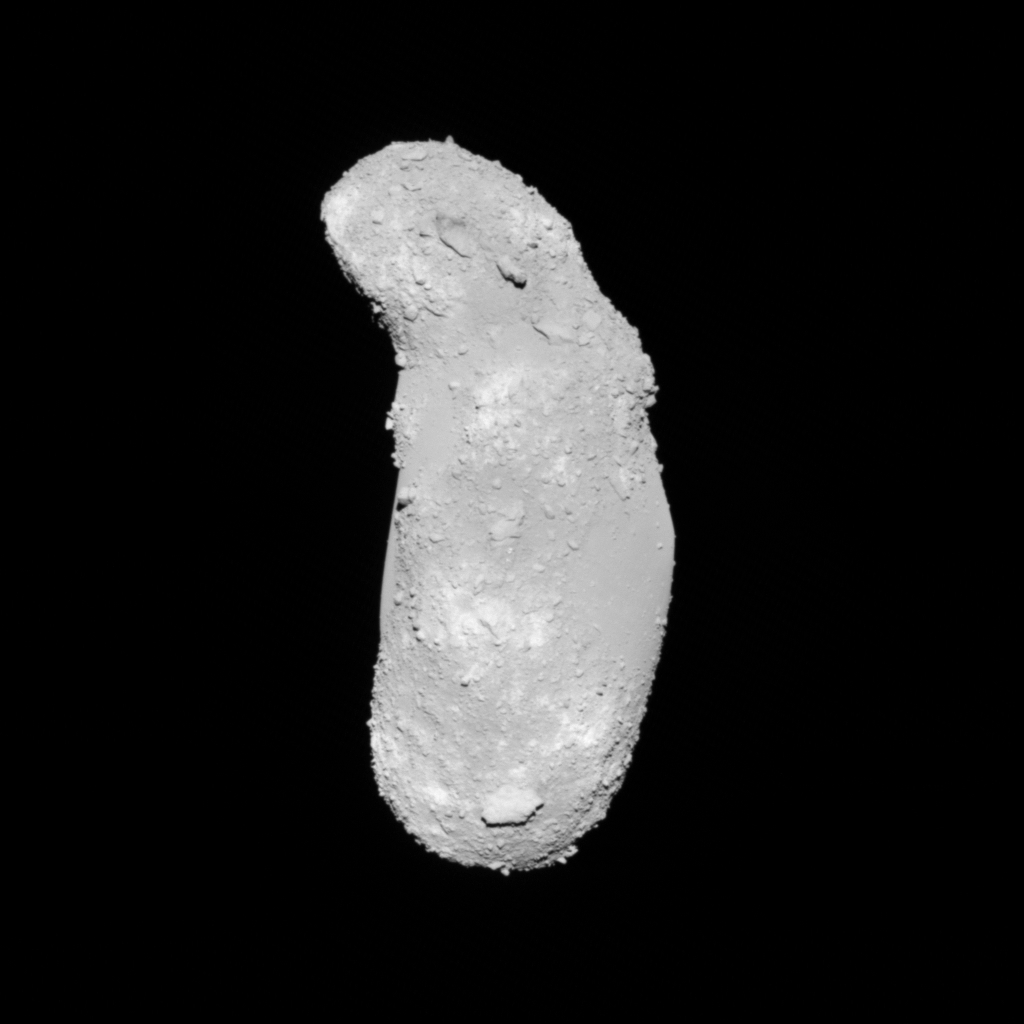}
        \includegraphics[width=0.1\textwidth]{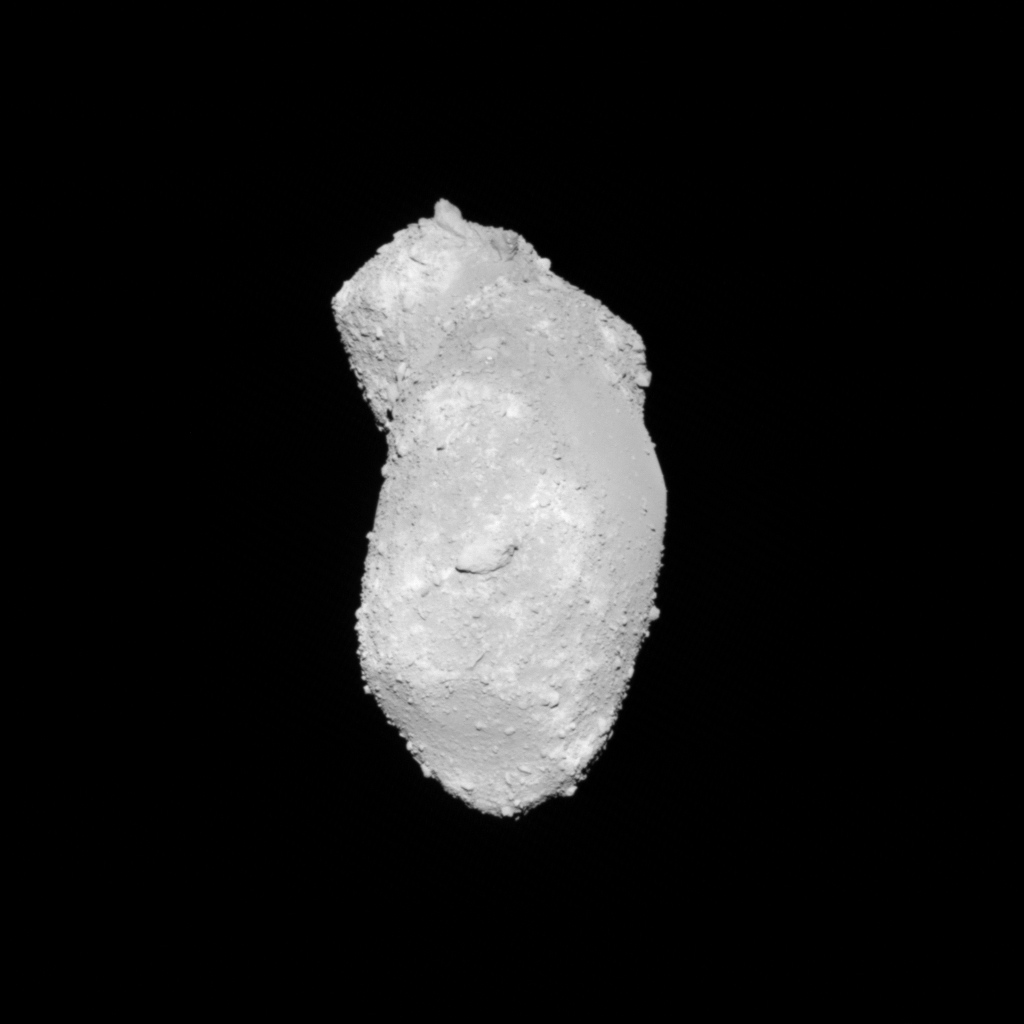}
        \includegraphics[width=0.1\textwidth]{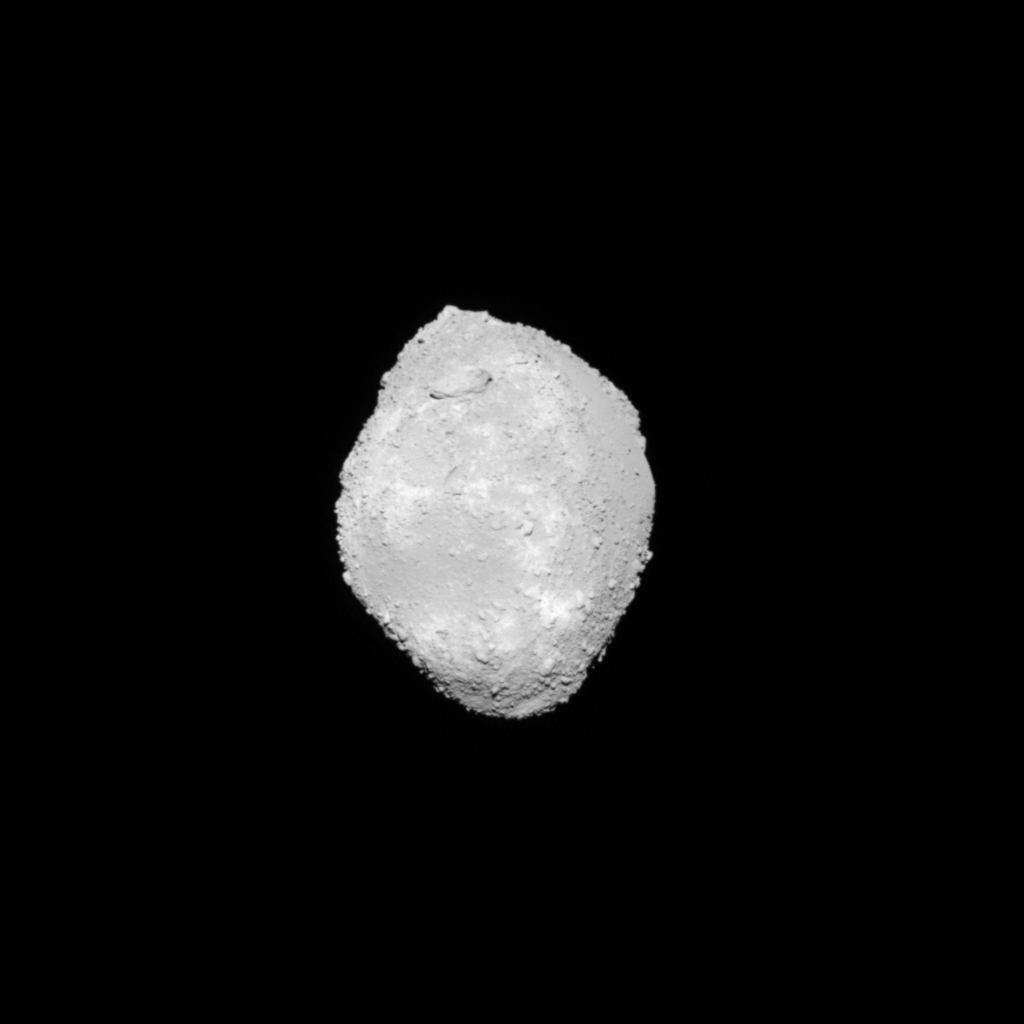}
	    \label{fig1_a}
    }
	\hfil
	\subfloat[Ryugu]{
        \includegraphics[width=0.1\textwidth]{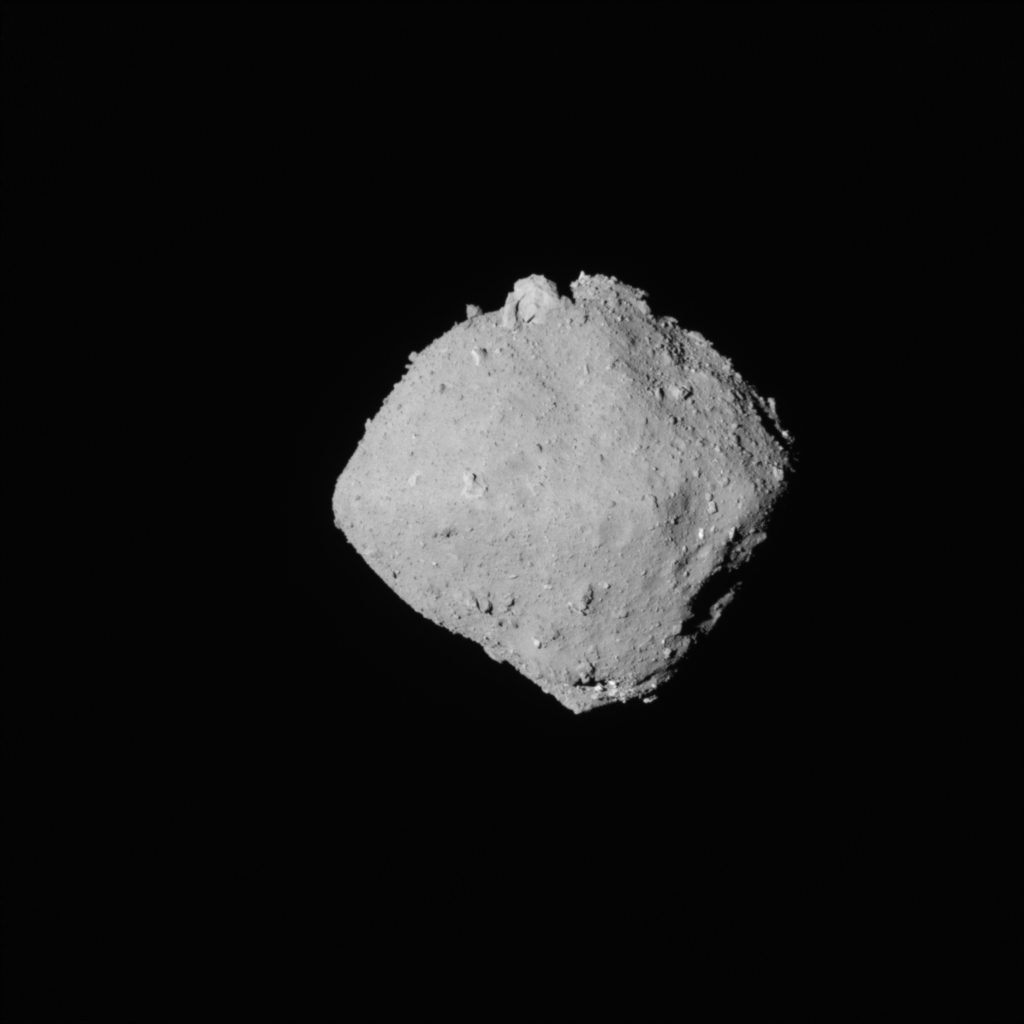}
        \includegraphics[width=0.1\textwidth]{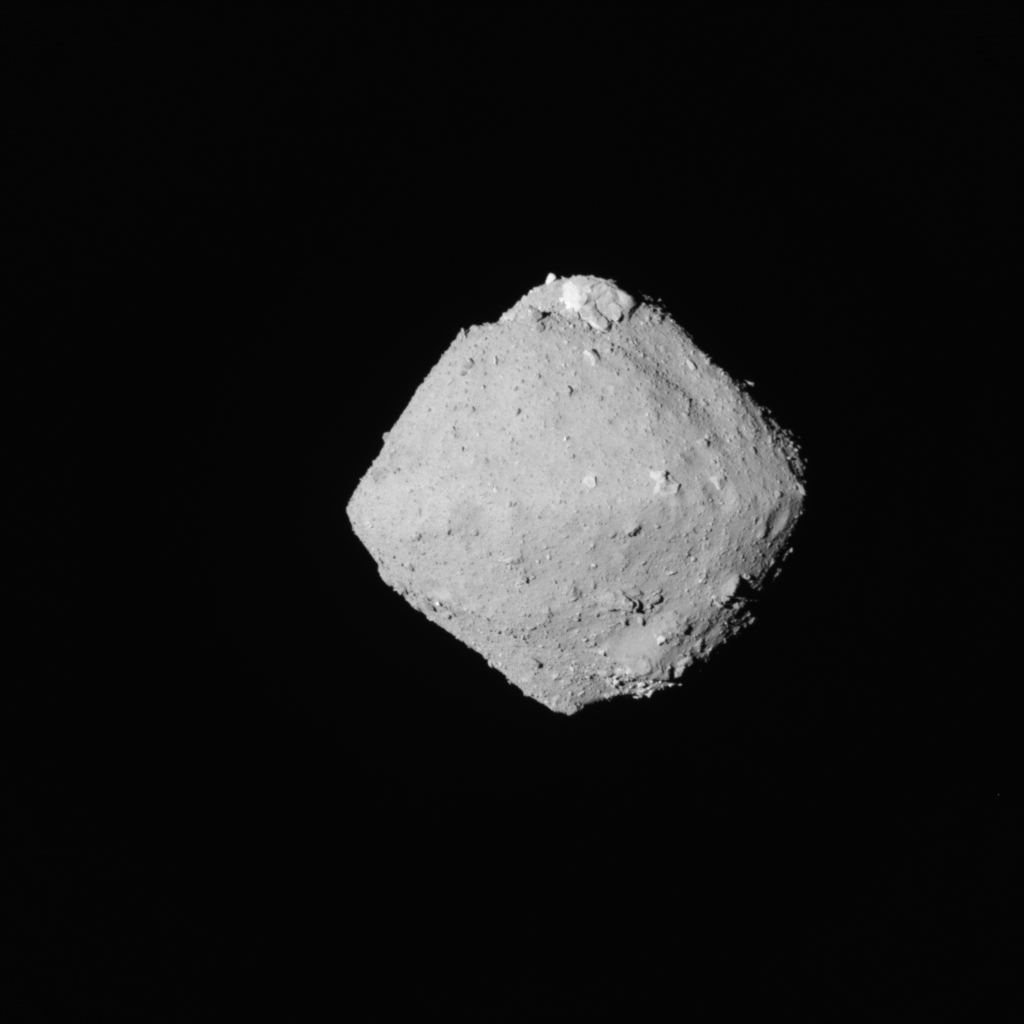}
        \includegraphics[width=0.1\textwidth]{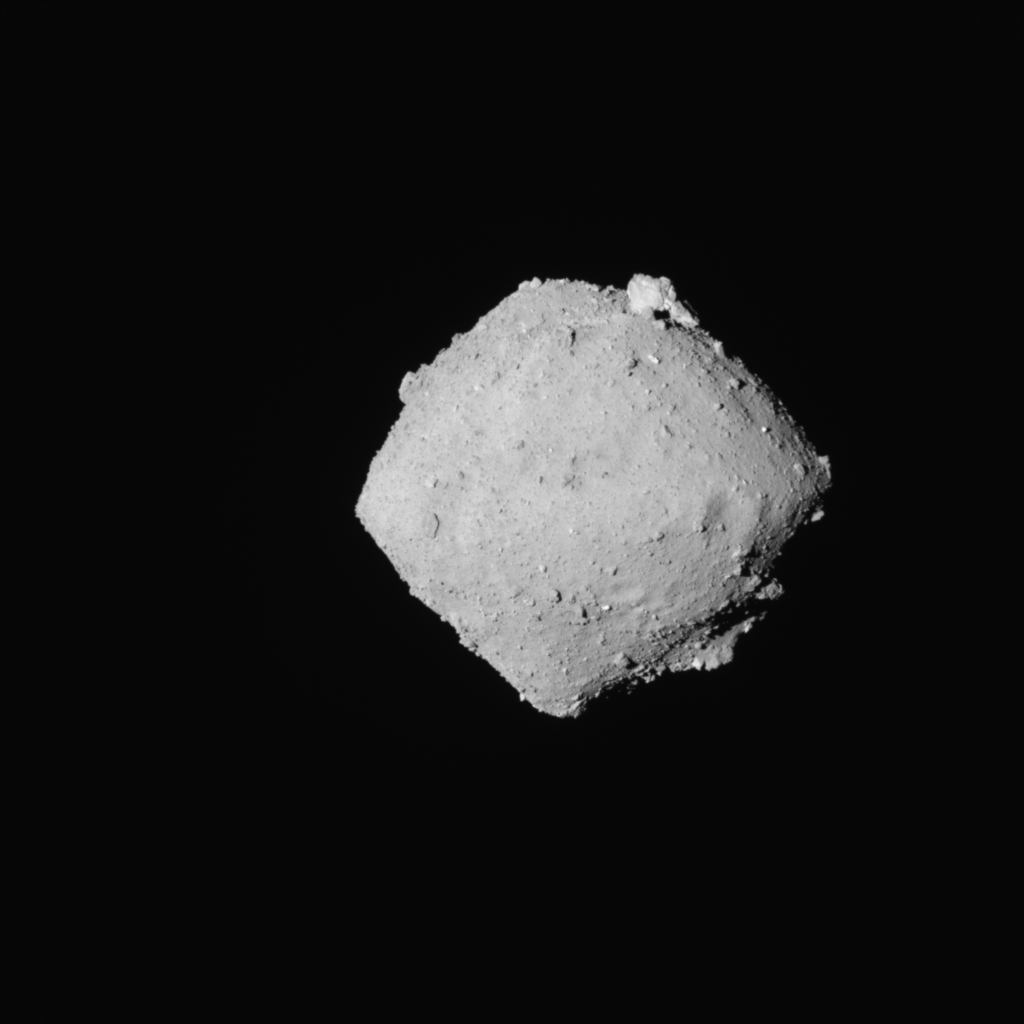}
        \includegraphics[width=0.1\textwidth]{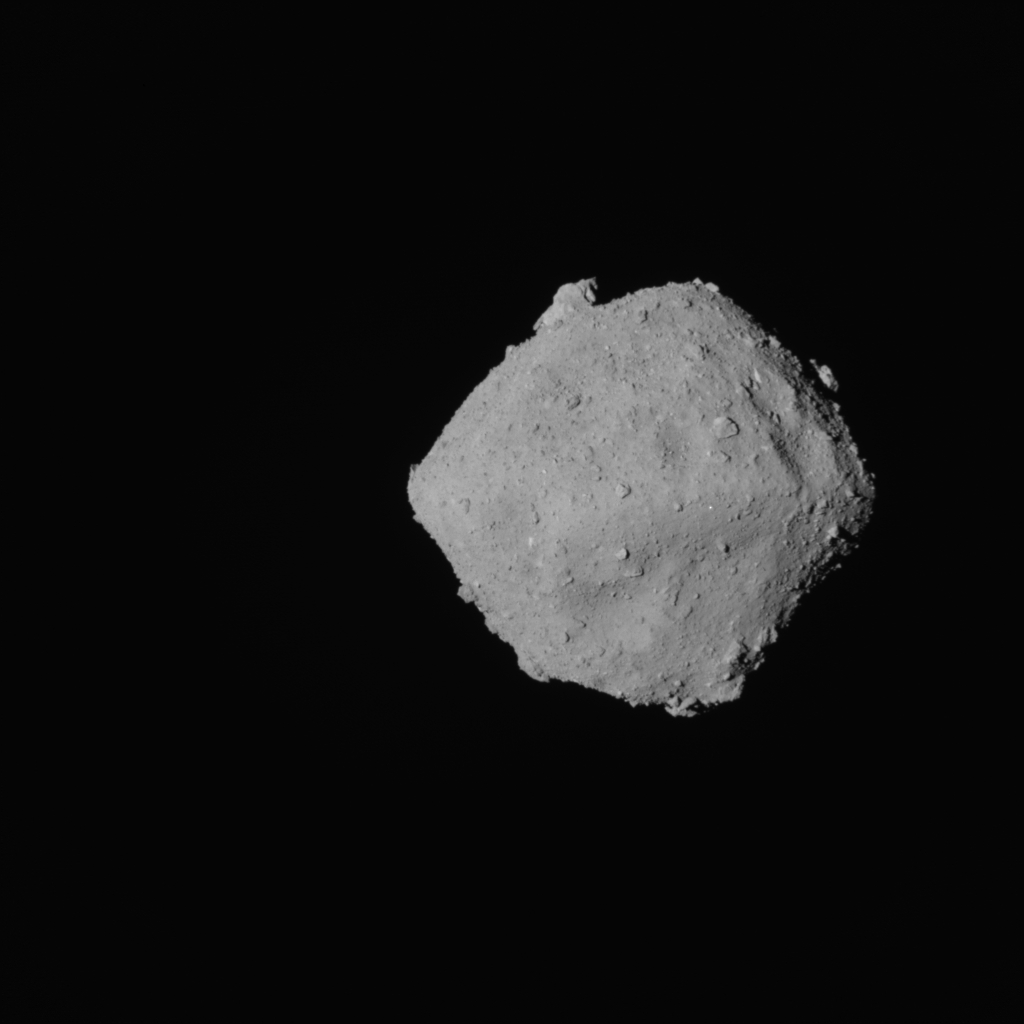}
	    \label{fig1_b}
    }
	\caption{Two mission objectives of \emph{Hayabusa} and \emph{Hayabusa2}. (a) Itokawa, a potato-shaped asteroid with 600-meter size, is named after Hideo Itokawa, a Japanese rocket pioneer. (b) Ryugu, a more primordial C-type asteroid with 900-meter size, is discovered on May 1999 by LINEAR telescope.
    }
	\label{fig1}
\end{figure}

Two near-Earth asteroids, Itokawa \cite{yoshikawa_hayabusa_2006} and Ryugu \cite{sugita_geomorphology_2019}, with complex 6-DoF motion are shown in Fig. \ref{fig1}. It is obvious that 3D visual tracking system is important to explore these two asteroids, which can provide object location, size, and pose. It’s also of great significance to spacecraft autonomous navigation, asteroid sample collection, and universe origin study. However, state-of-the-art 4-DoF trackers \cite{giancolaLeveragingShapeCompletion2019, qiP2BPointtoBoxNetwork2020, yinCenterBased3DObject2021} presented for automous driving are confused about heading angle of asteroid, which makes inaccurate 3D bounding-box estimation. Besides, some 6-DoF tracking methods \cite{prisacariuPWP3DRealTimeSegmentation2012, prisacariuSimultaneousMonocular2D2013, crivellaroRobust3DObject2018a} under strong assumptions are also impractical to track asteroid. To be honest, constructing an end-to-end deep network that predicts 6-DoF states of asteroid is pretty difficult. We therefore decompose 3D asteroid tracking problem into 3-DoF tracking and pose estimation. And this paper merely focus on the 3-DoF tracking part.

\begin{figure*}[ht]
    \centering
	\includegraphics[width=\textwidth]{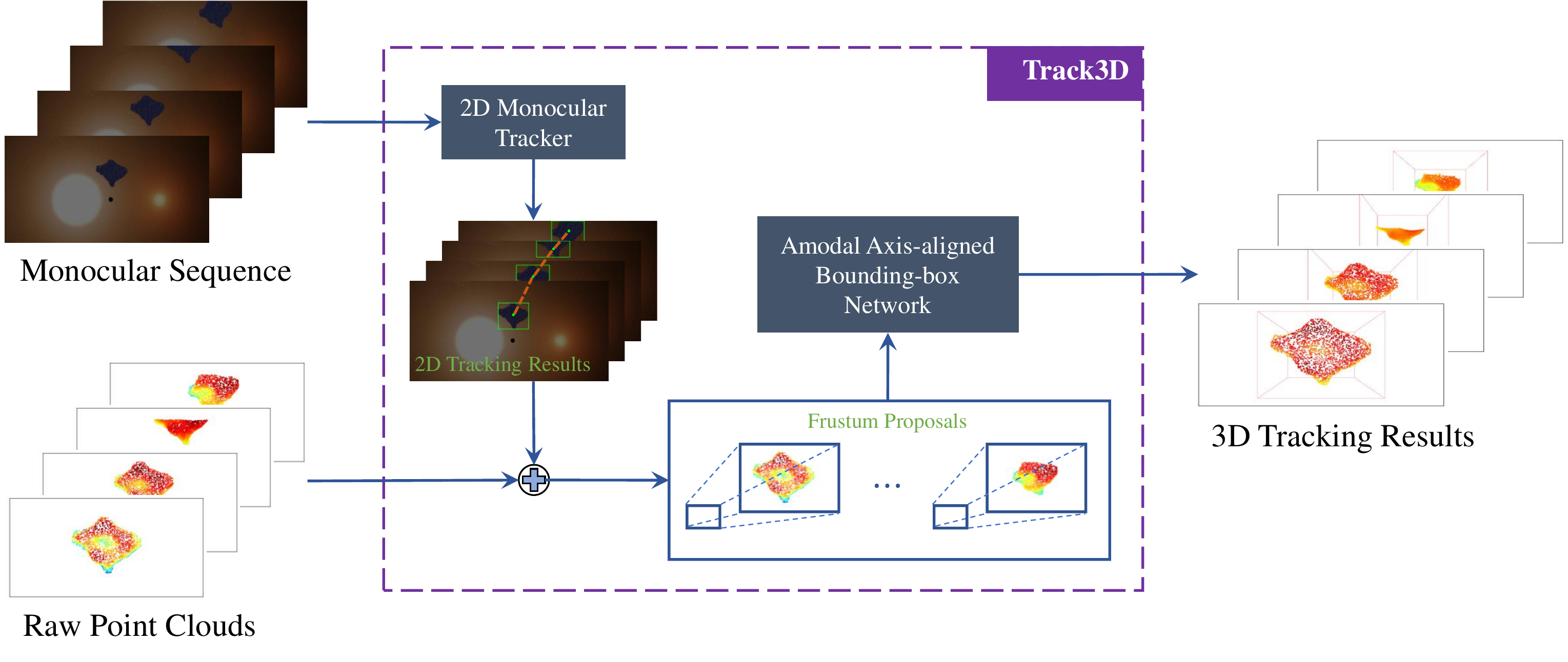}
    \caption{The deep-learning based 3D tracking framework for asteroid, named as Track3D, which consists of 2D monocular tracker and amodal axis-aligned bounding-box network, A3BoxNet. Experiment shows our framework can run at 77 FPS with high accuracy. And it has great generalization abitlity to 2D tracking algorithm.}
    \label{fig2}
    \centering
\end{figure*}

Inspired by the idea of 2D-driven 3D perception, we present a novel deep-learning based 3D asteroid tracking framework, Track3D. As shown in Fig. \ref{fig2}, it mainly consists of 2D monocular tracker and a light-weight amodal axis-aligned bounding-box network, A3BoxNet, which can predict accurate target center and size purely relied on partial object point cloud. Extensive experiments show that Track3D reaches state-of-the-art 3D tracking results (0.669 $AO^{3d}$ at 77 FPS) and has great generalization ability to 2D monocular tracker. Moreover, we discover that our framework with 2D-3D tracking fusion strategy can make significant improvement on 2D tracking performance. 

However, there are few studies on 3D asteroid tracking, as well as relevant dataset, which have greatly hindered the development of asteroid exploration. To this end, we construct the first large-scale 3D asteroid tracking dataset, by acquiring 148,500 binocular images, depth maps, and point clouds of diverse asteroids in various shapes and textures with physics engine. Benefitting from the power and convenience of physics engine, all the 2D and 3D annotations are automatically generated. Meanwhile, we also provide an evaluation toolkit, which includes 2D monocular and 3D tracking evaluation algorithms.

Our contributions in this paper are summarized as follows:
\begin{itemize}
    \item Considering different types of asteroid with various shapes and textures, we construct the first large-scale 3D asteroid tracking dataset, including 148,500 binocular images, depth maps, and point clouds. 

    \item The first 3-DoF asteroid tracking framework, Track3D, is also presented, which involves an 2D monocular tracker and A3BoxNet network. The experimental results show the impressive advancement and generalization of our framework, even based on poor 2D tracking algorithm.

    \item We propose a novel light-weight amodal bounding-box estimation network, A3BoxNet, which can predict accurate axis-aligned bounding-box of target merely with partial object point cloud. Randomly sampling 1024 points as network input, A3BoxNet can even achieve 0.712 $AO^{3d}$ and 0.345 $ACE^{3d}$ with up to 281.5 FPS real-time performance.
     
\end{itemize}

The rest of this paper is ordered in the subsequent sections. In Section \ref{section2}, we review some related works from two aspects: visual tracking in aerospace domain and 3D object tracking. In Section \ref{section3}, the constructing details of 3D asteroid tracking dataset are presented, involving simulation platform, annotation, and evaluation metrics. We also propose a novel deep-learning based 3D tracking framework in Section \ref{section4}, which consists of 2D monocular tracker, and a light-weight amodal bounding-box estimation network, A3BoxNet. Section \ref{section5} introduces more details about 3D tracking framework performance and ablation study. Finally, we make a conclusion in Section \ref{section6}.

\begin{figure*}[t]
    \centering
    \includegraphics[width=\textwidth]{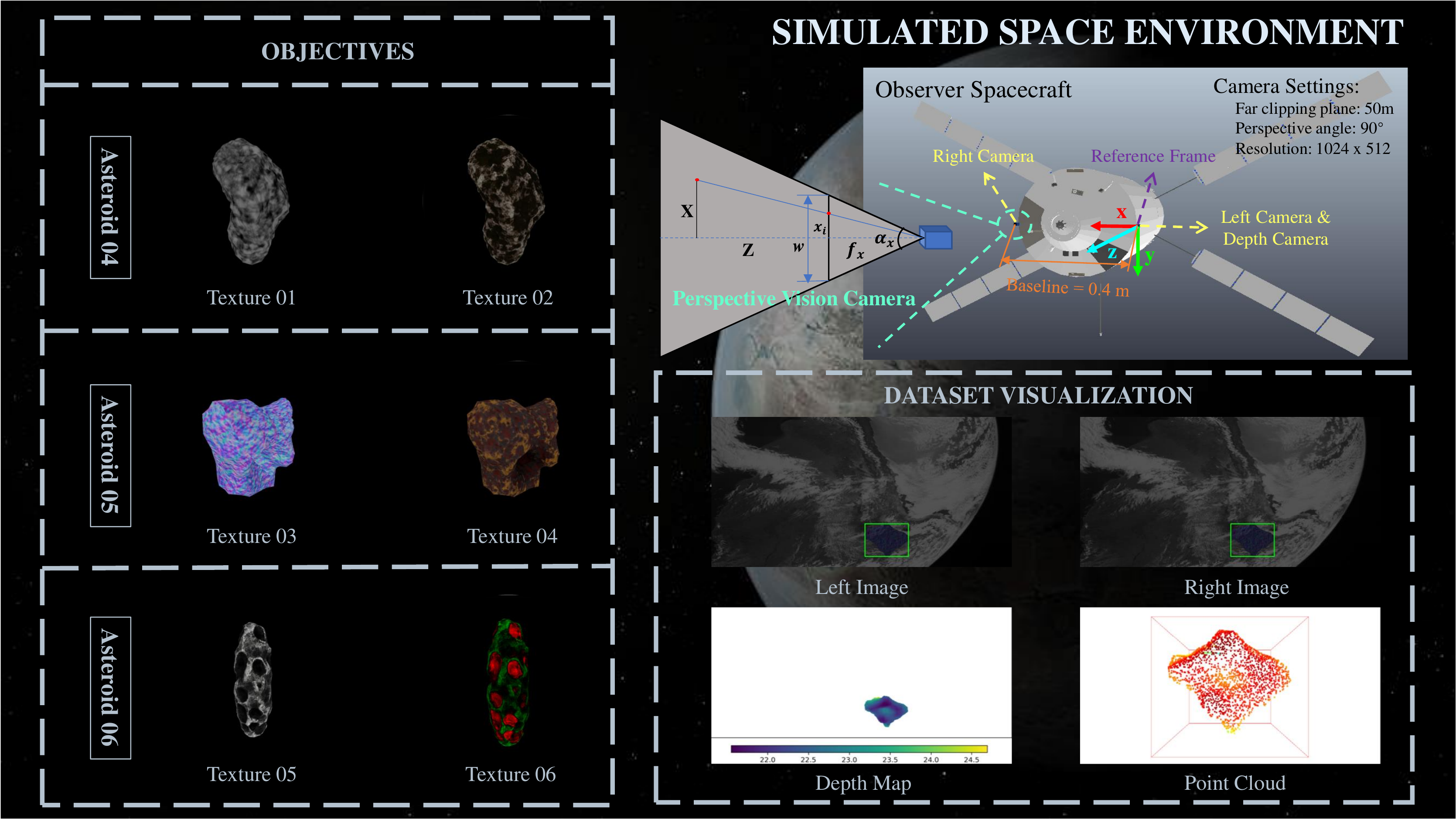}
    \caption{The simulated platform for constructing 3D asteroid tracking dataset. We design three types of 3D asteroid model with 6 colorful textures, of which size and luminosity are modified during data collection. The setups of visual observer spacecraft and imaging mechanism of perspective vision camera are also shown in upper right corner. We pre-align depth map and point cloud to reference frame (i.e. left camera coordinate system). The left and right vision cameras have the same parameters, including far clipping plane, perspective angle, and resolution. In addition, some screenshots of 3D asteroid tracking dataset with 2D and 3D annotations are also visualized at bottom right.}
    \label{fig3}
    \centering
\end{figure*}

\section{Related Work \label{section2}}
\subsection{Visual Tracking in Aerospace}
The goal of visual tracking is to precisely and robustly perceive real-time states (e.g. location, velocity and size) of an instereted target in complex and dynamic environment. As the basis for pose estimation, behavior understanding, and scene interpretation, visual object tracking has wide application in space debris removel \cite{agliettiActiveSpaceDebris2020,huangDexterousTetheredSpace2017}, space robotic inspection \cite{fourieFlightResultsVisionBased2014}, spacecraft rendezvous and docking \cite{petitVisionbasedSpaceAutonomous2011,sharmaPoseEstimationNoncooperative2018}. 

At present, most of visual trackers in aerospace domain mainly focus on feature-based method. Huang \cite{huangDexterousTetheredSpace2017} proposed a novel feature tracking algorithm. Firstly, it extracted features by SURF detector. And then, the pyramid-Kanade-Lucas-Tomasi (P-KLT) algorithm was adopted to match key-points between two adjacent frames. Finally, accurate target bounding box is obtained with Greedy Snake method. A feature-tracking scheme was also presented in \cite{volpePassiveCameraBased2018}  that combines traditional feature-point detector and frame-wise matching to track a non-cooperative and unknown satellite. Felicetti \cite{felicettiImagebasedAttitudeManeuvers2018} put forward an active space debris visual tracking method, in which the chaser satellite can keep the moving object in field of view of optical camera by continuously pose correction. Those feature-based trackers heavily relied on the manually designed feature detector and cannot handle extreme cases in space (e.g. illumination variation, scale variation, fast motion, rotation, truncation, and background clutters).

With many challenges and benchmarks emerging, such as OTB2015 \cite{wuObjectTrackingBenchmark2015}, YouTube-BB \cite{realYouTubeBoundingBoxesLargeHighPrecision2017}, ILSVRC-VID \cite{russakovskyImageNetLargeScale2015}, GOT-10K \cite{huangGOT10kLargeHighDiversity2019}, and Visual Object Tracking challenges \cite{kristanVisualObjectTracking2017, kristanSeventhVisualObject2019}, the development of generic object tracking is far beyond imagination \cite{lanLearningCommonFeatureSpecific2018}. Especially, deep learning based trackers \cite{bertinettoFullyconvolutionalSiameseNetworks2016, wangFastOnlineObject2018, liSiamRPNEvolutionSiamese2018, danelljanATOMAccurateTracking2018, zhangDeeperWiderSiamese2019, chenSiameseBoxAdaptive2020, ondrasovicSiameseVisualObject2021} have dominated the whole tracking community in recent years, because of its striking performance. Generic object trackers often follow one protocol that no prior knowledges is available. This hypothesis is naturally suitable for asteroid visual tracking, because of high uncertainty of vision tasks in space. In paper \cite{zhou2DVisionbasedTracking2021}, most of state-of-the-art generic trackers had been evaluated on space non-cooperative object visual tracking (SNCOVT) dataset, which provides firm research foundation for our work.

However, recent visual trackers mainly cope with RGB \cite{zhangOceanObjectawareAnchorfree2020, yanLearningSpatioTemporalTransformer2021, chenTransformerTracking2021}, RGB-Thermal \cite{lanModalitycorrelationawareSparseRepresentation2020, lanLearningModalityConsistencyFeature2019, lanRobustMultimodalityAnchor2019}, and RGB-Depth \cite{songTrackingRevisitedUsing2013} video sequences, which only provides poor target information in 2D space and heavily restricts practical applications of visual tracking. In contrast, 3D visual tracking is more promising and competitive. To this end, we propose a novel deep-learning based 3D tracking framework for asteroid exploration.

\begin{table*}[t]
    \centering
    \caption{The default object size categories and corresponding ratio of XYZ.}
    \label{table1}
    \begin{tabular}{ccccccccccccccc}
        \toprule
        category & 0 & 1 & 2 & 3 & 4 & 5 & 6 & 7 & 8 & 9 & 10 & 11 & 12 & 13 \\
        \midrule
        X & 1 & 1/2 & 1/3 & 2/3 & 1 & 1 & 1 & 1 & 1/2 & 2/3 & 1/2 & 2/3 & 1 & 1\\
        Y & 1 & 1 & 1 & 1 & 1/2 & 2/3 & 1 & 1 & 1/2 & 2/3 & 1 & 1 & 1/2 & 2/3 \\
        Z & 1 & 1 & 1 & 1 & 1 & 1 & 1/2 & 2/3 & 1 & 1 & 1/2 & 2/3 & 1/2 & 2/3 \\
        \bottomrule  
    \end{tabular}
    \centering
\end{table*}

\subsection{3D Object Tracking}
Although there are plenty of related works \cite{prisacariuPWP3DRealTimeSegmentation2012, heldPrecisionTrackingSparse2013, prisacariuSimultaneousMonocular2D2013, asvadi3DObjectTracking2016, crivellaroRobust3DObject2018a, kartObjectTrackingReconstruction2019, giancolaLeveragingShapeCompletion2019, qiP2BPointtoBoxNetwork2020, yinCenterBased3DObject2021}, the concept of 3D object tracking remains ambiguous. Hence that, we first define that 3D object tracking is to obtain 3D information of any target in real-time by leveraging various visual sensors, given initaial states at the beginning. According to the degrees of freedom of tracking result, all the 3D trackers can be concluded into three categories: 3-DoF, 4-DoF, and 6-DoF Tracker. Obviously, the more degrees of freedom, the more difficult 3D tracking task is. 

3-DoF tracking means that both 3D location and size of object (i.e. 3D axis-aligned bounding-box) should be estimated. However, most of researches only considered predicting 3D object center. In paper \cite{heldPrecisionTrackingSparse2013}, a color-augmented search algorithm was presented to obtain the position and velocity of vehicle. Asvadi et al. \cite{asvadi3DObjectTracking2016} utilized two parallel mean-shift localizers in image and pcd space, and made fusion for two localizations by Kalman filter. This algorithm can effectively achieve low 3D center error. To the best of our knowledge, the framework proposed in this paper is the first real 3-DoF tracker that can be applied for collision avoidance, 3D reconstruction, and pose estimation.

Comparing with 3-DoF methods, 4-DoF tracker often needs to predict an extra heading angle of target, which is original from 3D tracking requirement in autonomous driving. Giancola et al. \cite{giancolaLeveragingShapeCompletion2019} presented a novel 3D siamese tracker with the regularization of point cloud completion. However, the exhaustively searching for candidate shapes in this method would consume high computational cost. Qi et al. \cite{qiP2BPointtoBoxNetwork2020} also proposed point-wise tracking paradigm, P2B, which addressed 3D object tracking by potential center localization, 3D target proposal and verification. Since this method purely uses point cloud data as input, it is vulnerable to initial object point cloud and only achieves 0.562 $AO^{3d}$ at 40 FPS on KITTI tracking dataset \cite{geigerAreWeReady2012}. 

In paper \cite{prisacariuPWP3DRealTimeSegmentation2012, prisacariuSimultaneousMonocular2D2013}, 6-DoF tracking task was considered as joint 2D segmentation and 3D pose estimation problem, and the method looked for the pose that best segmented the target object from the background. Crivellaro et al. \cite{crivellaroRobust3DObject2018a} present a novel 6-DoF tracking framework with monocular image, including expensive 2D detector, local pose estimation and extended kalman filter. However, the non-textured 3D model of target should be given in this method, which heavily limits the scope of its application. To be honest, the asteroid tracking is one of 6-DoF tracking tasks. We think it is very hard to construct an end-to-end network that can directly and precisely predict the 6-DoF object states. Therefore, we decompose asteroid tracking problem into 3-DoF tracking and pose estimation. It is worthwhile noting that this paper simply focus on the 3-DoF tracking task of asteroid.

\section{3D Asteroid Tracking Dataset \label{section3}}
To promote the research on 3D tracking for asteroid, we construct a large-scale 3D asteroid tracking dataset, including binocular video sequences, depth maps and point clouds. In addition, 3D tracking evaluation toolkit is also provided for performance analysis. More details about 3D asteroid tracking dataset are introduced in this section.

\subsection{Dataset Construction}
There is no doubt that collecting real data to create large-scale 3D asteroid tracking dataset is impractical, like KITTI dataset \cite{geigerAreWeReady2012} and Princeton RGB-D Tracking dataset \cite{songTrackingRevisitedUsing2013}. Meanwhile, constructing dataset by ground simulation \cite{zhou2DVisionbasedTracking2021} is very expensive and limited. Inspired by UAV123 dataset \cite{muellerBenchmarkSimulatorUAV2016}, we therefore consider to collect rich 3D tracking data by virtual physics engine, V-rep. The power and convenience of physics engine make automatic data labelling possible, which greatly reduces constructing cost and improves annotation accuracy.

The critical foundation of dataset constructing based on physics engine is 3D modeling of asteroids with diverse shapes and textures. We create three types of 3D asteroid model (i.e. Asteroid04, 05, and 06) with 6 different textures, which are illustrated at the left of Fig. \ref{fig3}. From point of our view, one asteroid model with diferent textures can be considered as different fine-grained categories. In addition, we introduce 9 simulated space scenes into dataset construction. All the 3D asteroid models have been controlled by scripts to carry out random 6-DoF motion in simulated scenes. 

The detailed setup of vision sensors for data collection can be seen at upper right corner of Fig \ref{fig3}. Only two perspective vision cameras are equipmented with 0.4-meter baseline in x-axis direction on observer spacecraft, which can not only acquire binocular video sequences during simulation, but also achieve aligned depth maps and point clouds from camera buffer by the API of physics engine. The camera matrix is very vital for our 3D tracking framework in which point clouds should be projected to image plane, however, V-rep merely provides perspective angle $(\alpha_x, \alpha_y)$ and resolution $\left(W, H\right)$. To this end, we compute camera intrinsic matrix $M_i$ following Eq. \ref{eq_1} (more derived details are given in Appendix \ref{appendix1}):

\begin{equation}
    M_{i}=
    \begin{bmatrix}
       \frac{W}{2 \tan(\alpha_x/2)} & 0 & \frac{W}{2} \\
       0 &   \frac{H}{ 2 \tan(\alpha_y/2)} & \frac{H}{2}\\
       0 & 0 & 1
    \end{bmatrix} 
    \label{eq_1}
 \end{equation}
    
At final, we collect 360 sequences for training and 135 sequences for testing set. Each sequence has 300 frames binocular images, depth maps, and point clouds. Following the protocol proposed in \cite{huangGOT10kLargeHighDiversity2019}, there is no overlap on fine-grained categories between training and testing set. The screenshots of 3D asteroid tracking testing set with 2D and 3D annotations are shown in Fig. \ref{fig3}.

\begin{figure}[t]
    \centering
        \subfloat[Size classes]{\includegraphics[width=0.23\textwidth]{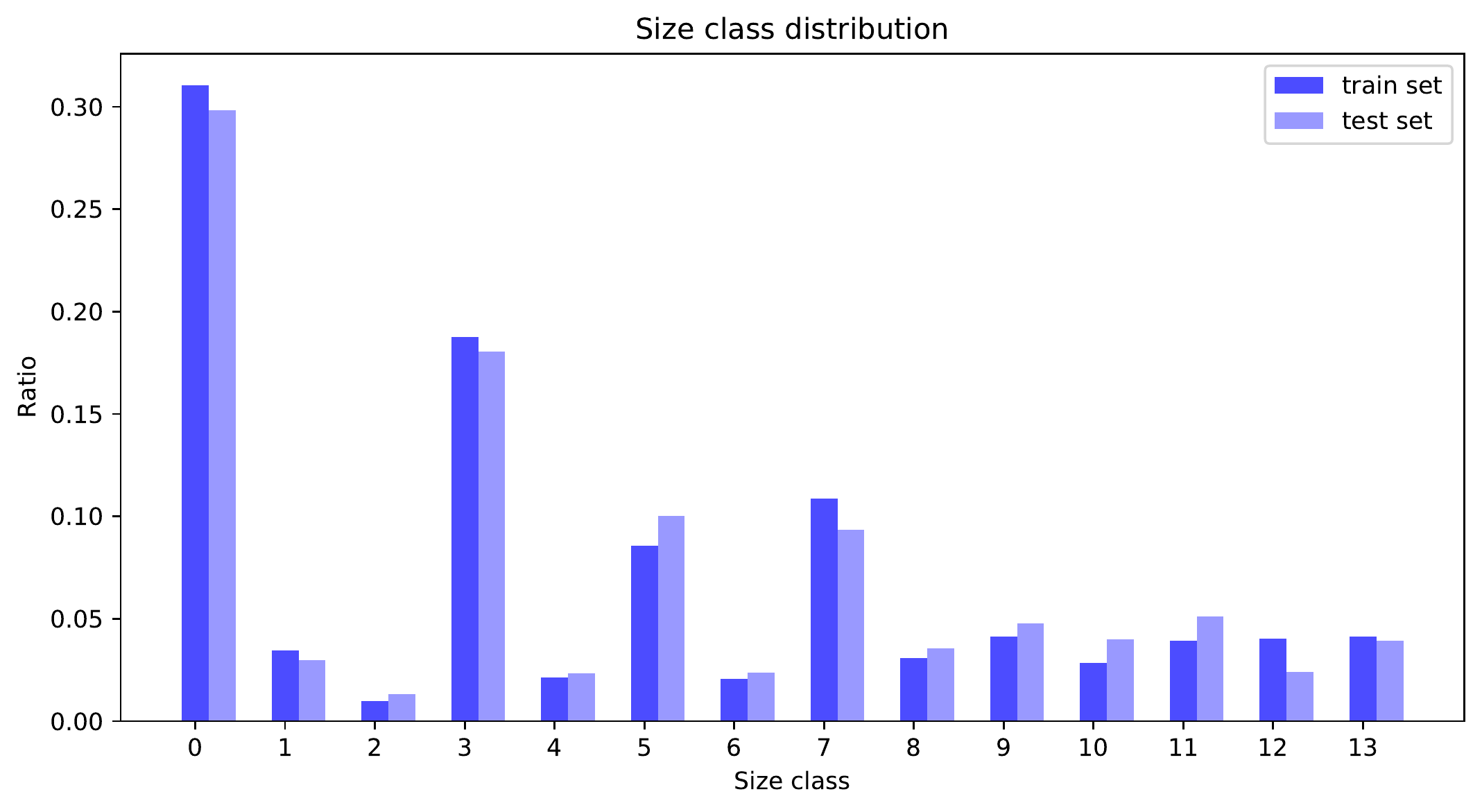} \label{fig4_a}} \hfil
        \subfloat[Object volumes]{\includegraphics[width=0.23\textwidth]{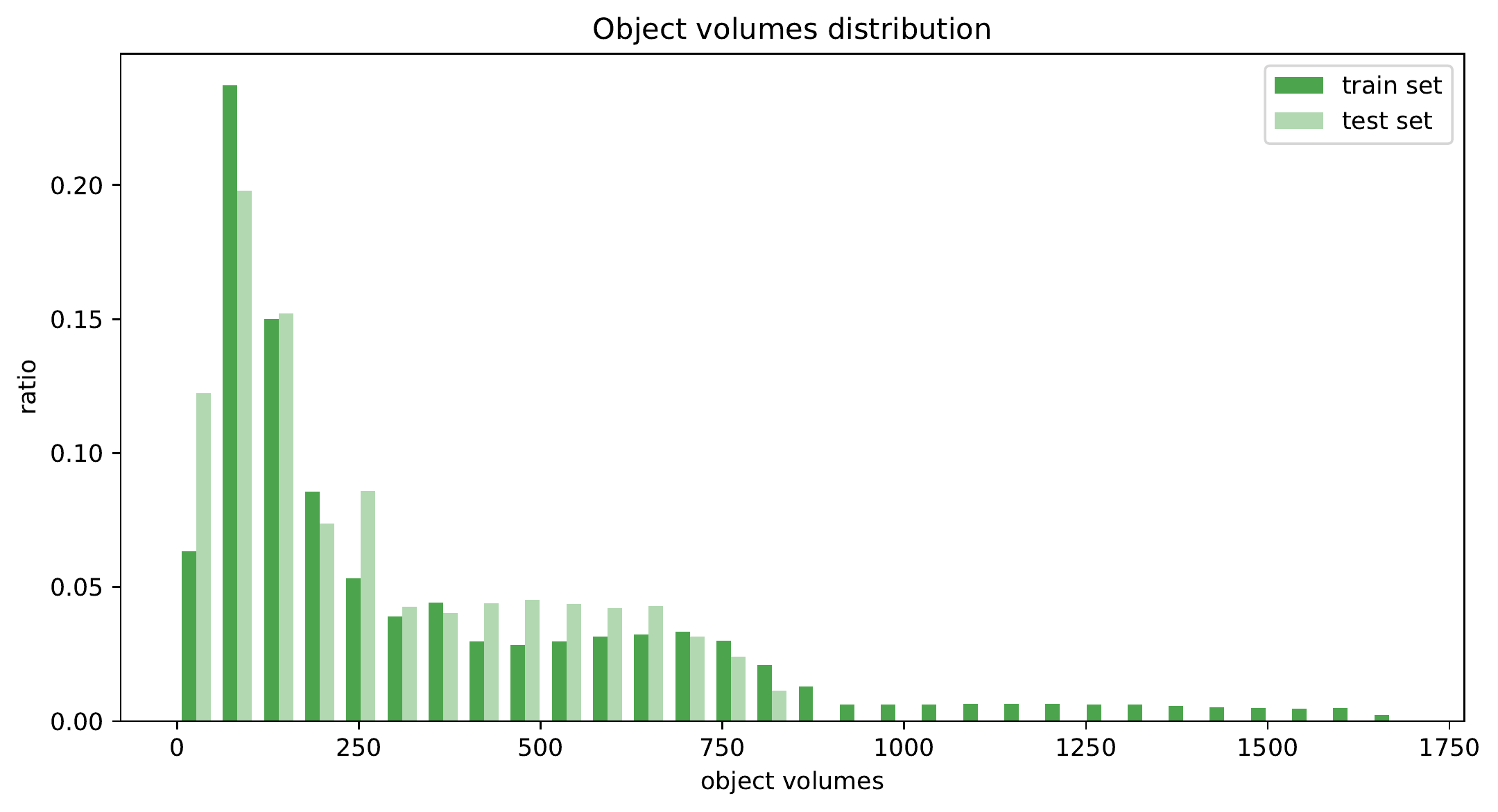} \label{fig4_b}} \vfill
        \subfloat[Motion speeds]{\includegraphics[width=0.23\textwidth]{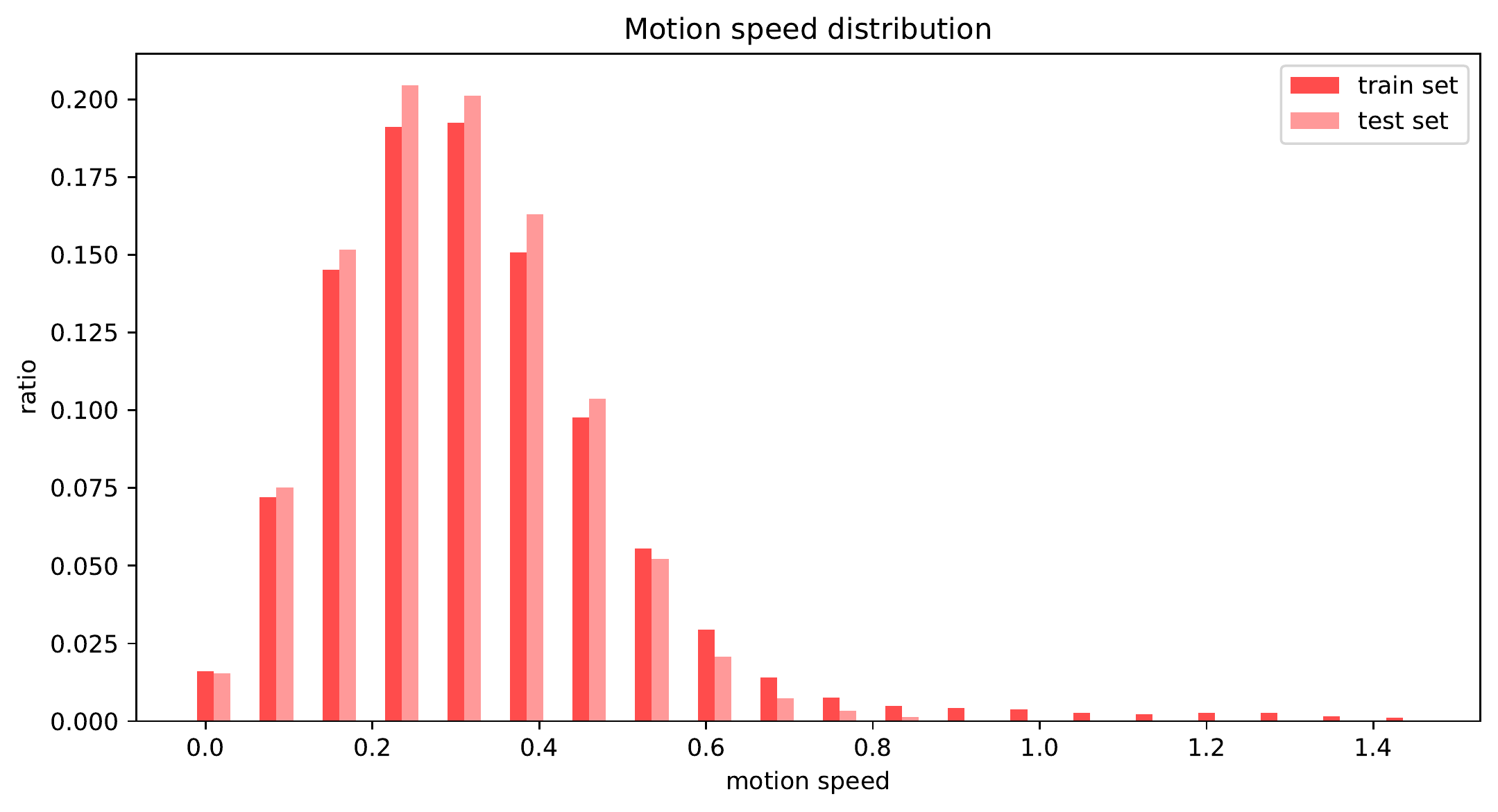} \label{fig4_c}} \hfil
        \subfloat[Illumination]{\includegraphics[width=0.23\textwidth]{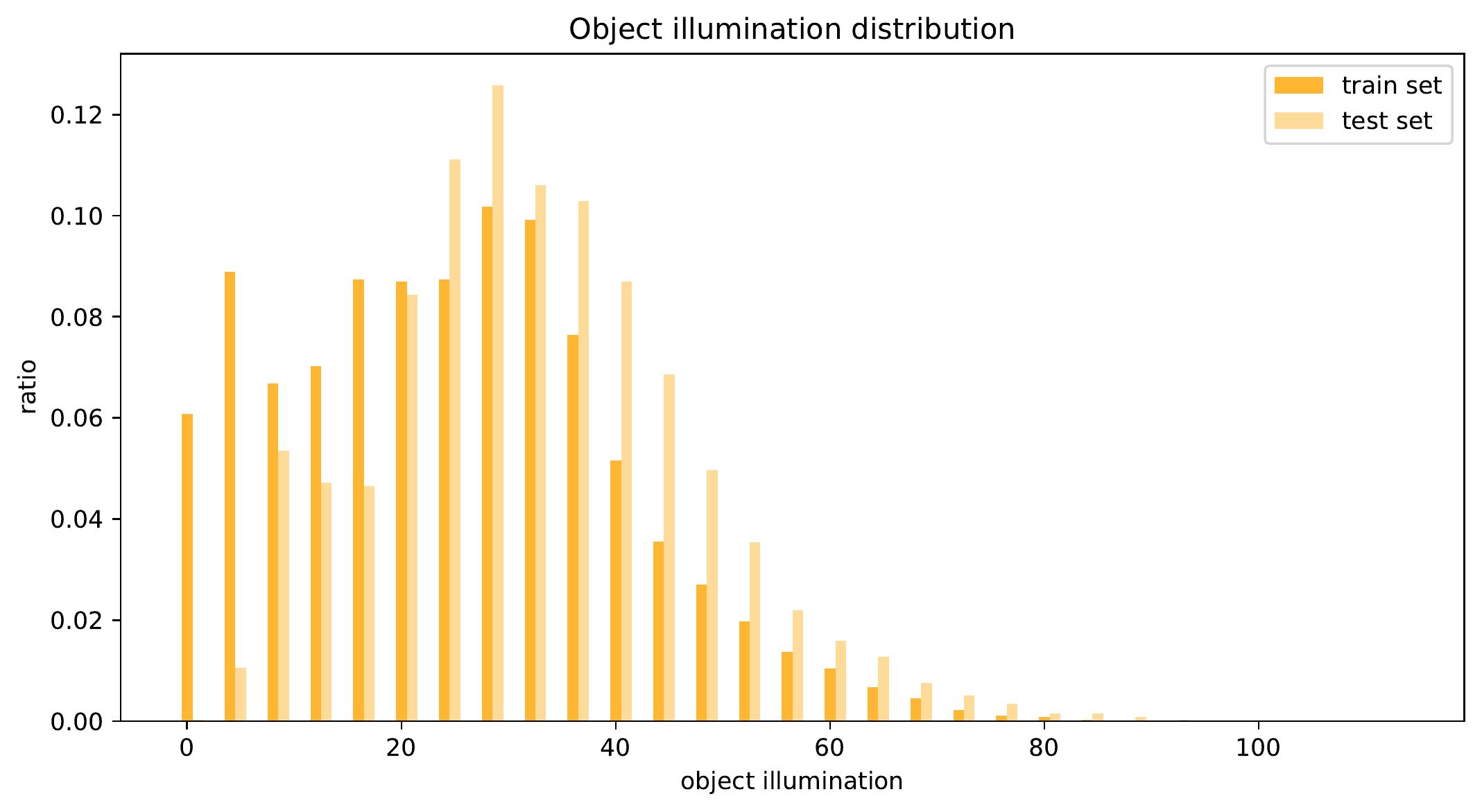} \label{fig4_d}} 
        \caption{The statistics of 3D asteroid tracking dataset.}
        \label{fig4}
    \centering
\end{figure}

\begin{figure*}[ht]
    \centering
	\includegraphics[width=0.8\textwidth]{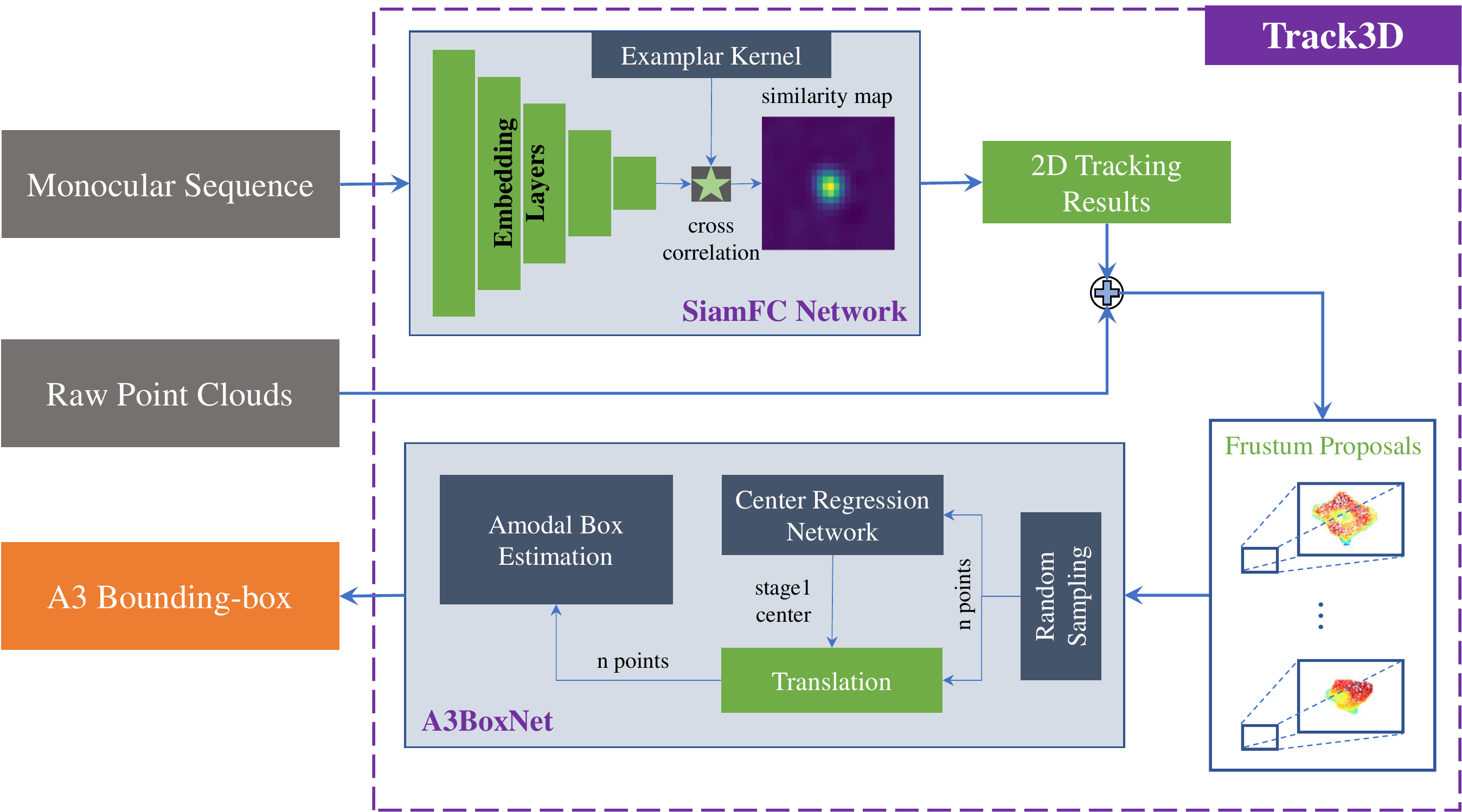}
    \caption{The detailed framework diagram of Track3D}
    \label{fig5}
    \centering
\end{figure*}

\subsection{Dataset Statistics}
In actual, our framework predicts the object size class and normalized size residuals, rather outputs an axis-aligned 3D bounding-box directly. Therefore, size classes should be predefined and the number of samples in each category has better to be balanced. To this end, we make a statistic analysis for size category of axis-aligned 3D bounding-boxes in 3D asteroid tracking dataset and define 14 default size categories summarized in Table \ref{table1}. The final distribution of size classes can be seen in Fig. \ref{fig4_a}. 

Meanwhile, the distribution of object volumes and motion speeds are shown in Fig. \ref{fig4_b} and \ref{fig4_c}, respectively. Because of random asteroid rotation and manual shape modification, the volume of samples is widely distributed (16 $m^3$ to 1600 $m^3$). It is worthwhile noting that object volume metioned here is the volume of axis-aligned 3D bounding-box, instead the real one. In addition, as we mentioned above that all the 3D asteroid models are controlled by scripts to carry out random 6-DoF motion, Fig. \ref{fig4_c} clearly demonstrates that the random translation of object obeys normal distribution. 

As paper \cite{zhou2DVisionbasedTracking2021} had proven that all the generic trakcers are vulnerable to illumination variation, we also consider this factor into our dataset. Furthermore, we have limited the illumination of all samples at low value (see in Fig. \ref{fig4_d}), which conforms to the reality of space environment.

\subsection{Evaluation Metrics}
The main metric utilized to analyse 2D tracking performance is average overlap, $AO$, which is also widely applied in visual tracking benchmarks like OTB\cite{wuObjectTrackingBenchmark2015}, VOT\cite{kristanSeventhVisualObject2019}, and GOT-10k \cite{huangGOT10kLargeHighDiversity2019}. In this paper, we also introduce this metric to evaluate 2D monocular tracker and 3D tracking results in bird's eye view (i.e. $AO^{2d}$ and $AO^{bev}$). Furthermore, We extend average overlap measurement to evaluate 3D tracking accuracy, named as $AO^{3d}$:
\begin{align}
    AO^{3d} &= \frac{1}{M} \sum_{s=1}^{M} \left( \frac{1}{N_{s}} \sum_{t=1}^{N_{s}} \Omega_{t}^{3d} \right);\\ 
    \nonumber \Omega_{t}^{3d} &= \frac{\tilde{A}_{t}^{3d} \cap A_{t}^{3d}}{\tilde{A}_{t}^{3d} \cup A_{t}^{3d}};
\end{align}
in which, $M$ is the sequences number in 3D asteroid tracking test set, $N_{s}$ denotes the length of \emph{s}-th sequence, and $\Omega^{3d}_{t}$ denotes the region overlap between axis-aligned 3D annotation  $\tilde{A}_{t}^{3d}$ and 3D tracking result $A_{t}^{3d}$ at \emph{t}-th frame. It is worthwhile noting that tracking restart mechanism is introduced to 3D tracking evaluation. That is 3D tracker is allowed to reinitialize, once the Intersection-over-Union (IoU) of 3D tracking result $\Omega^{3d}_{t}$ is zero.

Besides average overlap metric, we also adopt success rate, $SR$, as our indicator, which denotes the percentage of successfully tracked frames where the overlap exceeds a threshold. In this work, we take 0.5 as threshold to measure the success rate of one tracker. However, $SR$ at a specific threshold is not representative, we further introduce success plot presented by \cite{wuObjectTrackingBenchmark2015} into our evaluation tool.

Another intuitive technique index of tracker is location precision. Therefore, we propose 3D average center error metric ($ACE^{3d}$) that measures the mean distance between all the predicted trajectories and ground-truth trajectories:
\begin{align}
    ACE^{3d}  &= \frac{1}{M} \sum_{s=1}^{M} \Delta \left( \Gamma_s, \tilde{\Gamma}_s \right)
\end{align}
where $\Gamma_s = \left\{ p_t | t=1, \ldots, N_s \right\}$ and $\tilde{\Gamma}_s = \left\{ \tilde{p}_t | t=1, \ldots, N_s \right\}$ are the predictions and ground-truths of \emph{s}-th sequence, which consist of a series of object center in 3D space. $\Delta \left( \Gamma_s, \tilde{\Gamma}_s \right)$ is formulated as:
\begin{equation}
    \Delta\left(\Gamma_{s}, \tilde{\Gamma}_{s}\right)=\frac{1}{N_{s}} \sum_{t=1}^{N_{s}}\left\|p_{t}-\tilde{p}_{t}\right\|_{2}
\end{equation}
in which, $\left\| \cdot \right\|_{2}$ denotes Euclidean distance.

\begin{figure*}[t]
    \centering
	\subfloat[Center regression network]{\includegraphics[width=0.65 \textwidth]{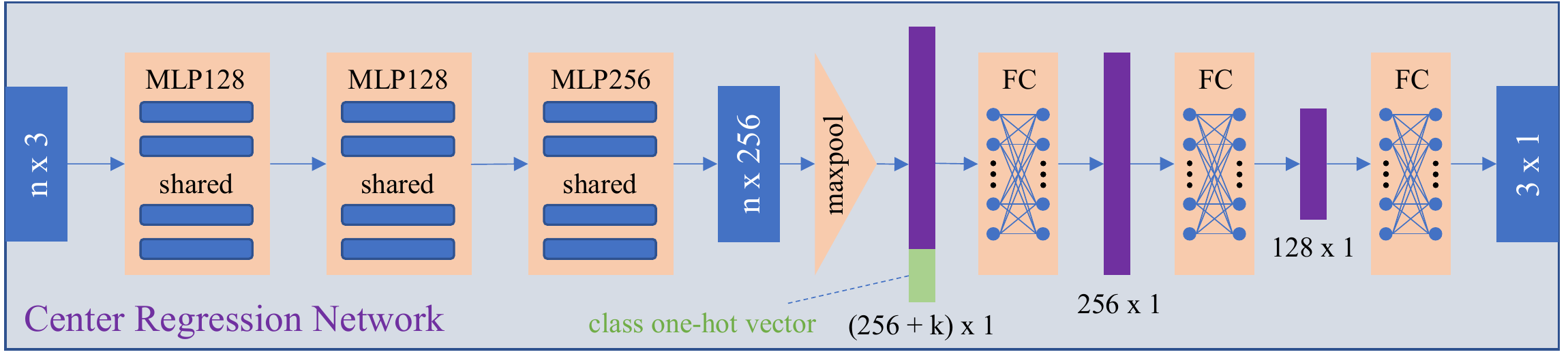} \label{fig6_a}} \hfil
    \subfloat[Amodal box estimation network]{\includegraphics[width=0.7 \textwidth]{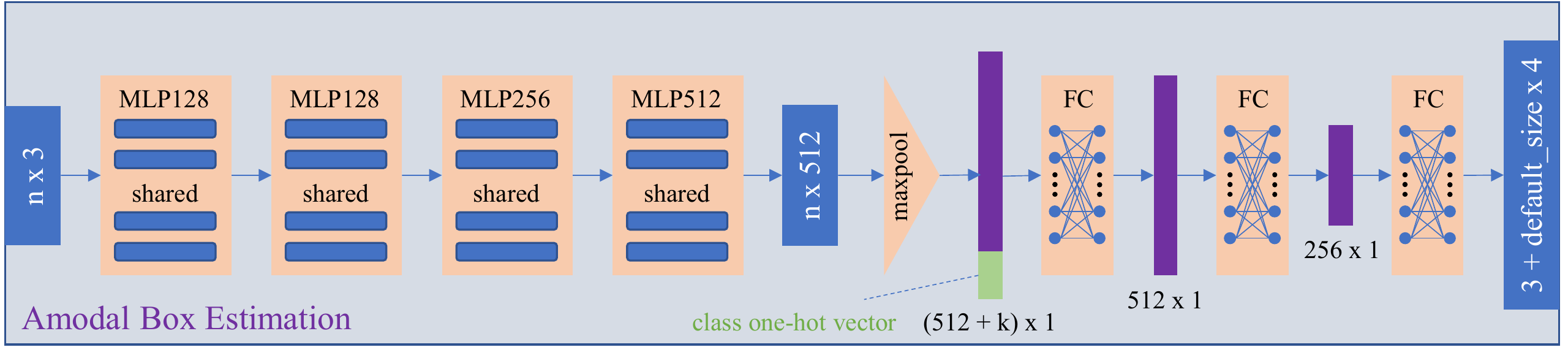} \label{fig6_b}}
    \caption{Two main module of A3BoxNet. (a) predicts coarse object center, (b) is to estimate object center residuals, object size category and normalized size residuals.}
    \label{fig6}
    \centering
\end{figure*}

\section{3D Tracking Framework \label{section4}}
Inspired by the 2D-drieven 3D perception, we propose a deep-learning based 3D tracking framework shown in Fig. \ref{fig5}, which involves 2D monocular tracker, SiamFC \cite{bertinettoFullyconvolutionalSiameseNetworks2016} and a novel light-weight amodal axis-aligned bounding-box network, A3BoxNet. Although binocular images and depth maps are also provided in our 3D asteroid tracking dataset, it is worthwhile noting that Track3D only utilizes monocular video sequence and corresponding point clouds, which reduces the complexity of 3D tracking framework and more conforms to real applications in aerospace. In addition, we will introduce a simple but effective 2D-3D tracking fusion strategy in the following subsections. 

\subsection{SiamFC}
The SiamFC algorithm utilizes full convolutional layers as an embedding layer, denoted as embedding function $\varphi$. The exemplar image $z$ and multi-scale search images $X=\left\{x_i | i = 1, 2, ..., S \right\}$ are mapped to a high-dimensional feature space with $\varphi$, which can be trained by a general large-scale dataset. And then similarity score maps are generated by cross-correlation between exemplar kernal and the tensors of multi-scale search images, where the 2D tracking result can be reached after post-processing. 

For this algorithm, the key is to learn discriminative embedding function $\varphi$. In this work, we train the backbone of SiamFC from scratch on ILSVRC-VID \cite{russakovskyImageNetLargeScale2015} with 50 epochs and further fine-tune it on our 3D asteroid tracking training set. Benefitting from the strong generalization ability of Siamese network, SiamFC can achieve a decent 2D tracking performance. Although there are multifarious deep-learning based 2D monocular trackers, like SiamRPN++ \cite{liSiamRPNEvolutionSiamese2018}, Ocean \cite{zhangOceanObjectawareAnchorfree2020}, STARK \cite{yanLearningSpatioTemporalTransformer2021}, TransT \cite{chenTransformerTracking2021}, which greatly outperform SiamFC in tracking accuracy, we think weak dependence on high-performance 2D monocular tracker can effectively guarantee the generalization ability of our framework and improve its running speed. 

Once tracking result $A_{t} $ is acquired from 2D monocular tracker, it can be used for frustum proposal extraction from raw points set $P_{\text{raw}} = \left\{ (x_i, y_i, z_i) \in \mathbb{R} ^3 | i=1, 2, ..., r \right\}$, as shown in Fig. \ref{fig5}. We first crop out $m$ points from raw point cloud $P_{\text{crop}} = \left\{ (x_i, y_i, z_i) |  1 \leqslant z_i \leqslant 45, i=1, 2, ..., r \right\}$, and then compute the projected point cloud in image plane $P_{\text{img}}=\left\{ (u_i, v_i) | i=1, 2, ..., m \right\}$ with camera matrix $\mathbf{M} \in \mathbb{R}^{3 \times 4}$. Therefore, we can achieve the frustum proposal $P_{\text{frustum}} =  \left\{ (x_i, y_i, z_i) | i=1, 2, ..., n \right\} $ by the indices of points which belong to $A_t$ region in $P_{\text{img}}$. 

\begin{figure}[t]
    \centering
        \includegraphics[width=0.4 \textwidth]{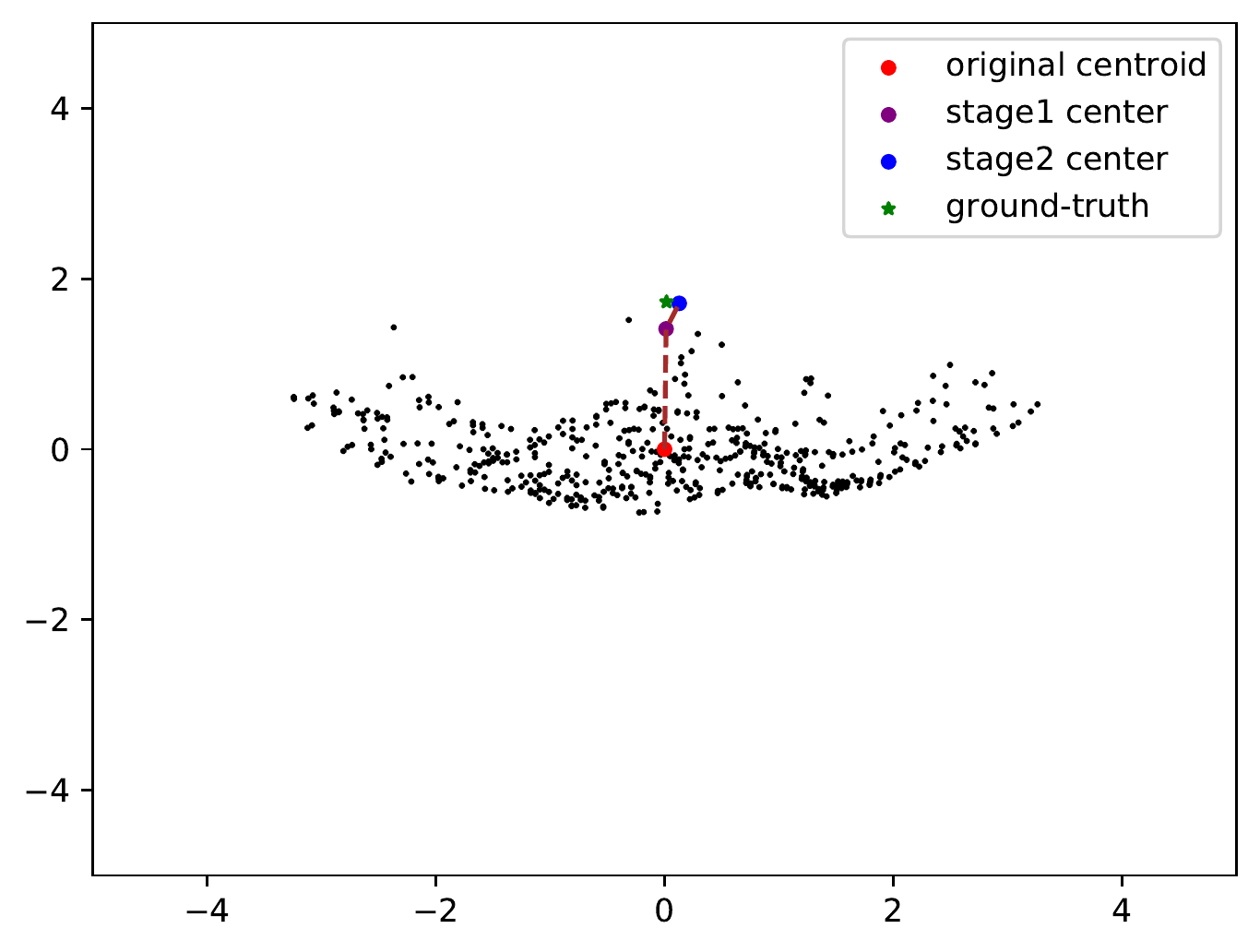}
        \caption{Two-stage object center prediction in bird's eye view. We first estimate the stage-1 center of input points with center regression network. And then center residuals are predicted by amodal box estimation network, also named as stage-2 center.}
        \label{fig7}
    \centering
\end{figure}

\begin{figure}[t]
    \begin{center}
        \includegraphics[width=0.4 \textwidth]{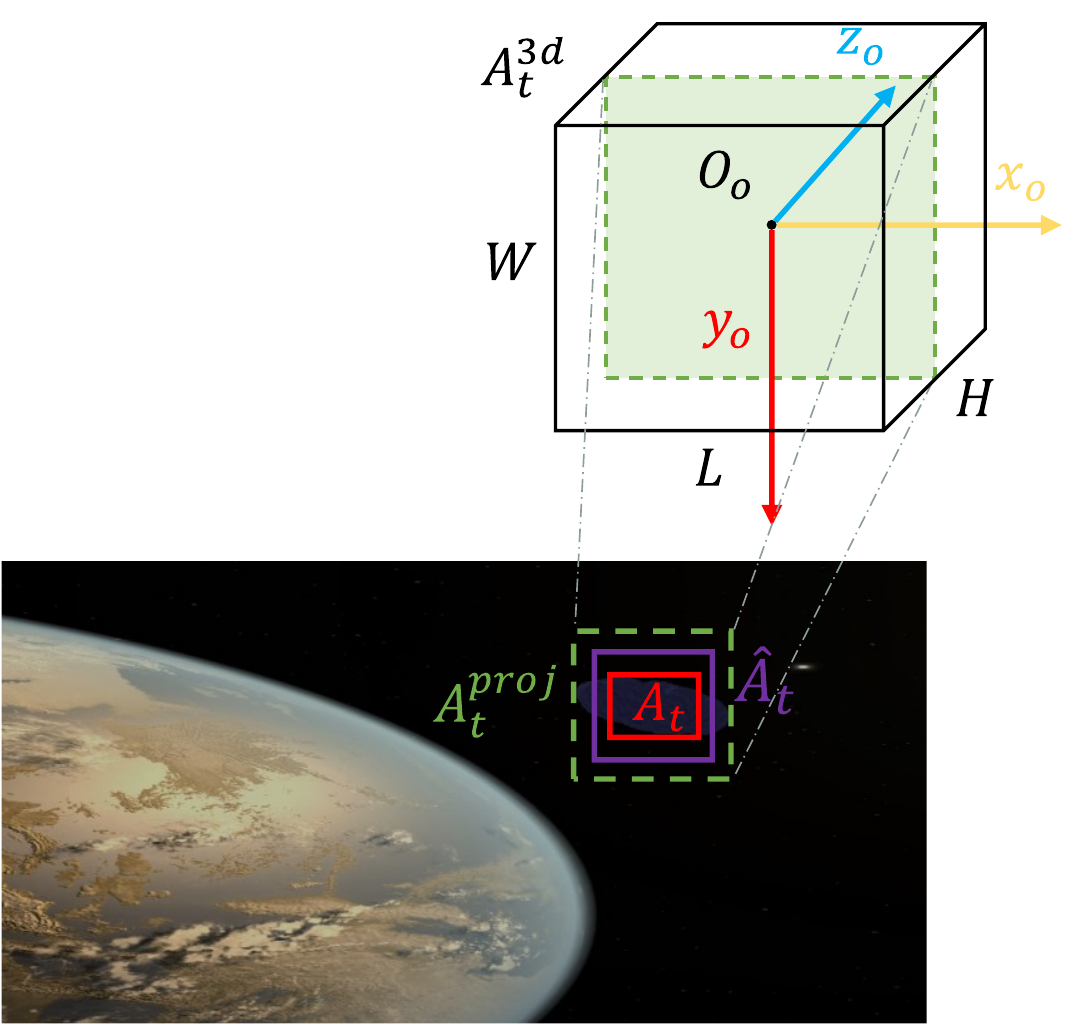}
        \caption{The schematic diagram of 2D-3D tracking fusion strategy. $A^{proj}_t$ is the projection of center cross-section plane of $A^{3d}_t$ in image plane and $A_t$ is the result of 2D monocular tracker at \emph{t}-th timestep. We compute fused tracking result $\hat{A}_t$ by the weighted average between $A^{proj}_t$ and $A_t$.}
        \label{fig8}
    \end{center}
\end{figure}

\begin{table*}[t]
    \centering
    \caption{The evaluation results on 3D asteroid tracking dataset. First two rows are the performances of two main modules in Track3D, and the 3rd row is a simple 3D tracking baseline as comparison. All the 2D tracking metrics of 3D tracker are computed by using 2D-3D fusion strategy which makes great improvement on 2D tracking performance. The top-2 evaluation results for each performance indicator are highlighted in \textcolor{green}{green} and \textcolor{blue}{blue}.}
    \label{table2}
    \begin{tabular}{ccccccc}
        \toprule
         \multirow{2}{*}{name} & \multirow{2}{*}{$AO^{2d}$} & $ACE^{2d}$ & \multirow{2}{*}{$AO^{bev}$}  & \multirow{2}{*}{$AO^{3d}$}  & $ACE^{3d}$ & \multirow{2}{*}{FPS}\\
         & & (pixel) & & & (meter) & \\
        \midrule
        SiamFC & 0.513 & 29.986 & - & - & - & 101.3\\
        A3BoxNet & - & - & \textcolor{green}{0.798} & \textcolor{green}{0.721} & \textcolor{green}{0.345} & \textcolor{green}{281.5} \\ \midrule
        3D Tracking Baseline & \textcolor{green}{0.568} & \textcolor{green}{23.767} & 0.309 & 0.165 & 0.876 & \textcolor{blue}{103.1} \\
        Track3D & \textcolor{blue}{0.541} & \textcolor{blue}{26.741} & \textcolor{blue}{0.756} & \textcolor{blue}{0.669} & \textcolor{blue}{0.570} & 77.0 \\
        \bottomrule  
    \end{tabular}
    \centering
\end{table*}

\subsection{A3BoxNet}
In this work, we assume that there are no other objects in the range of LiDAR perception, that is, all of the points in frustum proposal belong to tracked target, which conforms to practical scenario in space. Therefore, we propose a novel light-weight amodal axis-aligned bounding-box network, A3BoxNet, which makes directly prediction on frumstum proposal and no point segmentation is considered into. Fig. \ref{fig2} clearly shows that our A3BoxNet mainly consists of two modules: center regression network and amodal box estimation network. The former one is responsible for the estimation of stage-1 object center. Another is to predict object center residuals, object size category, and normalized size residuals. It is worthwhile noting that both center regression network and amodal box estimation network are only support fixed number of input points. To this end, we add random sampling at the beginning of A3BoxNet.

The architecture of center regression network and amodal box estimation network are both illustrated in Fig. \ref{fig6_a} and \ref{fig6_b}. It can be clearly seen that there is no significant difference between two networks which are both derived from PointNet \cite{qiFrustumPointNets3D2018}. In A3BoxNet, we use a number of MLP for high-level features extraction. And max-pooling layer is introduced to aggregate global features with symmetric property that is critical for unsorted point sets. The global feautres concatenated with one-hot vector of coarse-grained asteroid category can be further used for predicting stage-1 center, stage-2 center, and object size. 

We visualize the center predction progress of A3BoxNet in Fig. \ref{fig7}. The frustum proposal have been normalized by substracting the centroid of point sets firstly, which can improve the translation invariance of A3BoxNet and speed up convergence rate during training. In addtion, it shows that after two stage prediction, estimated object center is very close to the ground-truth. The predicted center is formulateds as follows:
\begin{equation}
    C_{pred} = \bar{C} + \Delta C_1 + \Delta C_2
\end{equation}
where $\bar{C}$ is the centroid of point cloud in frustum proposal, $\Delta C_1$ is the output of center regression network, $\Delta C_2$ is the center residuals predicted by amodal box estimation network. 

Except for estimating the center residuals of object, our amodal box estimation network also classifies object size to 14 predefined categories (see in Table \ref{table1}) as well as predicts normalized size residuals ($N_{size}$ scores for size classification, $N_{size} \times 3$ for size residuals regression). At final, we remap the predicted size category and normalized size residual to original scale by multiplying the largest length of enclosing bounding-box of point clouds. 

To train our A3BoxNet, we use a joint loss function $\mathcal{L}_{joint}$ to simultaneously optimize both two submodules:
\begin{equation}
    \mathcal{L}_{joint} = \mathcal{L}_{center-net} + \mathcal{L}_{box-net}
\end{equation}
in which, 
\begin{equation}
    \mathcal{L}_{box-net} = \mathcal{L}_{center\_res} +  \mathcal{L}_{size\_cls} + \mathcal{L}_{size\_res}
\end{equation}
more details about optimization function of A3BoxNet are given in Appendix \ref{appendix2}. In addition, A3BoxNet is trained on 3D asteroid tracking dataset with 25 epoches and 32 batch size. All the inputs are fixed number of point sets that randomly sampled from object point cloud.

\subsection{2D-3D Tracking Fusion Strategy}
We believe the fusion of 3D tracking and 2D monocular tracking results can make significant improvement on 2D tracking performance. To this end, we propose a simple and effective 2D-3D tracking fusion strategy (as shown in Fig. \ref{fig8}). At first, we calculate the projection of center cross-section plane of 3D bounding-box in image plane, $A^{proj}_t$. And then, 2D monocular tracking result $A_t$ is weighted with $A^{proj}_t$:
\begin{equation}
    \hat{A}^t = \lambda_1 \cdot A^{proj}_t + \lambda_2 \cdot A^t
\end{equation}
in this work, we set $\lambda_1 = 0.3$ and $\lambda_2 = 0.7$. 

\begin{figure*}[t]
    \centering
    \subfloat[3D tracking baseline]{\includegraphics[width=0.23 \textwidth, height=0.14 \textheight]{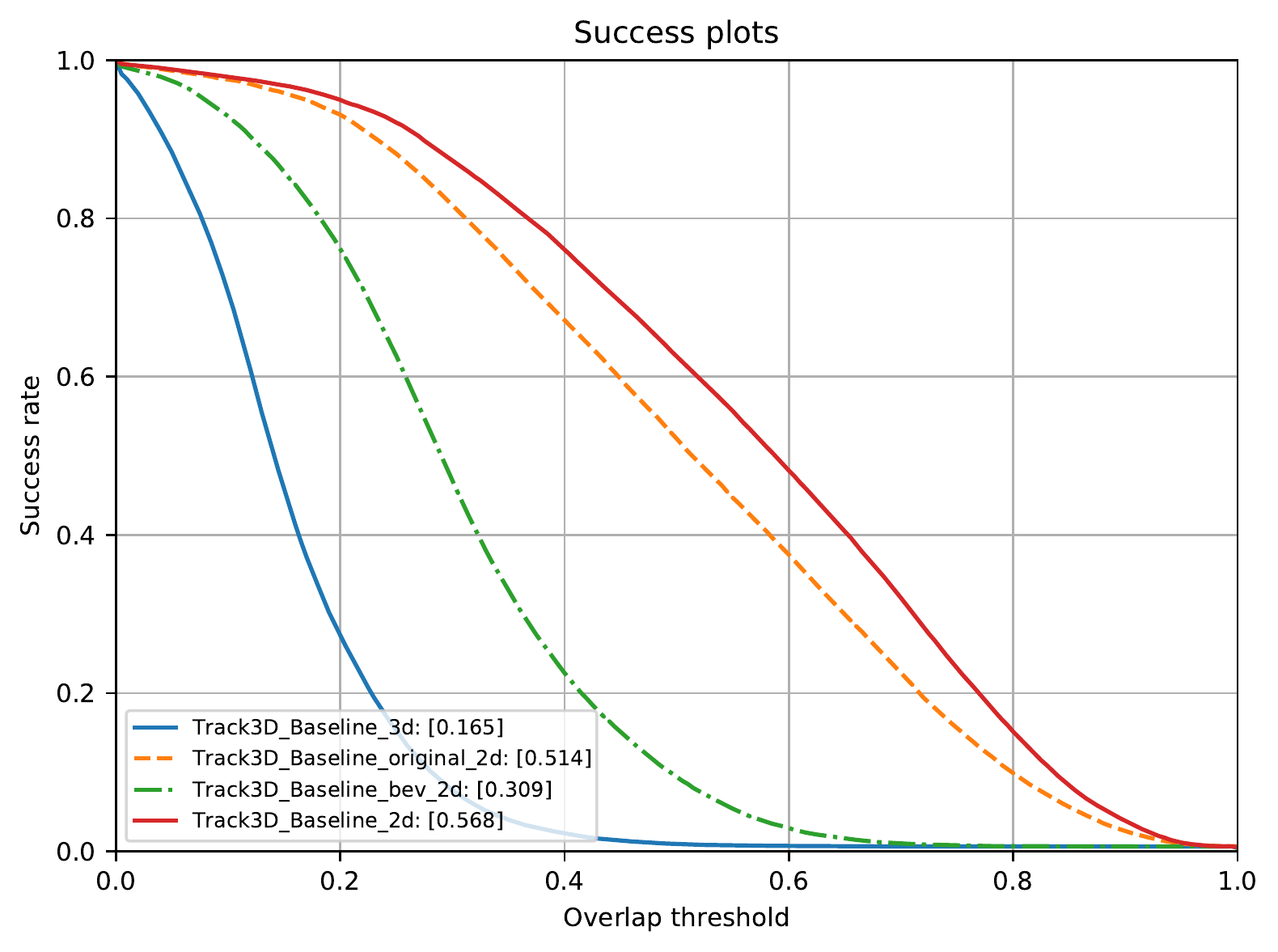}\label{fig11_a}} 
    \hfil
    \subfloat[Track3D]{\includegraphics[width=0.23 \textwidth, height=0.14 \textheight]{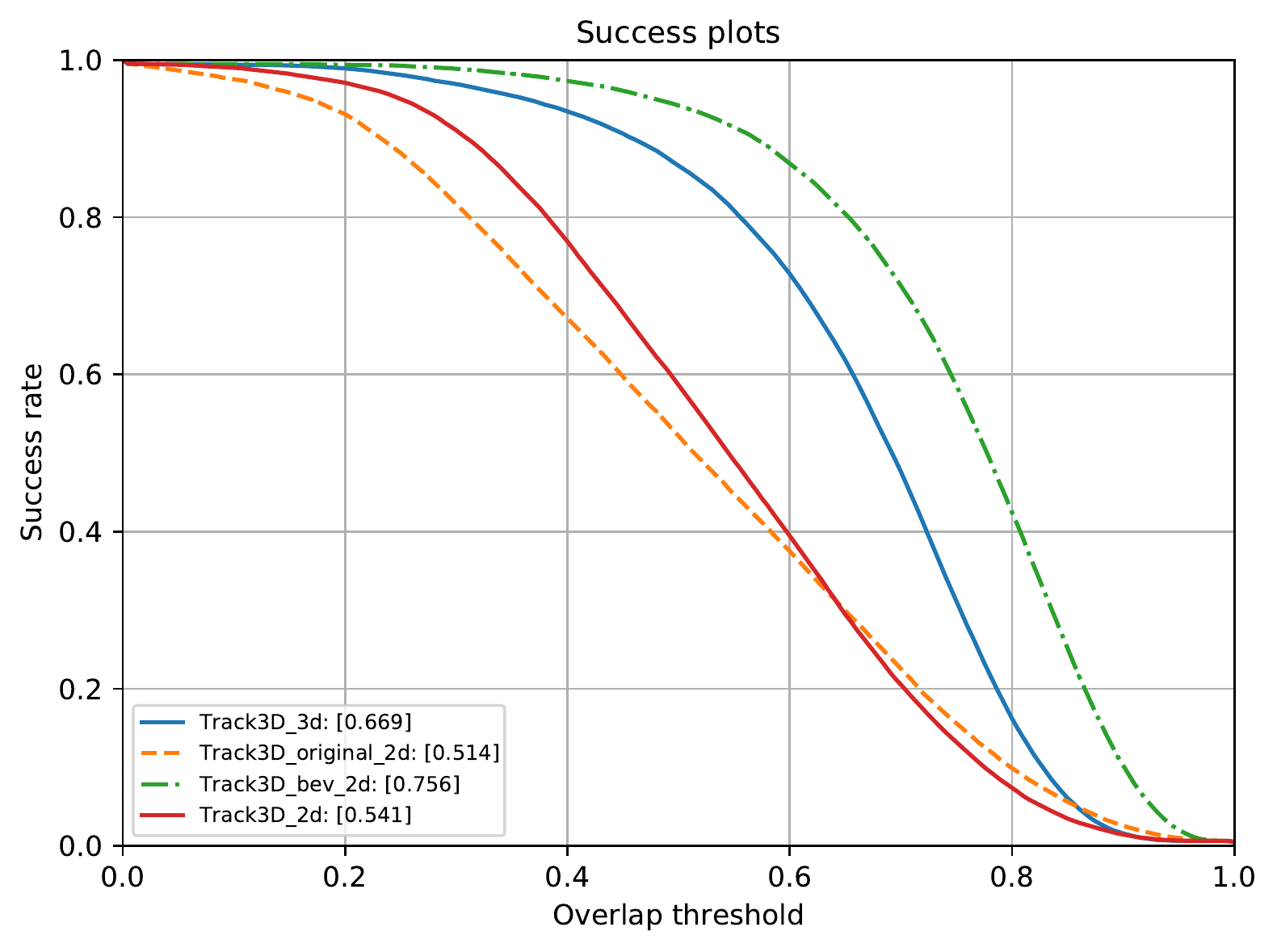}\label{fig11_b}}
    \hfil
    \subfloat[2D precision comparison]{\includegraphics[width=0.23 \textwidth, height=0.14 \textheight]{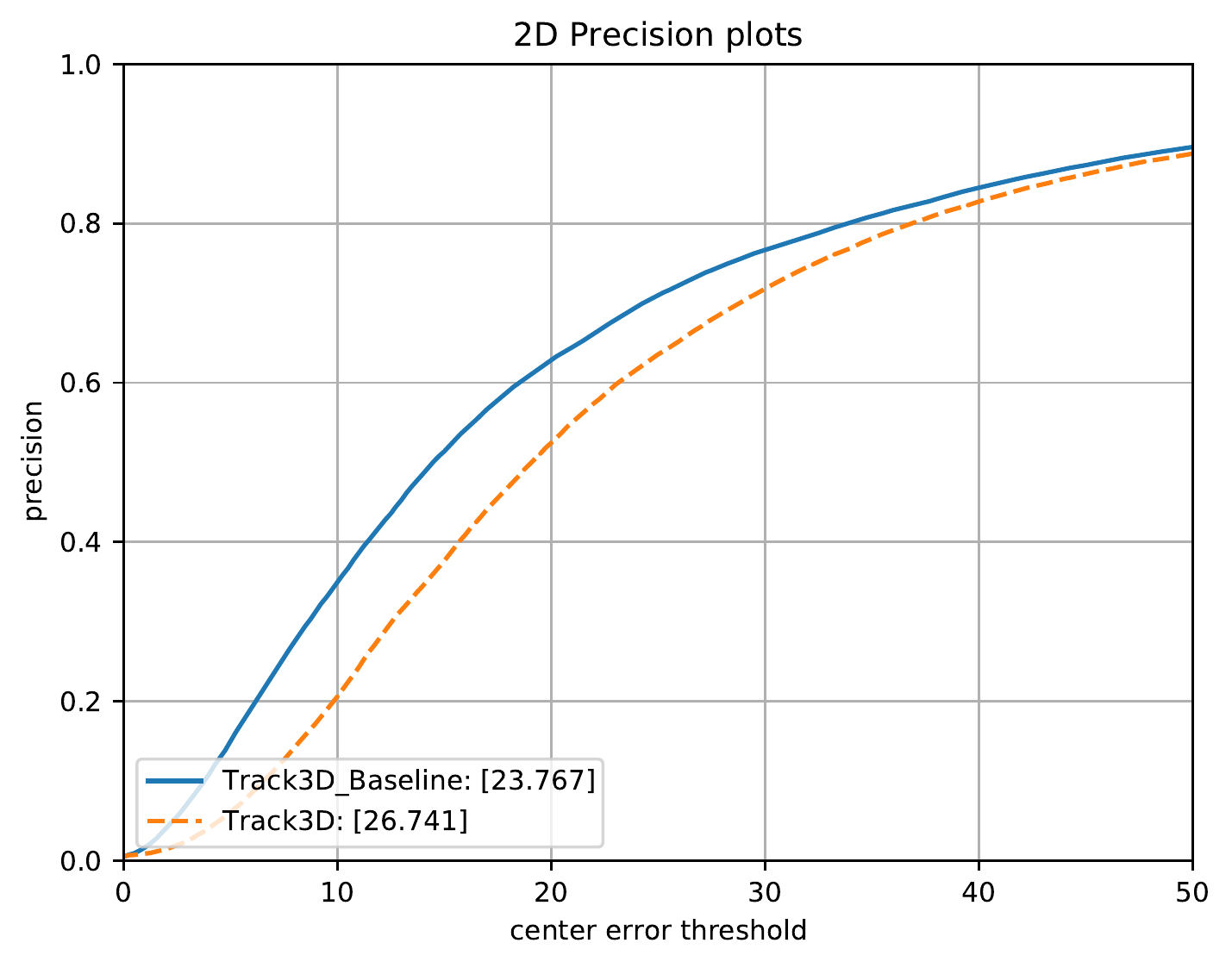}\label{fig11_c}}
    \hfil
    \subfloat[3D precision comparison]{\includegraphics[width=0.23 \textwidth, height=0.14 \textheight]{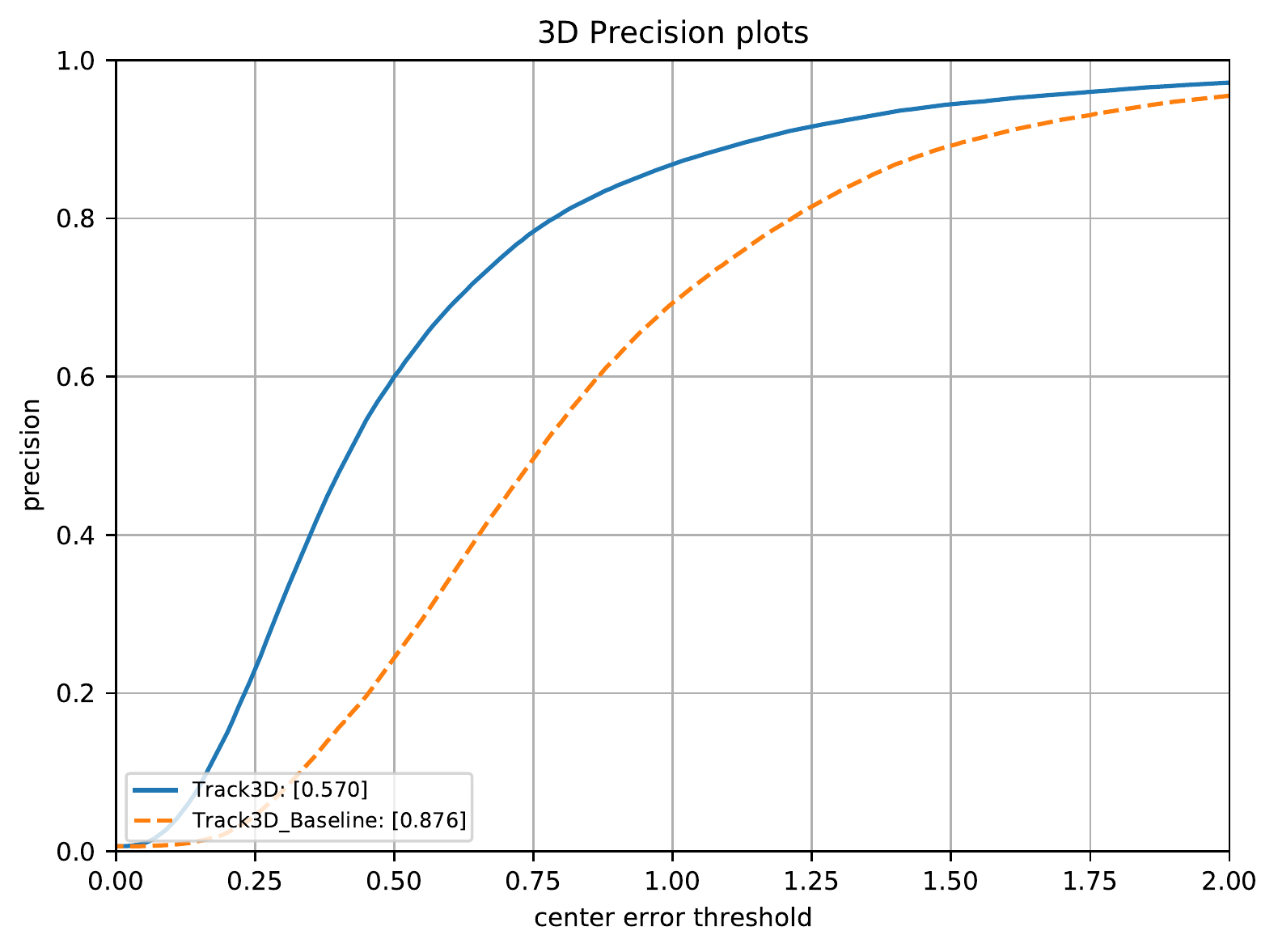}\label{fig11_d}}
    \caption{The performance comparison between 3D tracking baseline and Track3D. (a) and (b) are respectively success plots of 3D tracking baseline and Track3D with $AO^{2d}$, $AO^{bev}$, and $AO^{3d}$ metrics.}
    \label{fig10}
    \centering
\end{figure*}

\begin{figure*}[ht]
    \centering
    \begin{tabular}{cccc}
       sequence 0001 & sequence 0022 & sequence 0031 & sequence 0052
       \\
       \includegraphics[width=0.22 \textwidth]{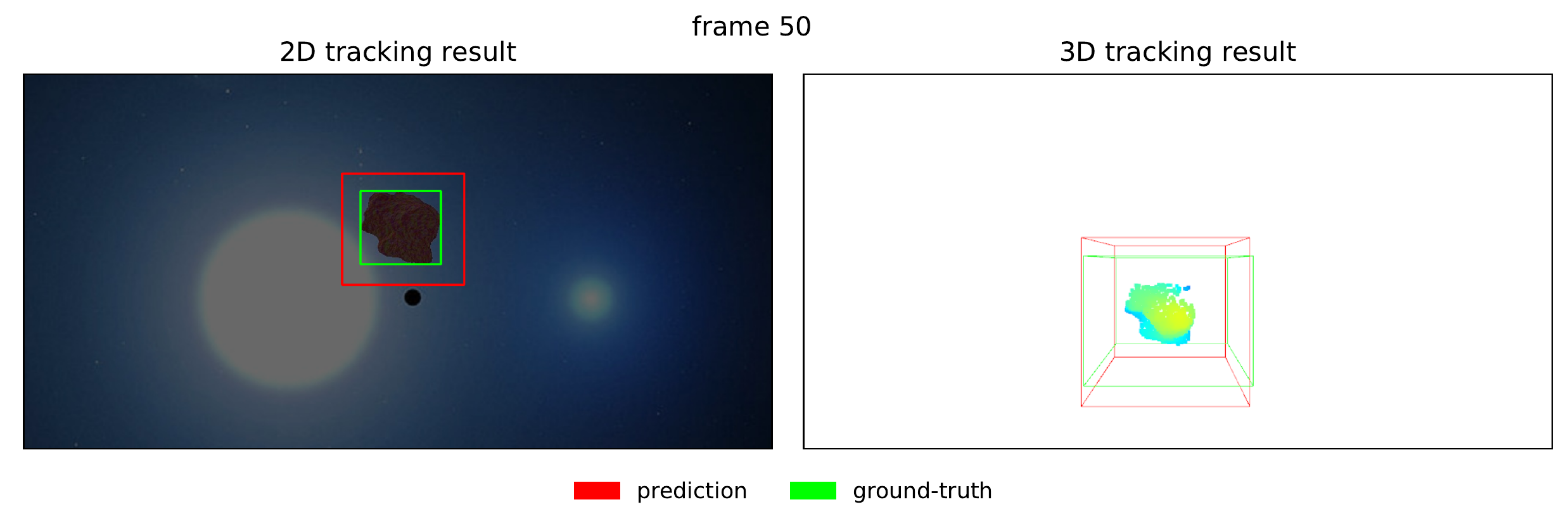} & 
       \includegraphics[width=0.22 \textwidth]{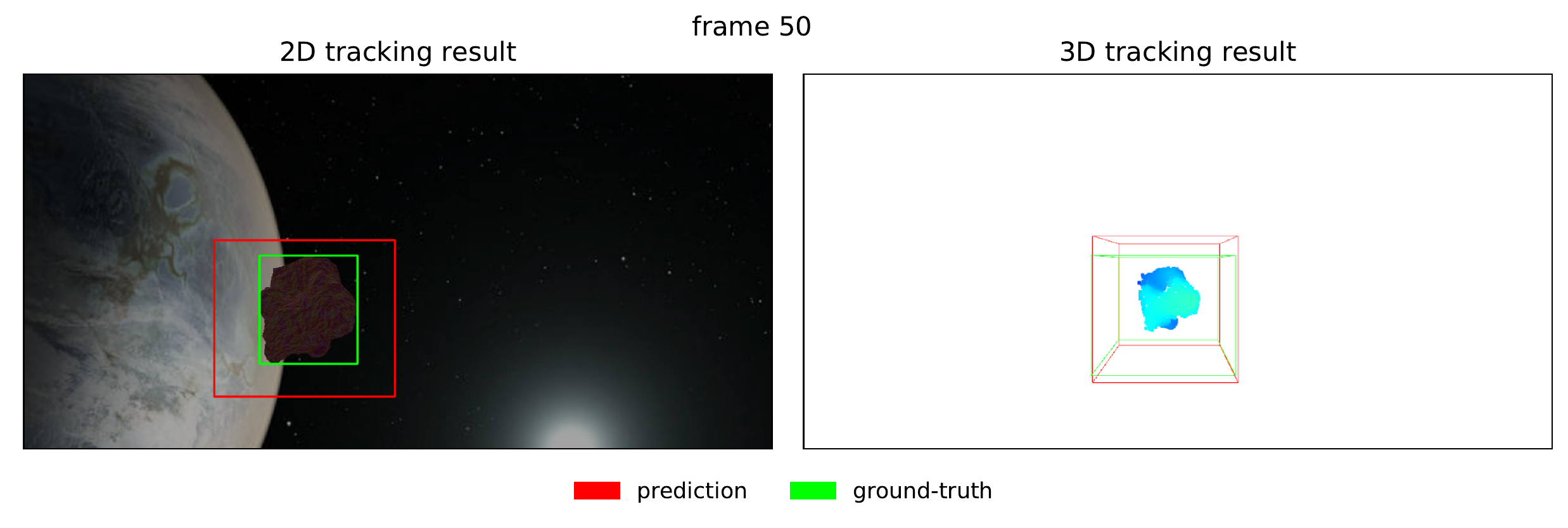} & 
       \includegraphics[width=0.22 \textwidth]{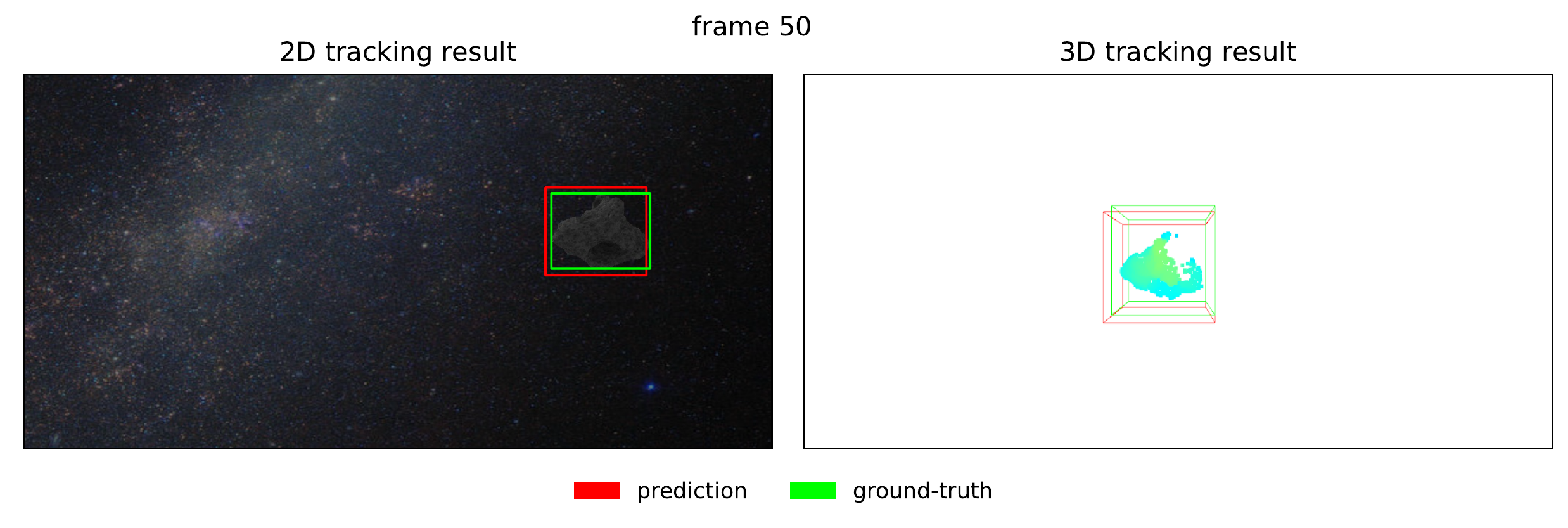} & 
       \includegraphics[width=0.22 \textwidth]{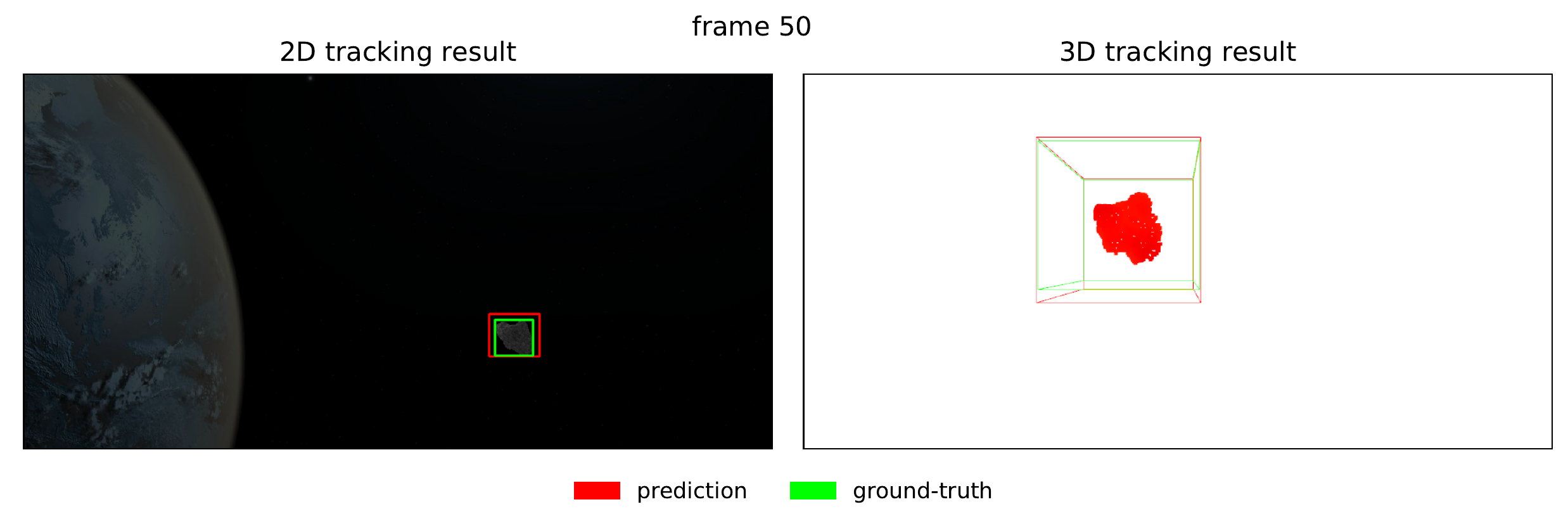} 
       \\
       \includegraphics[width=0.22 \textwidth]{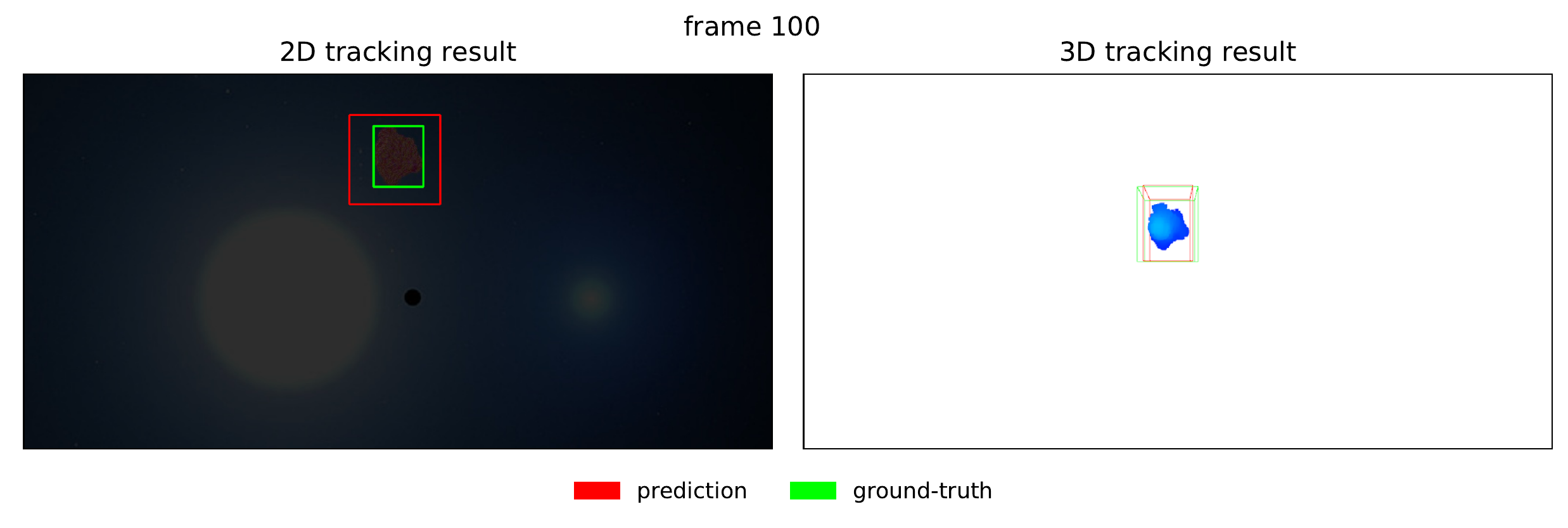} & 
       \includegraphics[width=0.22 \textwidth]{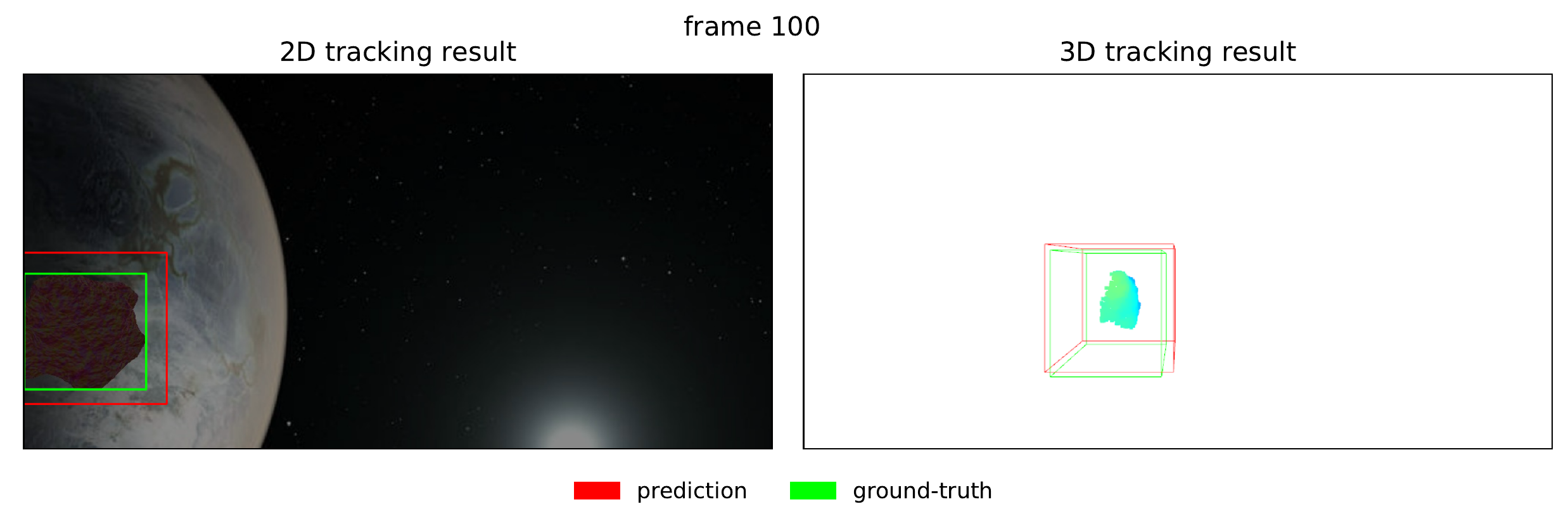} & 
       \includegraphics[width=0.22 \textwidth]{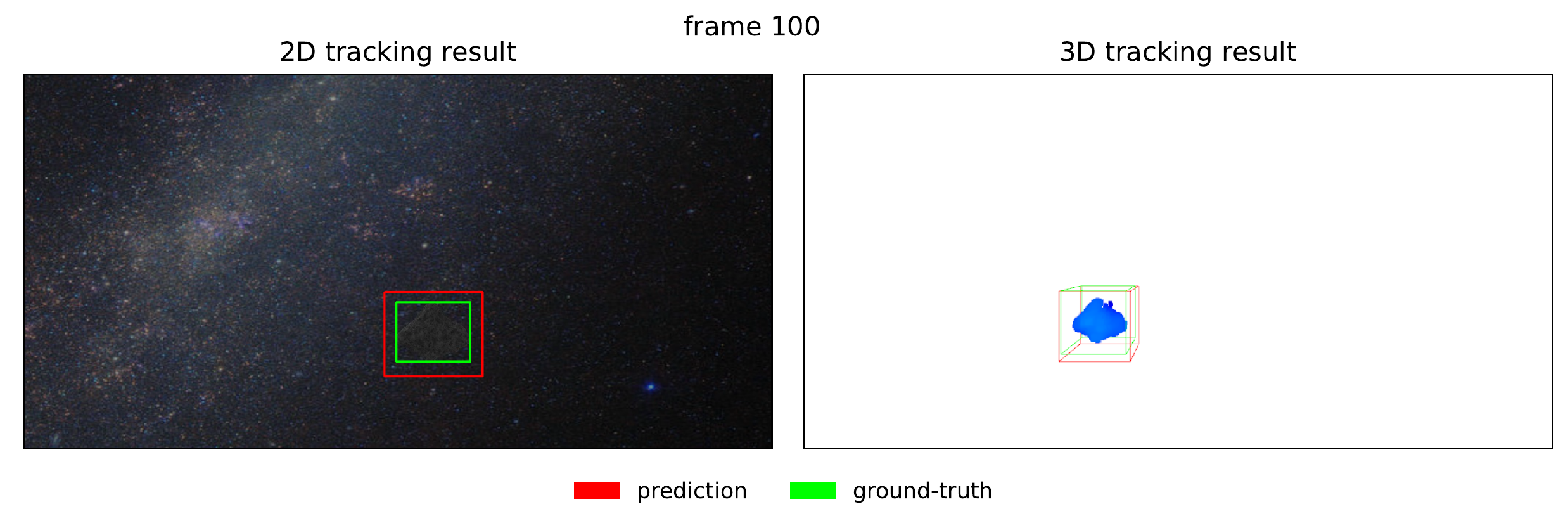} & 
       \includegraphics[width=0.22 \textwidth]{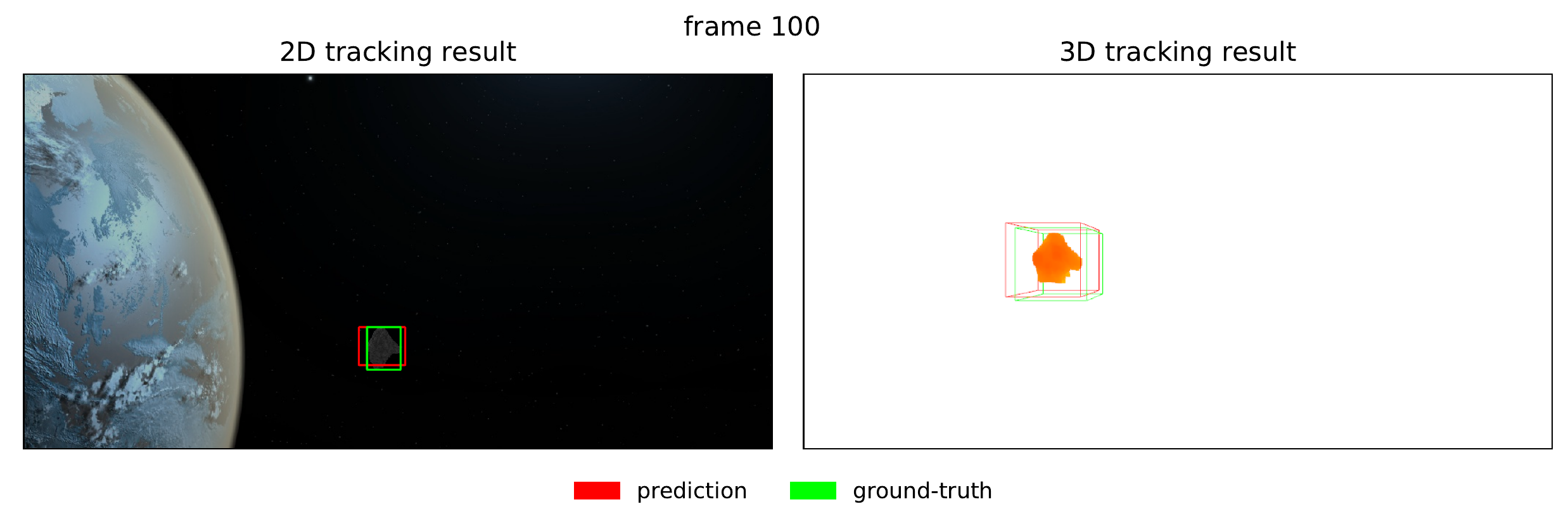} 
       \\
       \includegraphics[width=0.22 \textwidth]{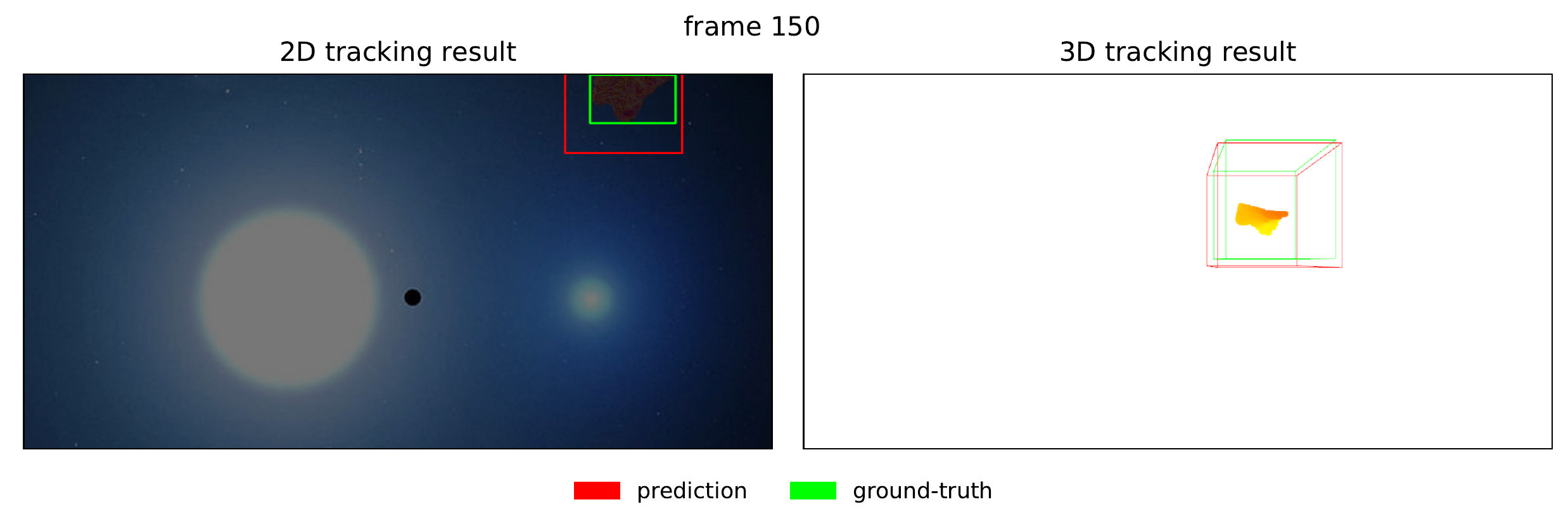} & 
       \includegraphics[width=0.22 \textwidth]{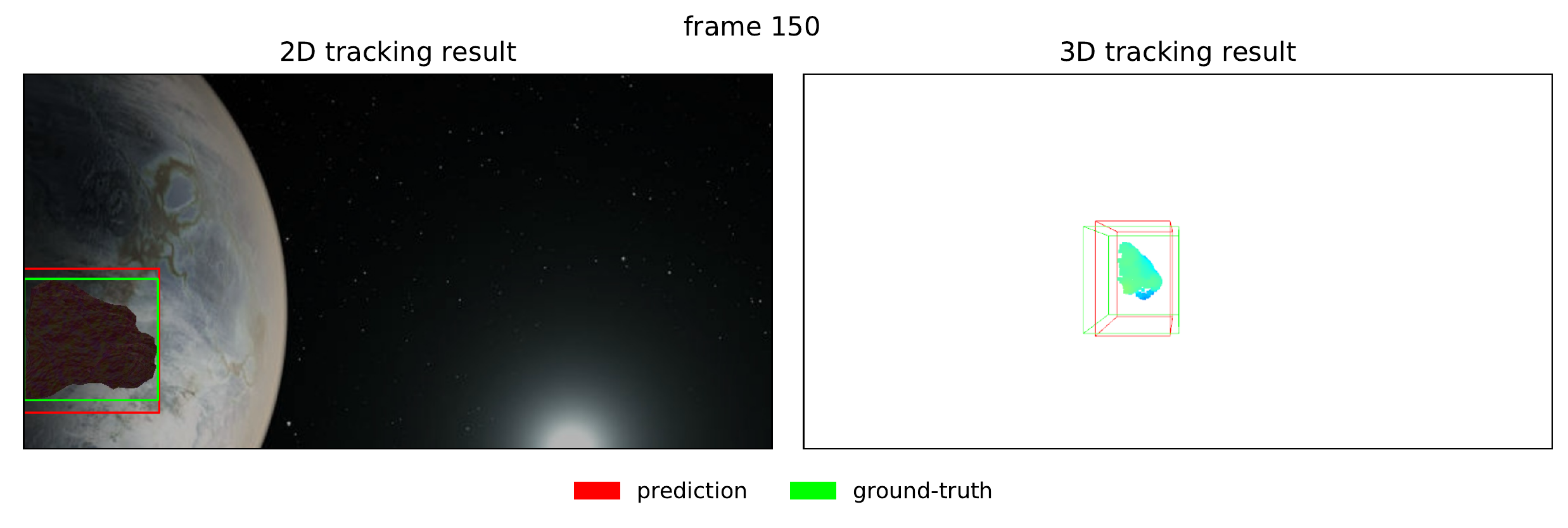} & 
       \includegraphics[width=0.22 \textwidth]{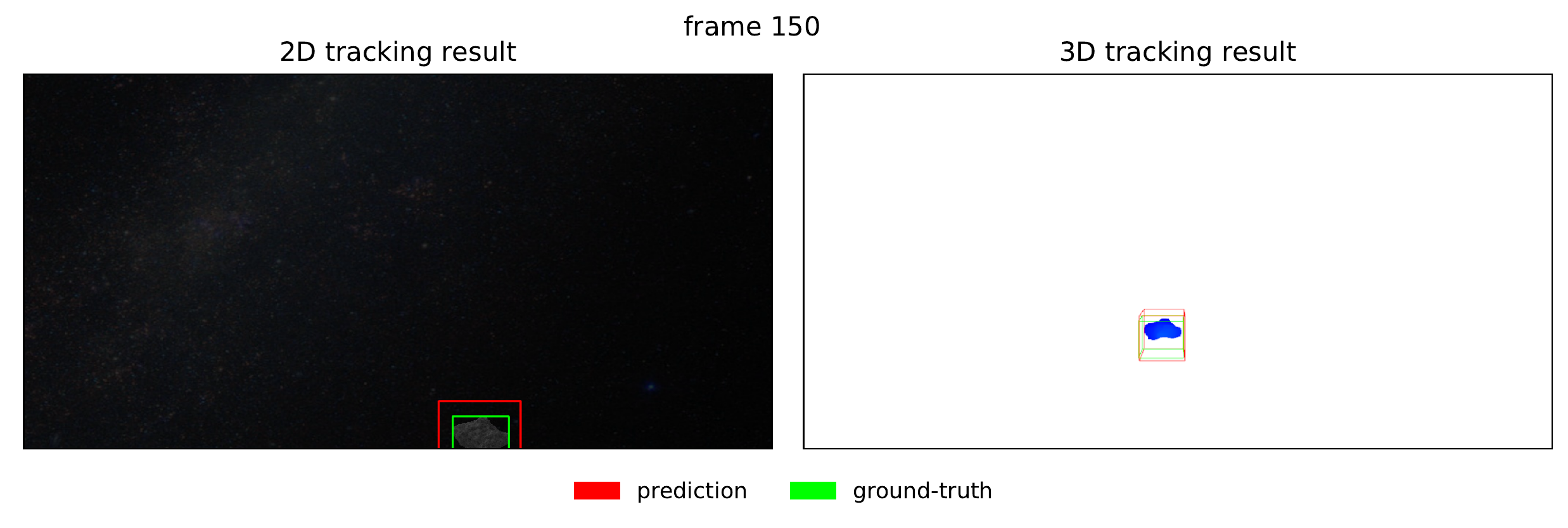} & 
       \includegraphics[width=0.22 \textwidth]{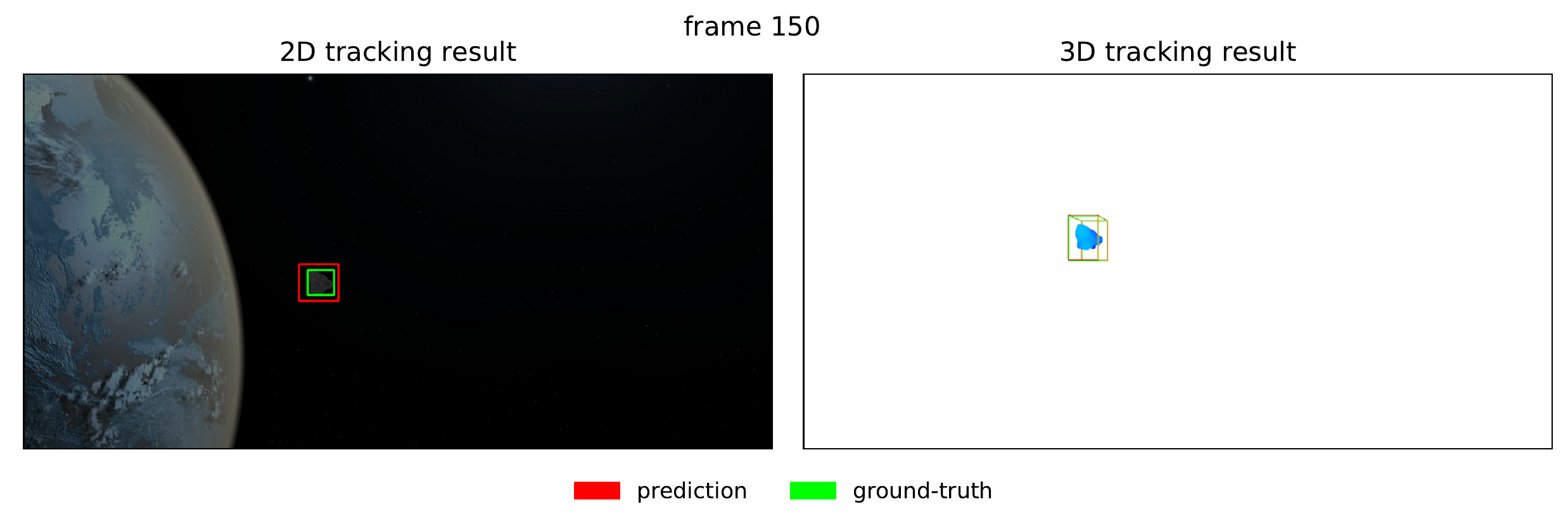} 
       \\
       \includegraphics[width=0.22 \textwidth]{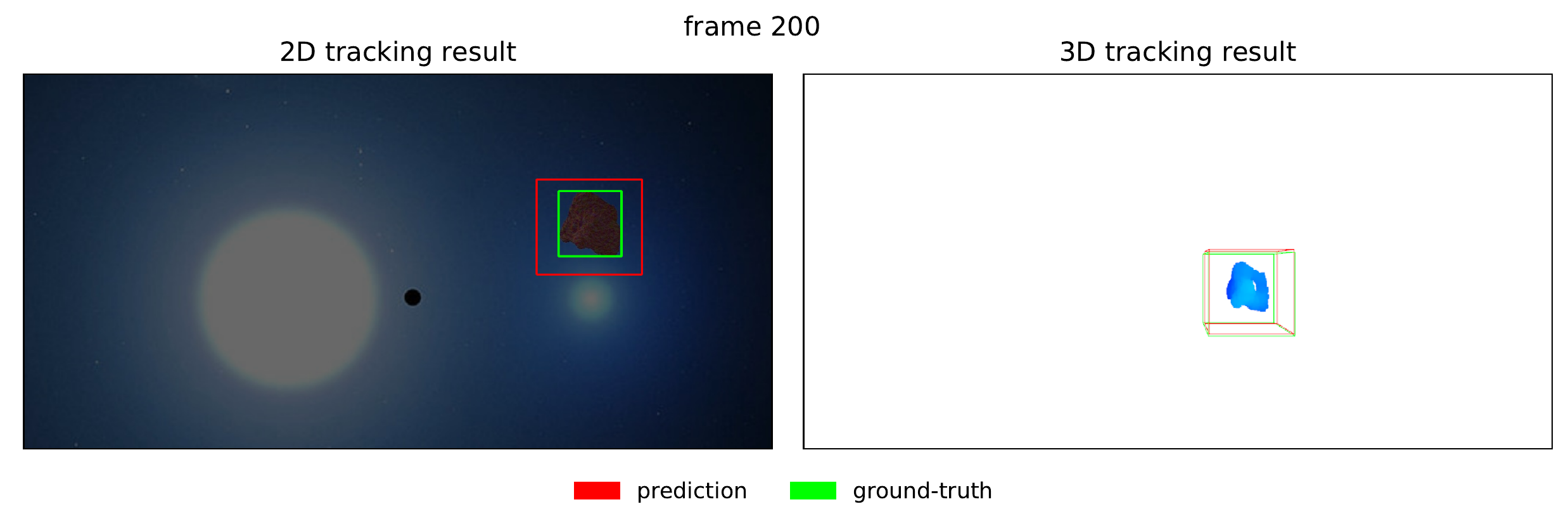} & 
       \includegraphics[width=0.22 \textwidth]{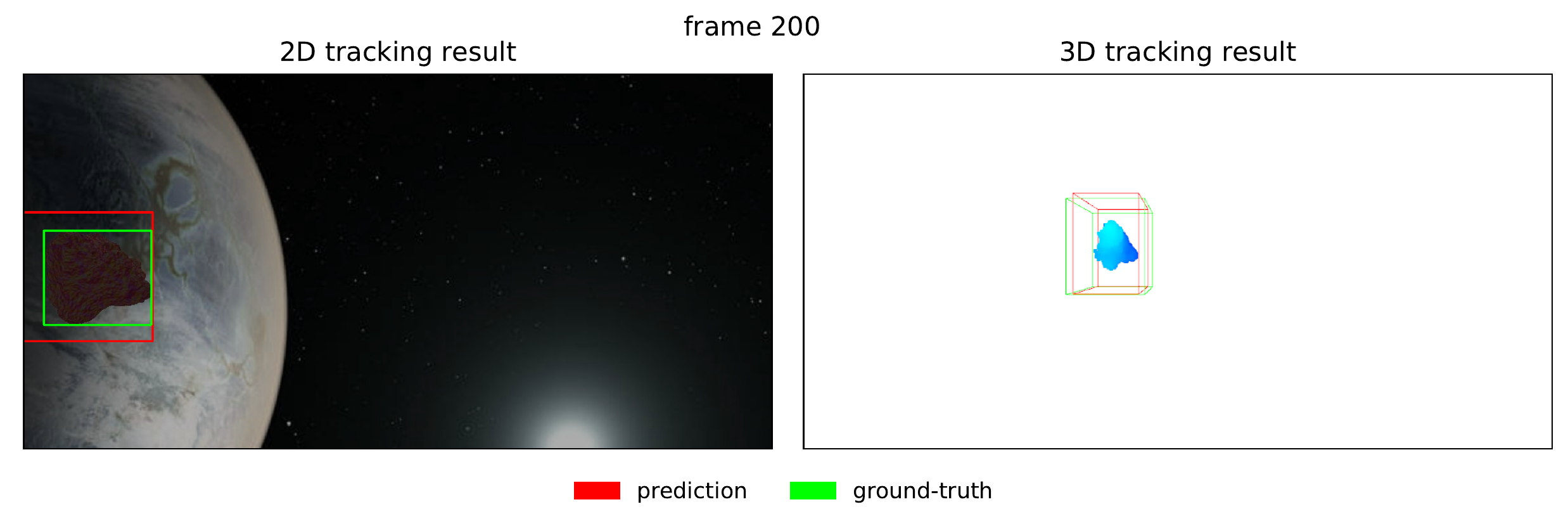} & 
       \includegraphics[width=0.22 \textwidth]{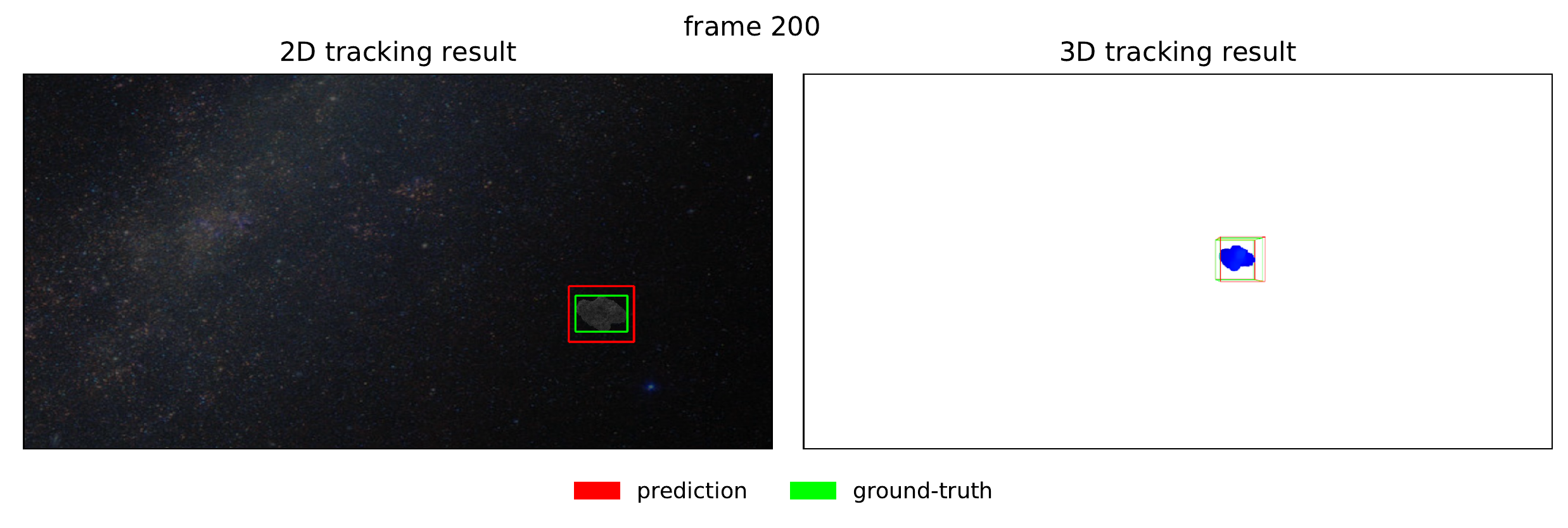} & 
       \includegraphics[width=0.22 \textwidth]{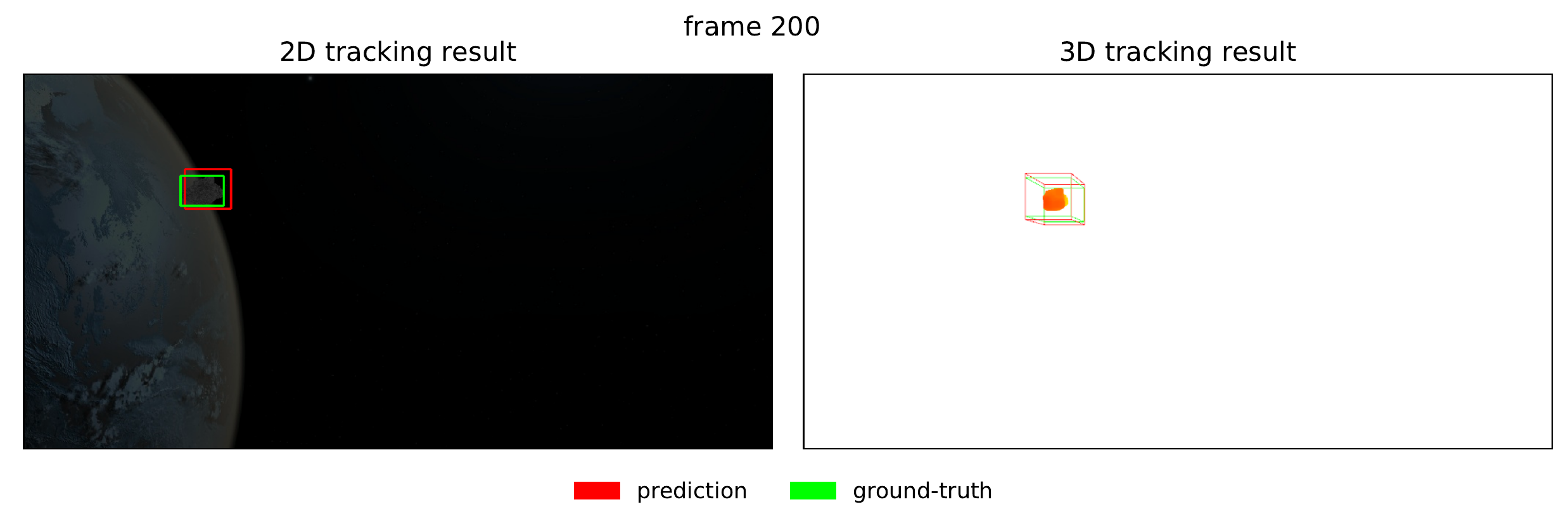} 
   \end{tabular}
   \caption{The tracking results of Track3D on 3D asteroid tracking test set.}
   \label{fig11}
   \centering
\end{figure*}

\begin{figure*}[ht]
    \centering
    \subfloat[sequence 0001]{\includegraphics[width=0.23 \textwidth]{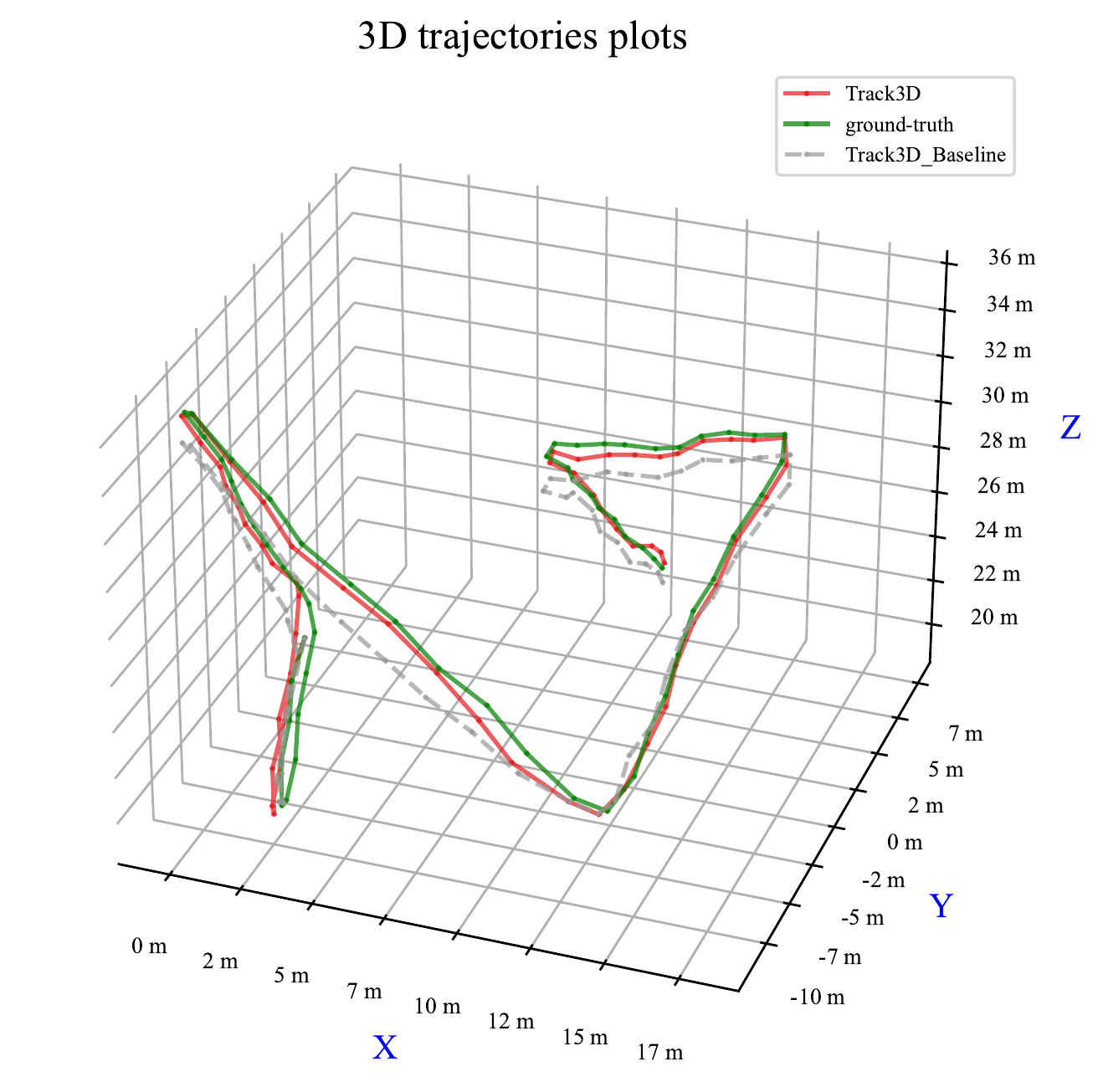}\label{fig12_a}}
    \hfil
    \subfloat[sequence 0022]{\includegraphics[width=0.23 \textwidth]{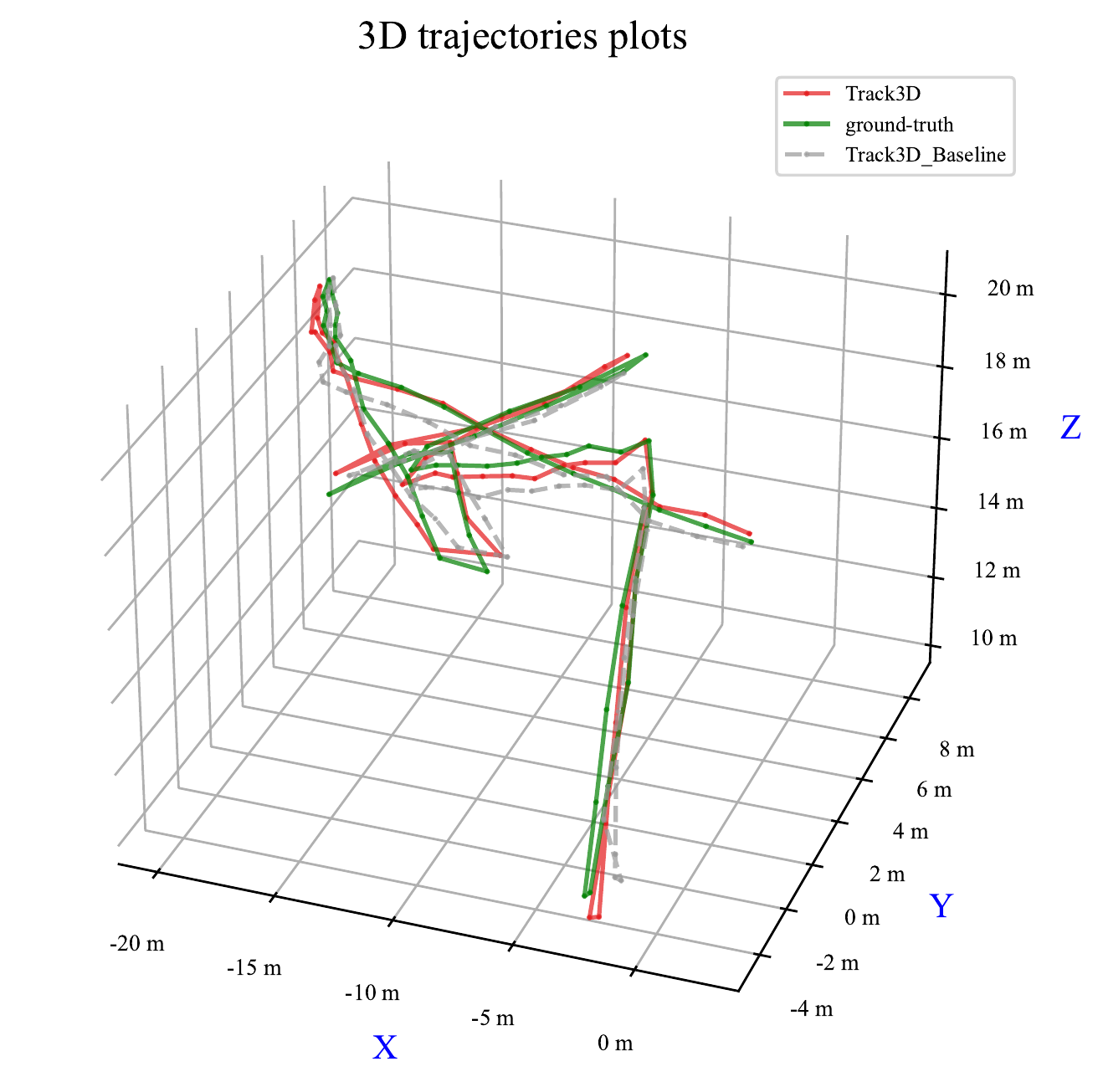}\label{fig12_b}} 
    \hfil
    \subfloat[sequence 0031]{\includegraphics[width=0.23 \textwidth]{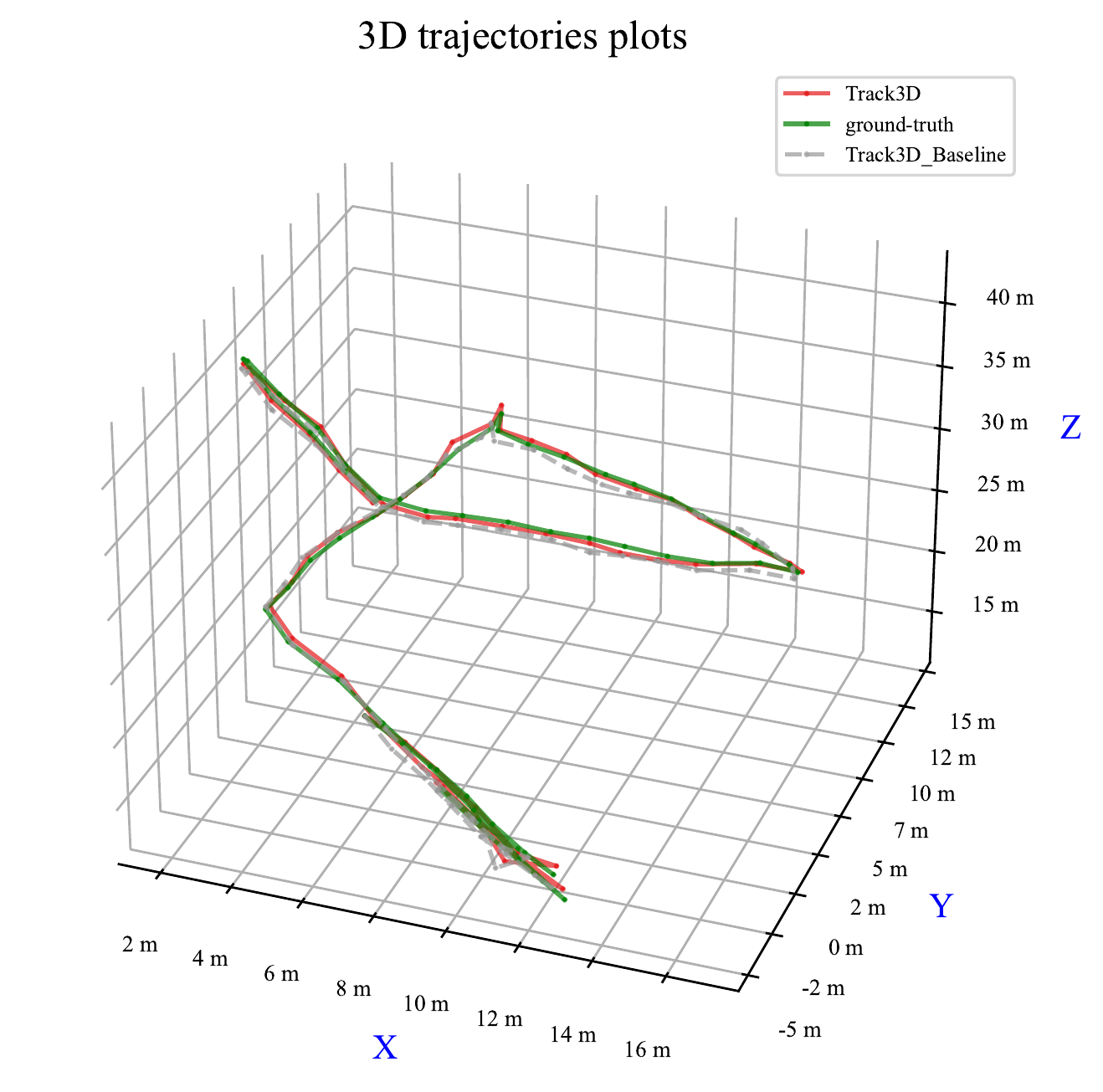}\label{fig12_c}} 
    \hfil
    \subfloat[sequence 0052]{\includegraphics[width=0.23 \textwidth]{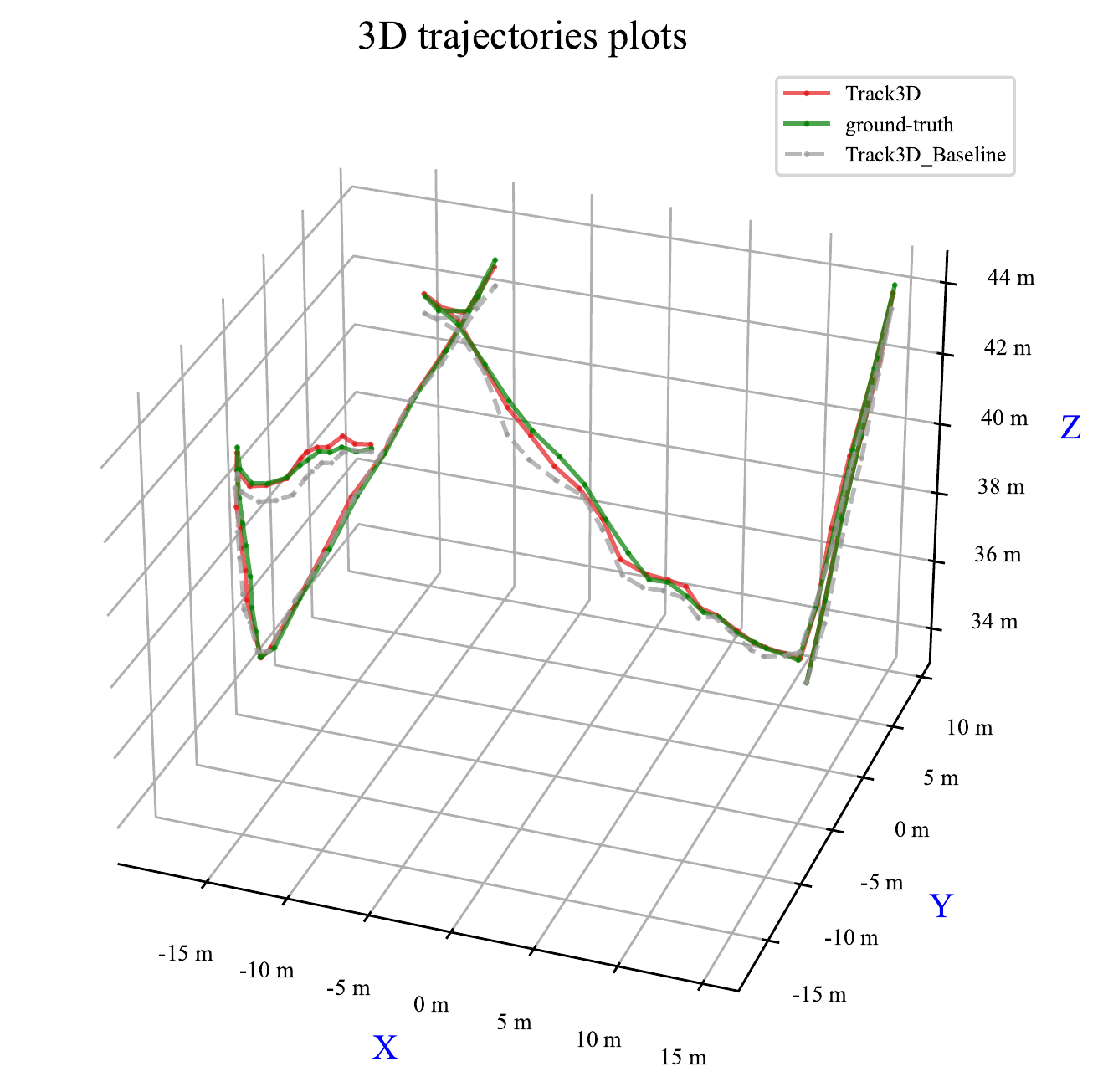}\label{fig12_d}}
    % \subfloat[sequence 0061]{\includegraphics[width=0.25 \textwidth]{imgs/trajectories/0061_001_trajectory.pdf}\label{fig14_e}}
    \caption{3D tracking trajectories on 4 sequences in 3D asteroid tracking test set.}
    \label{fig12}
    \centering
\end{figure*}

\begin{figure*}[t]
    \centering
    \subfloat[2D success plots]{\includegraphics[width=0.4\textwidth]{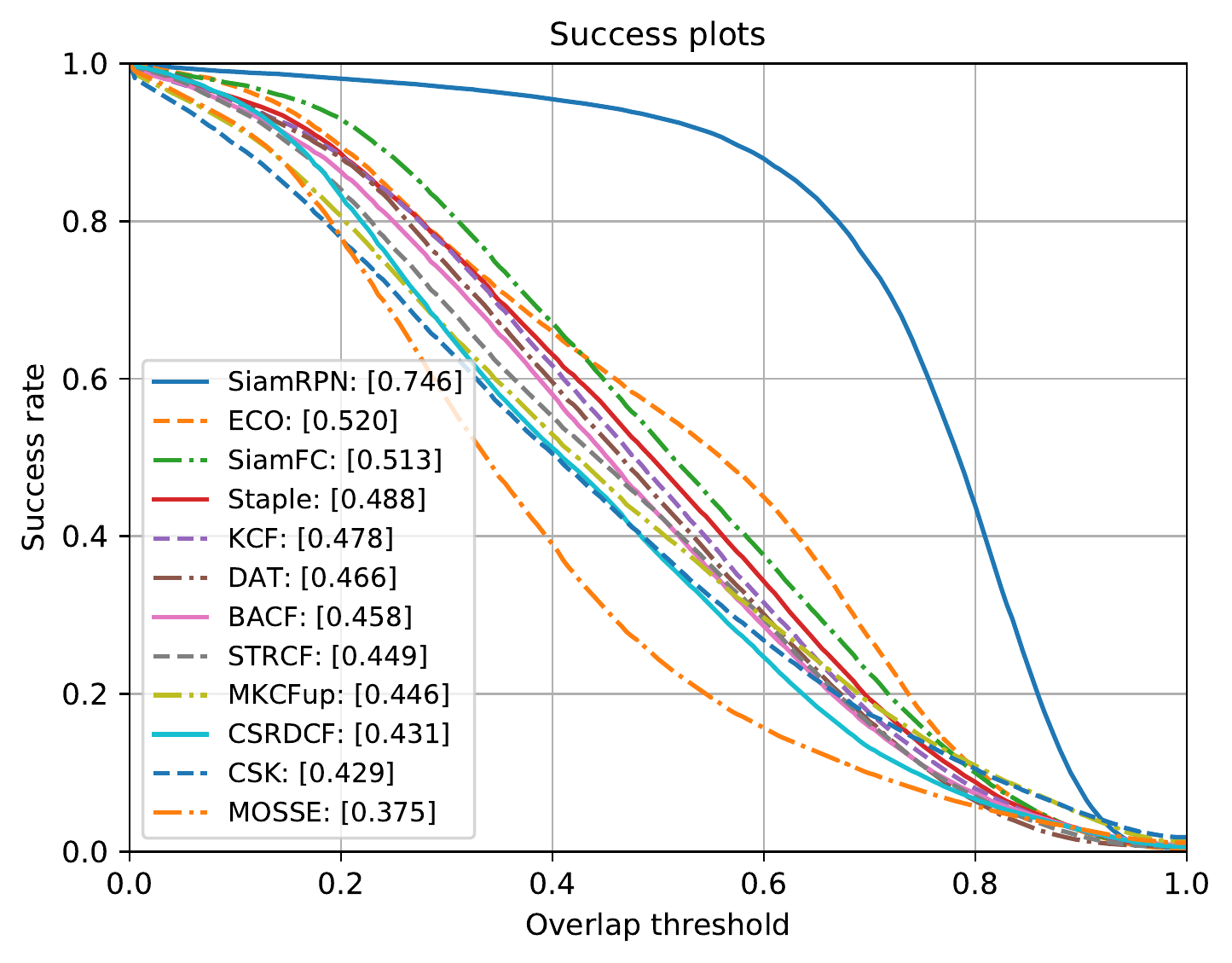} \label{fig14_a}} \hfil
    \subfloat[2D precision plots]{\includegraphics[width=0.4\textwidth]{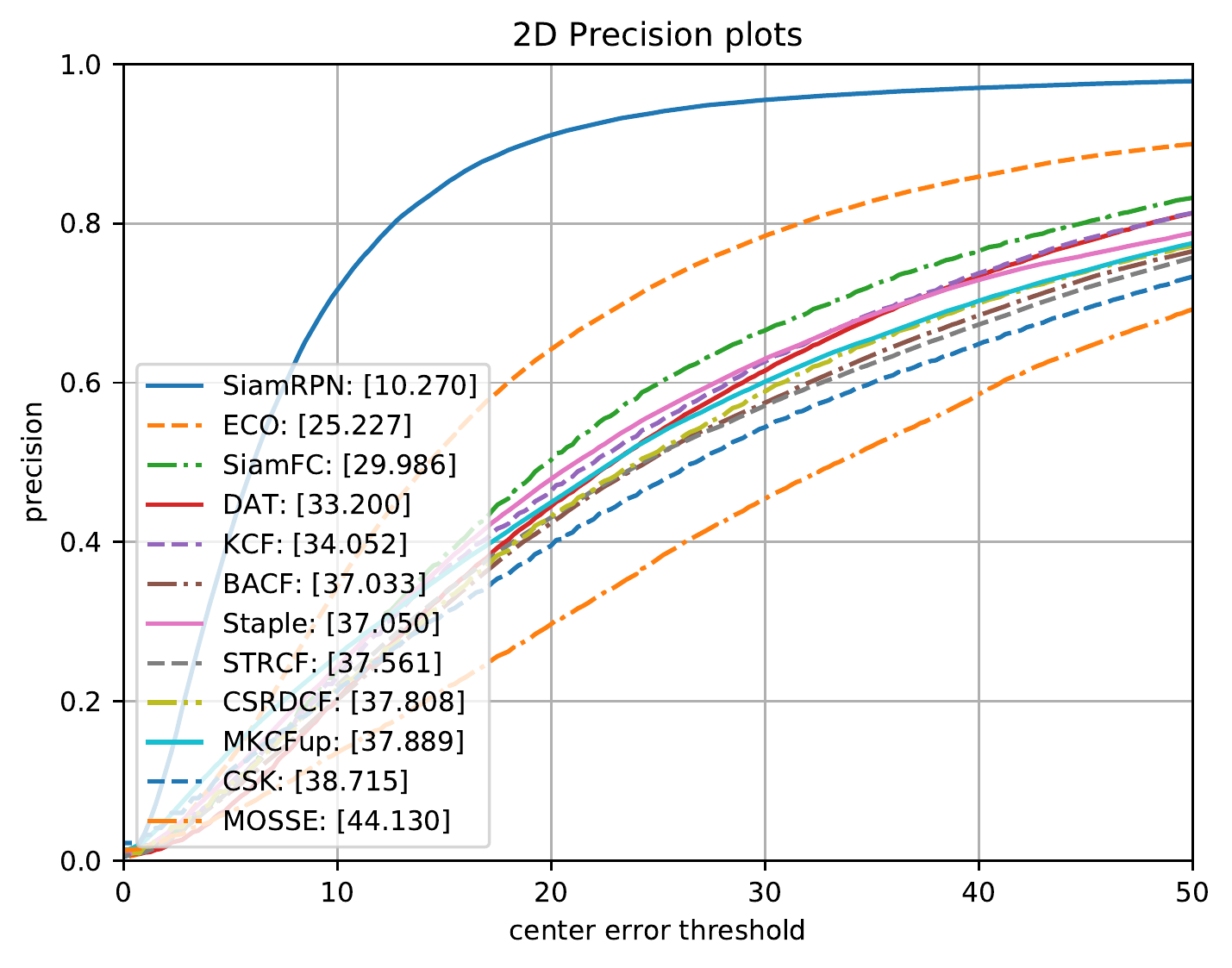} \label{fig14_b}}
    \caption{The 2D tracking performance of 12 classic monocular trackers.} 
    \label{fig14}
    \centering
\end{figure*}

\begin{figure*}[t]
    \centering
    \subfloat[3D success plots]{\includegraphics[width=0.4\textwidth]{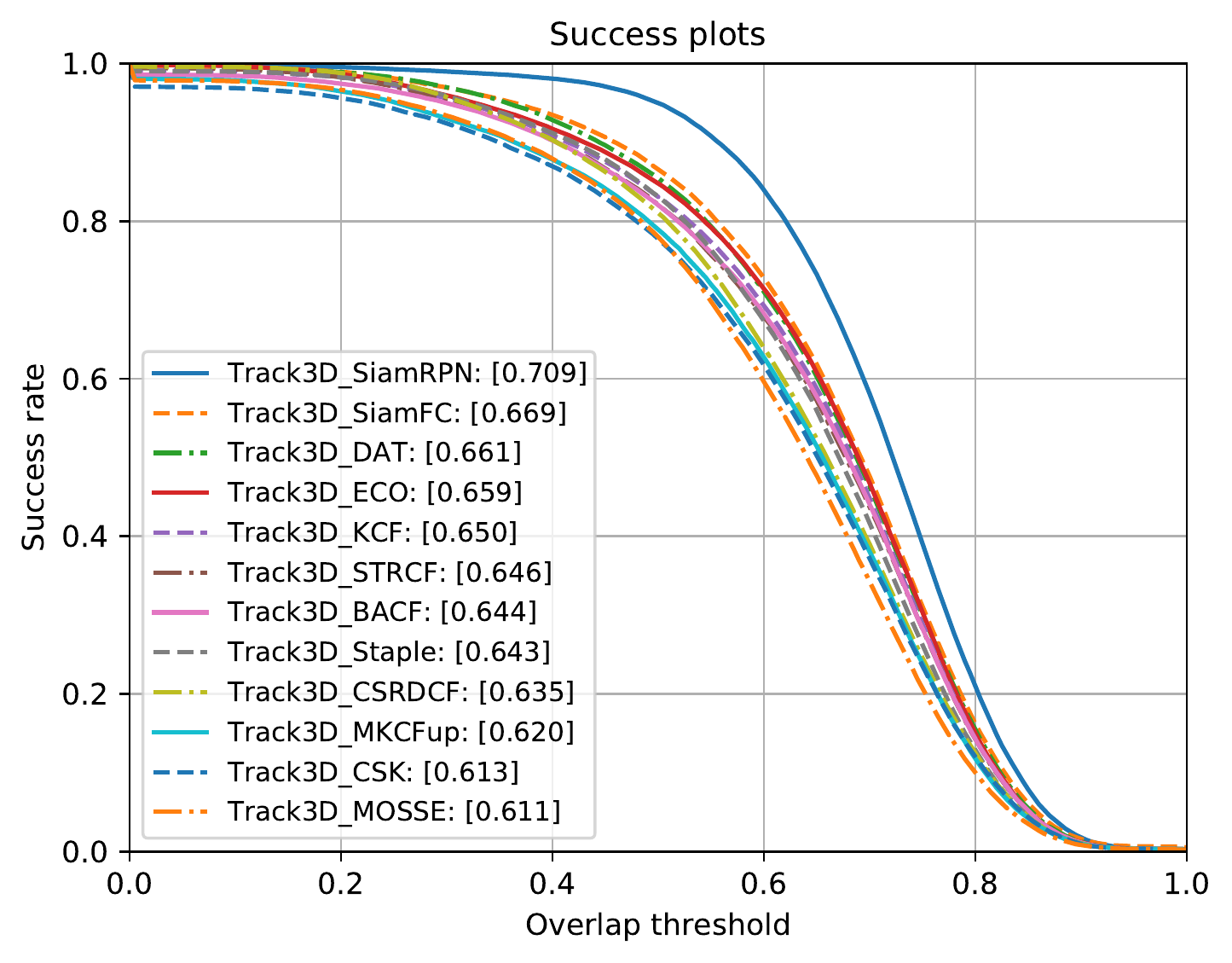} \label{fig15_a}} \hfil
    \subfloat[3D precision plots]{\includegraphics[width=0.4\textwidth]{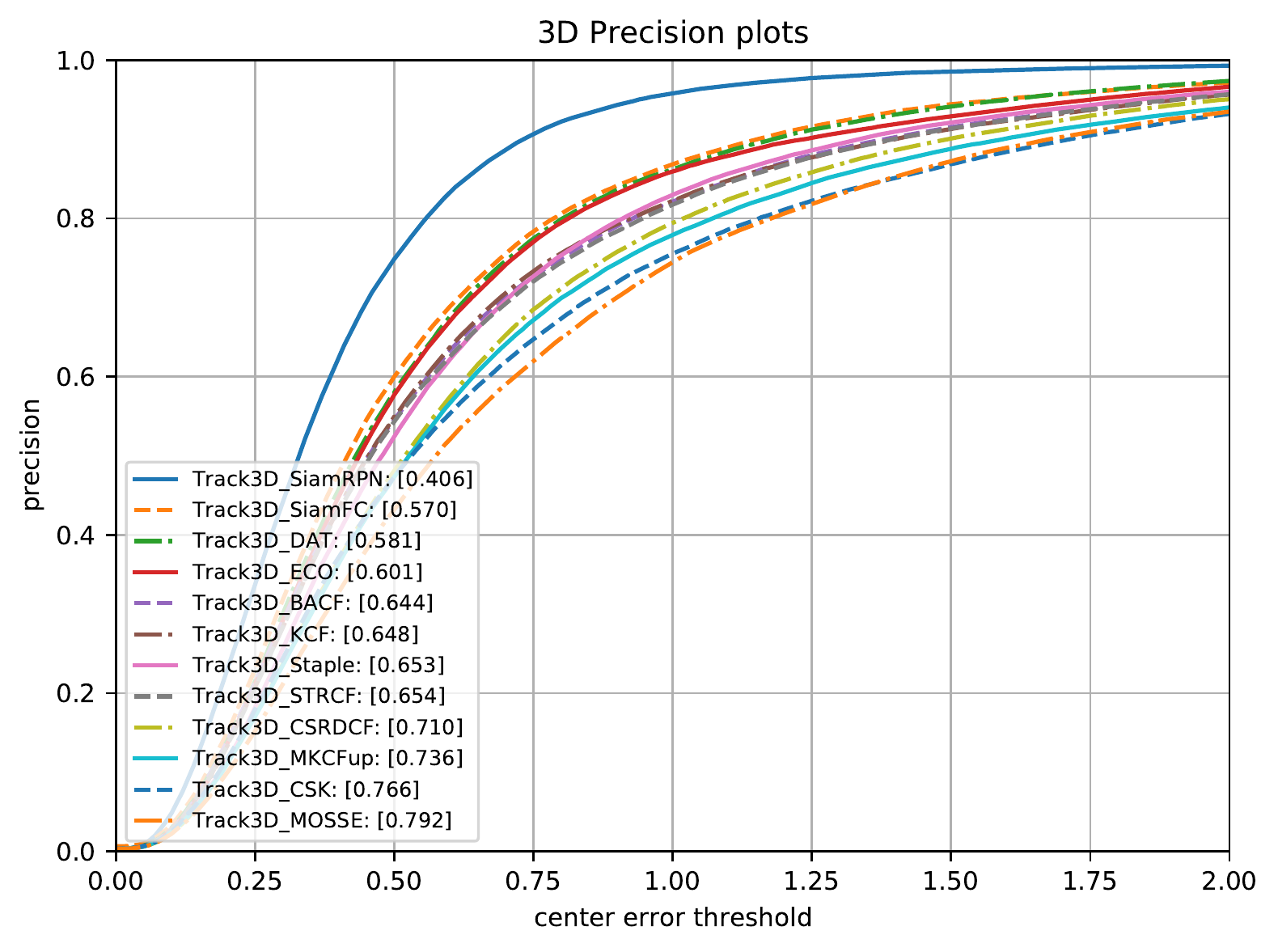} \label{fig15_b}} \vfil
    \caption{The 3D tracking performance of Track3Ds under 12 different monocular trackers.} 
    \label{fig15}
    \centering
\end{figure*}

\begin{figure}[t]
    \centering
        \includegraphics[width=0.45\textwidth]{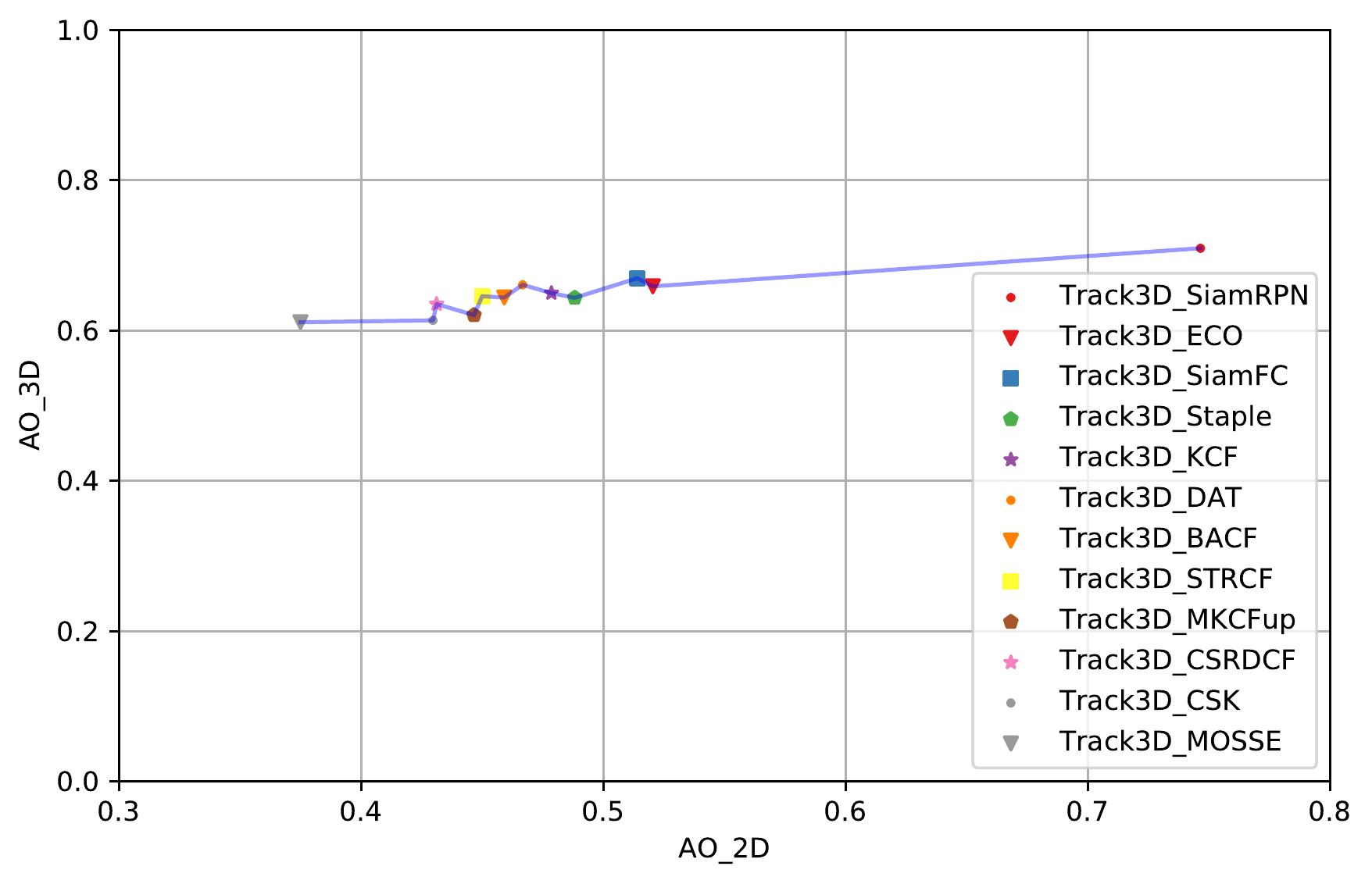}
        \caption{the connection between 3D tracking accuracy and 2D tracking performance, which demonstrates great generalization ability of our framework.}
        \label{fig16}
    \centering
\end{figure}

\section{Experiments \label{section5}}
In this section, we have introduced a simple 3D tracking baseline algorithm as comparison and implemented extensive experiments with Track3D on 3D asteroid tracking dataset to demonstrate the advancement and effectiveness of our framework. We also find out that our framework with 2D-3D fusion strategy can not only handle 3D tracking challenge, but also make improvement on 2D tracking performance. In addition, valuable ablation study is also considered. All the experiments are carried out with Intel i9-9900k@3.60GHz CPU and Nvidia RTX 2080Ti GPU.

\subsection{Framework Performance}
Considering that there are a few researh results in the scope of 3-DoF tracking at present, therefore, we replace the A3BoxNet of our framework with minimum enclosing 3D bounding-box algorithm (i.e. computing the length of aixs-aligned 3D bounding-box by $\max \left\{ P_{\text{frustum}} \right\} - \min \left\{ P_{\text{frustum}} \right\}$) to realise a simple 3D tracking baseline as comparison. Both 2D and 3D tracking evaluation results of 3D tracking baseline are summarized at 3rd row in Table \ref{table2}. And its success plots are shown in Fig. \ref{fig11_a}. We find out that the 3D tracking baseline with 2D-3D fusion strategy improves 2D monocular tracking performance about 10.5\%, however, it achieves quite poor performance in 3D space. 

The overall performance of deep-learning based 3D tracking framework, also named as Track3D, is illustrated in Fig. \ref{fig11_b}. It reaches excellent evaluation results on 3D asteroid tracking test set (0.669 $AO^{3d}$ and 0.570 $ACE^{3d}$) with high real-time performance (77.0 FPS). We visualize part of tracking results of Track3D in Fig. \ref{fig11}, which demonstrates the effectiveness of our 3D tracking method. It clearly shows that our framework can estimate accurate 3D bounding-box even under extreme truncation (see 150-th frame of sequence 0001 in Fig. \ref{fig11}). The 3D trajectory plots in Fig. \ref{fig12} also intuitively show our framework predicts precise 3D object location, which greatly outperforms 3D tracking baseline. In addition, it can be seen in the first colum of Table \ref{table2} that Track3D with 2D-3D fusion strategy can also make significant improvement on 2D tracking performance. 

\subsection{Module Performance}
Extensive experiments are implemented in this subsection that explore how two main modules of Track3D (i.e. 2D monocular tracker and A3BoxNet) make influences on final 3D tracking performance, which points the way to design effective 3D tracking framework for future work. 

Firstly, plenty of classic monocular trackers have been evaluated on left video sequences of 3D asteroid tracking dataset, such as SiamFC \cite{bertinettoFullyconvolutionalSiameseNetworks2016}, SiamRPN \cite{liHighPerformanceVisual2018}, ECO \cite{danelljanECOEfficientConvolution2017}, Staple \cite{bertinettoStapleComplementaryLearners2016}, KCF \cite{henriquesHighSpeedTrackingKernelized2015}, DAT \cite{posseggerDefenseColorbasedModelfree2015}, BACF \cite{kianigaloogahiLearningBackgroundAwareCorrelation2017}, STRCF \cite{liLearningSpatialTemporalRegularized2018}, MKCFup \cite{tangHighSpeedTrackingMultiKernel2018}, CSRDCF \cite{lukezicDiscriminativeCorrelationFilter2017}, CSK \cite{henriquesExploitingCirculantStructure2012}, and MOSSE \cite{bolmeVisualObjectTracking2010}. The evaluation results are illustrated in Fig. \ref{fig14} which straightforward shows the accuracy of 12 monocular trackers are distributed in a large range from 0.375 to 0.746. SiamFC adopted in our framework only achieves intermediate 2D tracking performance in both accuracy and precision metrics.

And then, we evaluate the whole framework under different monocular trackers respectively and plot 3D success and precision curves in Fig. \ref{fig15_a} and \ref{fig15_b}. We found that 3D evaluation curves become much dense comparing with 2D curves of Fig. \ref{fig12}, which denotes that 2D monocular trackers just have slight influence on 3D tracking performance of our framework. In other words, Track3D can still work even based on poor 2D monocular tracker. We further plot the relationship between 2D tracking performance and framework accuracy in Fig. \ref{fig16}, which intuitively demonstrates the generalization ability of Track3D.

\begin{figure}[t]
    \centering
    \includegraphics[width=0.4 \textwidth]{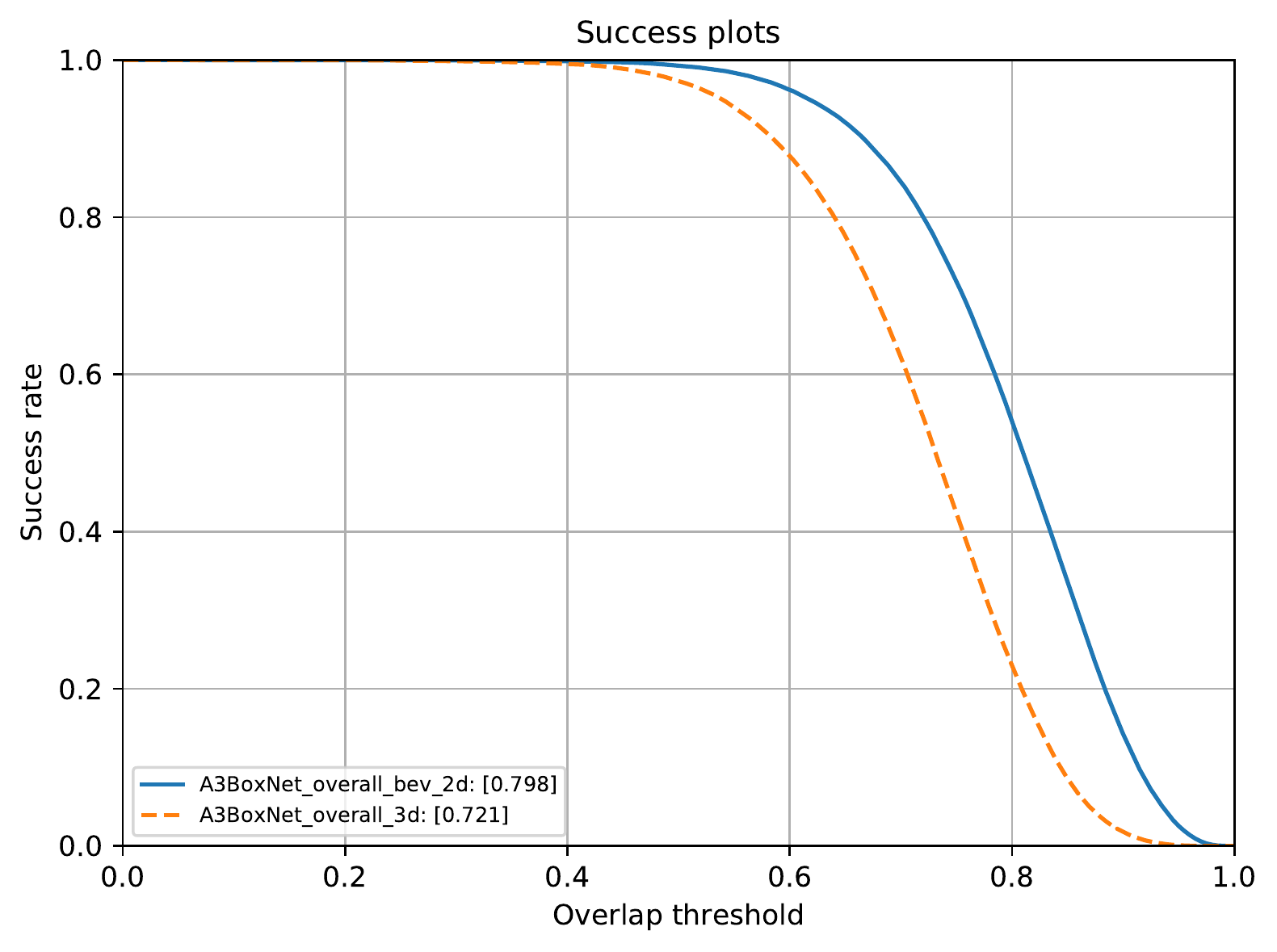} \vfil
    \includegraphics[width=0.4 \textwidth]{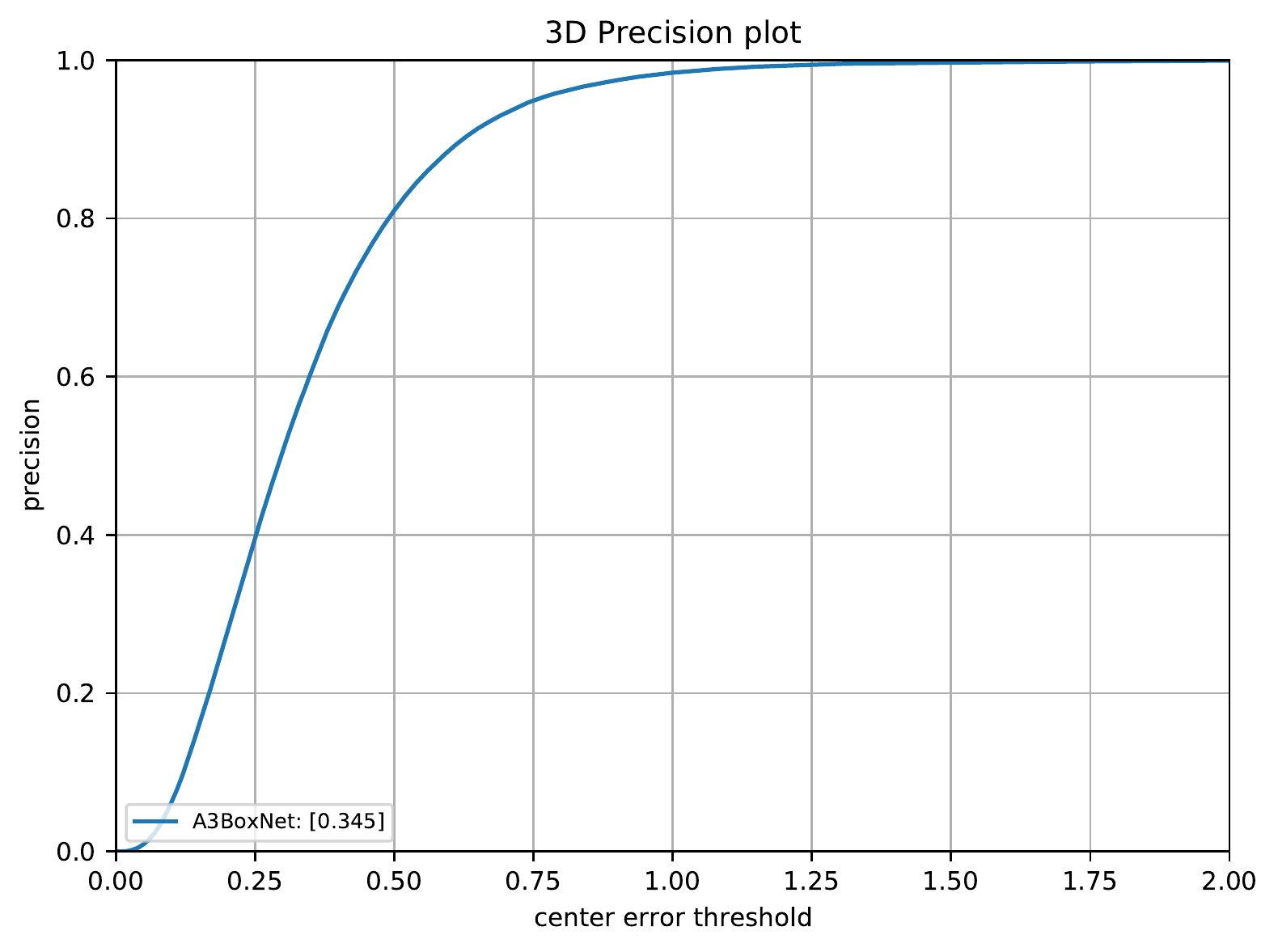}
    \caption{The performance of A3BoxNet with 1024 input points.}
    \label{fig9}
    \centering
\end{figure}

Meanwhile, We evaluate amodal bounding-box estimation network, A3BoxNet, on 3D asteroid tracking dataset by randomly sampling 1024 points from frustum proposal. Its performance is also illustrated in Fig. \ref{fig9}. It can be clearly seen that our A3BoxNet predicts high accurate axis-aligned bounding-box purely with partial object points. Furthermore, our network is able to run at 281.5 FPS, which totally satisfies the requirement of application on edge computing device. 

\begin{figure}[t]
    \centering
    \includegraphics[width=0.4\textwidth]{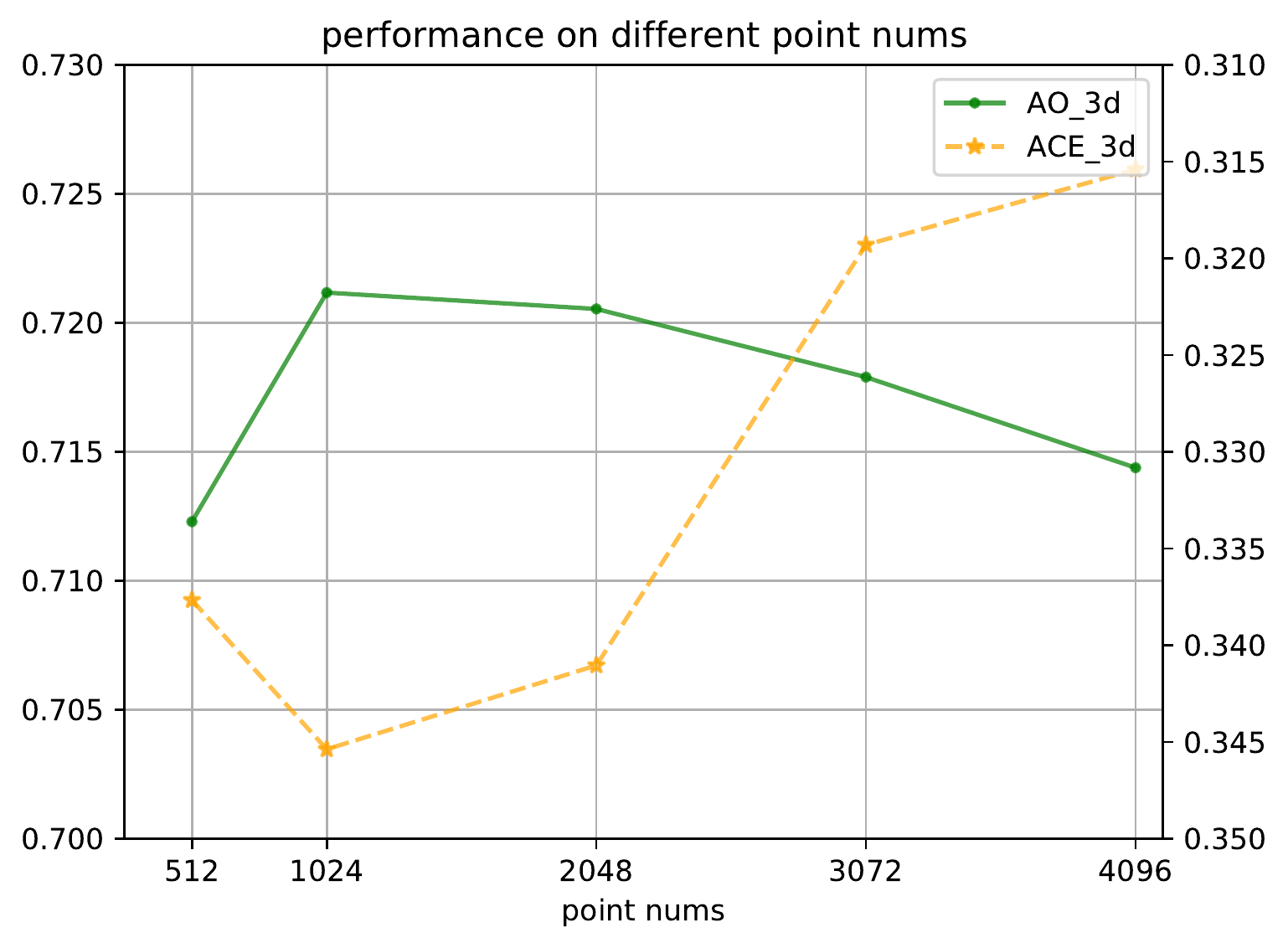}
    \caption{The pefromance of A3BoxNet under different numbers of point set. There is a contradictory between accuracy and precision of A3BoxNet.}
    \label{fig13}
    \centering
\end{figure}

\begin{table}[t]
    \centering
    \caption{The ablation study on one-hot category vector.}
    \label{table3}
    \begin{tabular}{cccc}
        \toprule
        name & $AO^{3d}$ & $AO^{bev}$ & $ACE{3d}$ \\
        \midrule
        A3BoxNet & \textcolor{green}{0.721} & \textcolor{green}{0.798} & 0.345 \\
        without category & 0.715 & 0.792 & \textcolor{green}{0.316} \\
        \bottomrule
    \end{tabular}
    \centering
\end{table}

We also study how the number of input points affects the performance of amodal axis-aligned bounding-box network. We retrain the A3BoxNet by randomly sampling different numbers of point set (e.g. 512, 1024, 2048, 3072, and 4096 points) from object point clouds in 3D asteroid tracking training set. And all the models are trained from scratch with 25 epoches and 32 batch size. The evaluation results are plotted in Fig. \ref{fig13}, which clearly shows accuracy and precision of A3BoxNet have contradictory with the number of input points. Once the accuracy of model increasing, the corresponding precision will decrease. 

Besides, the influence of object category on A3BoxNet is further studied. We remove one-hot vector of object category in center regression network and amodal box estimation network, and retrain A3BoxNet from scratch with 1024 point sets. The performance comparison between original A3BoxNet and A3BoxNet without category information is summarized in Table \ref{table3}, which proves the object category can make a slight improvement on the 3D accuracy of A3BoxNet.

\section{Conclusion \label{section6}}
In this work, we construct the first large-scale 3D asteroid tracking dataset, which involves  148,500 binocular images, depth maps, and point clouds. All the 2D and 3D annotations are automatically generated, which greatly guarantees the quality of tracking dataset and reduce the cost of data collection. The 3D asteroid tracking dataset will be public on website (\url{http://aius.hit.edu.cn/12920/list.htm}). Meanwhile, we propose a deep-learning based 3D visual tracking framework, Track3D, which mainly consists of classic 2D monocular tracker and a novel light-weight amodal axis-aligned bounding-box network. The state-of-the-art 3D tracking performance and great generalization ability of our framework have been demonstrated by sufficient experiments. We also find that Track3D with 2D-3D tracking fusion strategy also makes improvement on 2D tracking performance. In future work, We will further apply our Track3D method to normoal cases like automous driving and robot picking.

\appendices
\section{The perspective camera matrix \label{appendix1}}
In section \ref{section3}, we have mentioned that camera matrix is very important for Track3D to extract frustum proposal. However, physics engine V-rep only provides perspective angle $(\alpha_x, \alpha_y)$ and resolution $(W, H)$. To this end, we derive the camera matrix by perspective projection principle, which is also illustrated in Fig. \ref{fig3}. 

Suppose that the virtual focus of perspective camera in both x and y axes are $(f_x, f_y)$, the size of image plane is $(w, h)$, and an object point $P=(X, Y, Z)$ in camera coordinate system is projected at $p=(x_i, y_i)$ in image plane. Fig. \ref{fig3} clearly shows that:
\begin{equation}
    \frac{X}{Z} = \frac{x_i}{f_x} \label{eq_9}
\end{equation}
and, 
\begin{equation}
    \frac{w/2}{f_x} = \tan\left( \frac{\alpha_x}{2}\right) \label{eq_10}
\end{equation}
Meanwhile, the transformation from image coordinate system to pixel coordinate system in x axis is formulated as:
\begin{align}
    u &= (x_i + \frac{w}{2}) \cdot \frac{W}{w} \label{eq_11} \\ 
    \nonumber &=x_i \cdot \frac{W}{w} + \frac{W}{2} 
\end{align}
Substitute Eq. \ref{eq_9} and \ref{eq_10} into Eq. \ref{eq_11}, it can be obtained:
\begin{equation}
    u = \frac{W}{ 2 \tan(\frac{\alpha_x}{2})Z} X+  \frac{W}{2} \label{eq_12}
\end{equation}
in which, the parameter $f_x$ is eliminated. Similarly, the transformation in y-axis direction from camera coordinate system to pixel coordinate system is also obtained:
\begin{equation}
    v = \frac{H}{ 2 \tan(\frac{\alpha_y}{2})Z} Y+  \frac{H}{2} \label{eq_13}
\end{equation}

We further rewrite Eq. \ref{eq_12} and \ref{eq_13} in homogeneous matrix form:
\begin{align}
    \begin{bmatrix}
       u \\
       v \\
       1
    \end{bmatrix}=
    \begin{bmatrix}
       \frac{W}{ 2 \tan(\frac{\alpha_x}{2})Z} & 0 & \frac{W}{2Z} \\
       0 &   \frac{H}{ 2 \tan(\frac{\alpha_y}{2})Z} & \frac{H}{2Z}\\
       0 & 0 & 1/Z
    \end{bmatrix} 
    \begin{bmatrix}
       X \\
       Y \\
       Z
    \end{bmatrix} \label{eq_14}
\end{align}
To eliminate the $Z$ variable in the transformation matrix, we multiply both sides of Eq. \ref{eq_14} by $Z$:
\begin{align}
    Z\begin{bmatrix}
       u \\
       v \\
       1
    \end{bmatrix}=
    \begin{bmatrix}
       \frac{W}{2 \tan(\alpha_x/2)} & 0 & \frac{W}{2} \\
       0 &   \frac{H}{ 2 \tan(\alpha_y/2)} & \frac{H}{2}\\
       0 & 0 & 1
    \end{bmatrix} 
    \begin{bmatrix}
       X \\
       Y \\
       Z
    \end{bmatrix} 
\end{align}

Because we set left camera coordinate system as reference frame, the camera matrix of left perspective camera can be formulated as:
\begin{align}
    M^L =  \begin{bmatrix}
        \frac{W}{2 \tan(\alpha_x/2)} & 0 & \frac{W}{2} & 0\\
        0 &   \frac{H}{ 2 \tan(\alpha_y/2)} & \frac{H}{2} & 0\\
        0 & 0 & 1 & 0
     \end{bmatrix} 
\end{align}

\section{Traning objectives \label{appendix2}}
In this work, we utilize a joint loss function $\mathcal{L}_{joint}$ to optimize A3BoxNet:
\begin{equation}
    \mathcal{L}_{joint} = \mathcal{L}_{center-net} + \mathcal{L}_{box-net}
\end{equation}
where, $\mathcal{L}_{center-net}$ adopt huber loss function:
\begin{equation}
    \mathcal{L}_{center-net}=
    \left\{
        \begin{array}{ll}
        0.5\alpha^{2}, & \alpha<1 \\
        \alpha - 0.5,  & \text {otherwise}
        \end{array}
    \right.
\end{equation}
in which $\alpha = \left\| \hat C - (\bar C + \Delta C_1)\right\|_2$, $\hat C$ is 3D center label, $\bar c$ is the centroid of points in frustum proposal, and $\Delta C_1$ is the prediction of center regression network. 

In addition, 
\begin{equation}
    \mathcal{L}_{box-net} = \mathcal{L}_{center\_res} +  \mathcal{L}_{size\_cls} + \mathcal{L}_{size\_res}
\end{equation}
where $\mathcal{L}_{size\_cls}$ utilizes softmax cross entropy loss function:
\begin{equation}
    \mathcal{L}_{size\_cls} = -\sum_{i=1}^{N_{size}} \hat{y}_i \cdot \log \left( \frac{e^{y_i}}{\sum_{j=1}^{N_{size}} e^{y_j}}\right)
\end{equation}
in which $\hat{y}$ is $N_{size}$ dimensional one-hot vector of size category label, $y$ is the partial outputs of amodal box estimation network, of which dimension is also $N_{size}$. 

Furthermore, $\mathcal{L}_{center\_res}$ and $\mathcal{L}_{size\_res}$ both use huber loss function. $\mathcal{L}_{center\_res}$ is formulated as:
\begin{equation}
    \mathcal{L}_{center-net}=
    \left\{
        \begin{array}{ll}
        0.5\beta^{2}, & \beta<2 \\
        2(\beta - 1),  & \text {otherwise}
        \end{array}
    \right.
\end{equation}
in which, $\beta = \left\| \hat C - (\bar C + \Delta C_1 + \Delta C_2)\right\|_2$, $\Delta C_2$ is 3D center residuals predicted by amodal box estimation network. And $\mathcal{L}_{size\_res}$ is as follows:
\begin{equation}
    \mathcal{L}_{center-net}=
    \left\{
        \begin{array}{ll}
        0.5 \gamma^{2}, & \gamma<1 \\
        \gamma - 0.5,  & \text {otherwise}
        \end{array}
    \right.
\end{equation}
where $\gamma = \left\| \hat R - r * \max \left\{\hat S\right\}\right\|_2$, $\hat S$ and $\hat R$ are the size and size residual label, respectively. $r$ is normalized size residual corresponding to the size category predicted by amodal box estimation network. 

\section*{Acknowledgment}
This work was kindly supported by the National Key R\&D Program of China through grant 2019YFB1312001.

% \ifCLASSOPTIONcaptionsoff
%   \newpage
% \fi

\bibliographystyle{IEEEtran}
\bibliography{IEEEabrv, references, my_work, Object_Tracking}

\newpage

\begin{IEEEbiography}[{\includegraphics[width=0.9in,height=1.25in,clip]{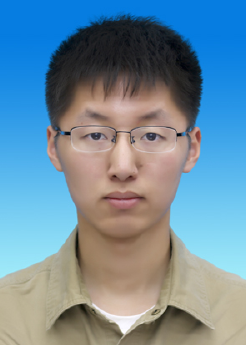}}]{Dong Zhou}
	was born in Hunan, China, in 1996. He received the B.S degree in automation from Harbin Engineering University, Harbin, China, in 2018. He is currently working toward the Ph.D. degree in the Department of Control Science and Engineering, Harbin Institute of Technology, Harbin, China. His research interests include space non-cooperative object visual tracking, 3D computer vision, and deep learning. 
\end{IEEEbiography}

\vspace{-120 mm}

\begin{IEEEbiography}[{\includegraphics[width=0.9in,height=1.25in,clip,keepaspectratio]{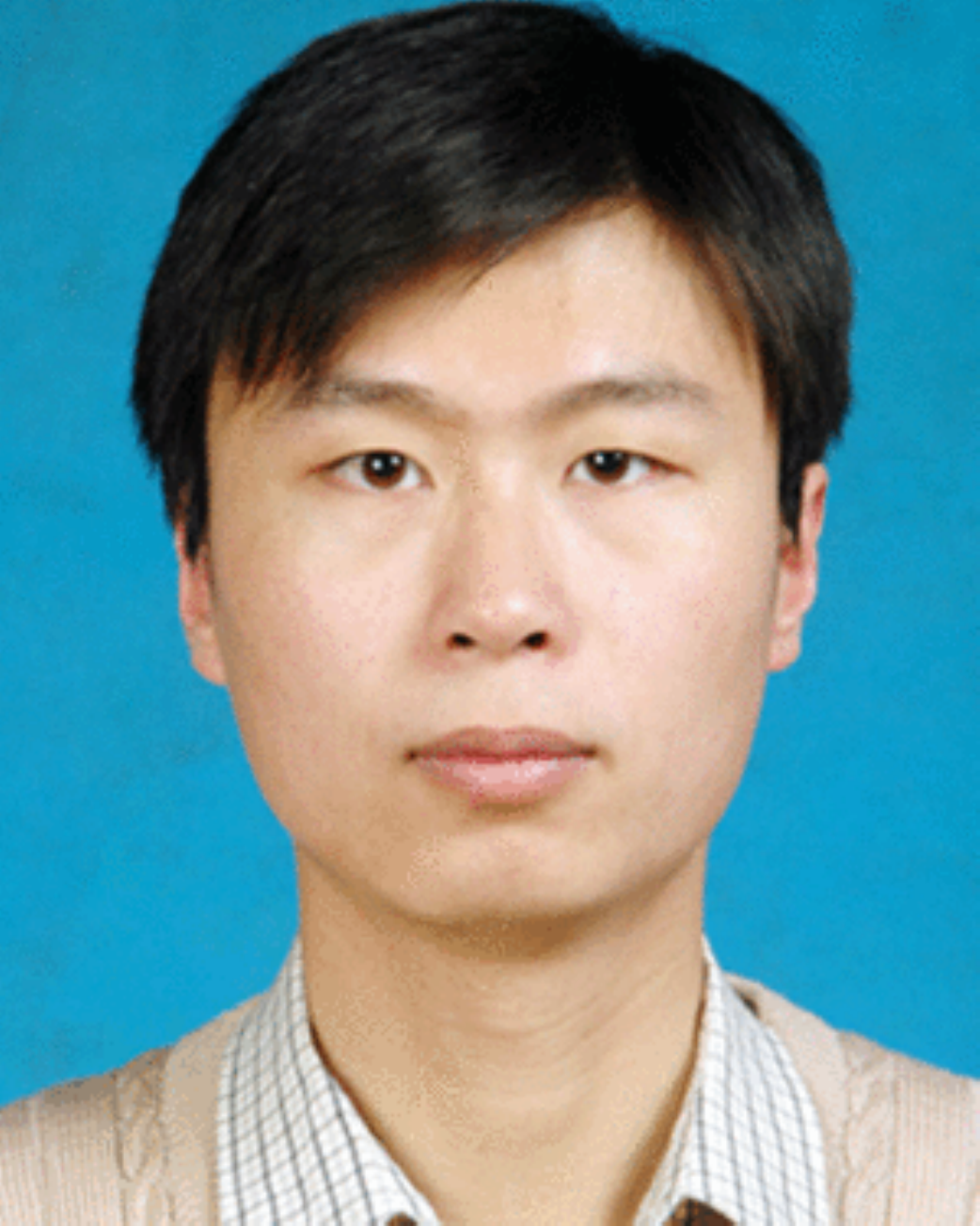}}]{Guanghui Sun}
	 was born in Henan Province, China, in 1983. He received the B.S. degree in Automation from Harbin Institute of Technology, Harbin, China, in 2005, and the M.S. and Ph.D. degrees in Control Science and Engineering from Harbin Institute of Technology, Harbin, China, in 2007 and 2010, respectively. He is currently a professor with Department of Control Science and Engineering in Harbin Institute of Technology, Harbin, China. His research interests include machine learning, computer vision, and aerospace technology.
\end{IEEEbiography}

\vspace{-120 mm}

\begin{IEEEbiography}[{\includegraphics[width=1in,height=1.25in,clip,keepaspectratio]{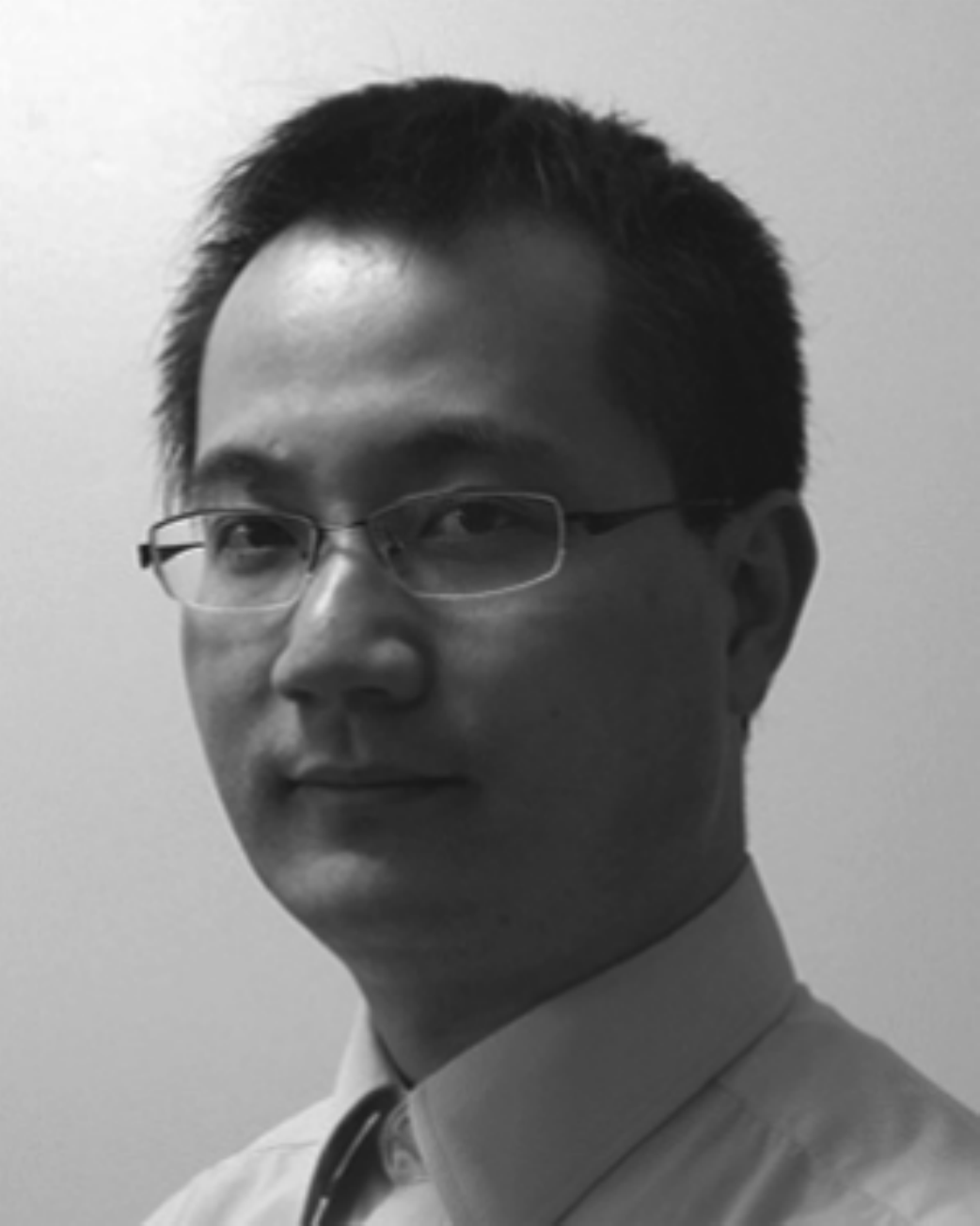}}]{Xiaopeng Hong}
	received the Ph.D. degree in computer application and technology from the Harbin Institute of Technology, China, in 2010. He was a Docent with Center for Machine Vision and Signal Analysis, University of Oulu, Finland, where he had been a Scientist Researcher from 2011 to 2018. He is currently a Distinguished Research Fellow with Xi’an Jiaotong University, China. He has published over 30 articles in mainstream journals and conferences such as the IEEE T-PAMI, T-IP, CVPR, ICCV, AAAI, and ACM UbiComp. His current research interests include multi-modal learning, affective computing, intelligent medical examination, and human-computer interaction. 
\end{IEEEbiography}

\end{document}